\documentclass{article}

\usepackage[preprint]{neurips_2026}

\usepackage[utf8]{inputenc} %
\usepackage[T1]{fontenc}    %
\usepackage[hidelinks]{hyperref}       %
\usepackage{url}            %
\usepackage{booktabs}       %
\usepackage{amsfonts}       %
\usepackage{nicefrac}       %
\usepackage{microtype}      %
\usepackage{xcolor}         %
\usepackage{colortbl}      %
\usepackage[table]{xcolor} %
\usepackage{enumitem}

\usepackage{microtype}
\usepackage{graphicx}
\usepackage{placeins}
\usepackage{subcaption}
\usepackage[labelfont=bf]{caption}
\usepackage{booktabs} %
\usepackage{multirow} %
\usepackage[dvipsnames]{xcolor} %

\usepackage{xurl} %

\usepackage{pgfplots}
\usepgfplotslibrary{groupplots}
\pgfplotsset{compat=1.18}
\usepgfplotslibrary{patchplots}
\tikzset{warmupline/.style={thick, dashed, red}, earlystopline/.style={thick, dashed, blue}}
\usetikzlibrary{arrows.meta,positioning,fit,calc,patterns}
\usepackage{svg}

\usepackage{pifont}
\newcommand{\xmark}{\ding{55}}
\newcommand{\cmark}{\ding{51}}

\usepackage{amsmath}
\usepackage{amssymb}
\usepackage{mathtools}
\usepackage{amsthm}
\usepackage{enumitem}

\usepackage{enumitem}

\usepackage[capitalize,noabbrev]{cleveref}

\theoremstyle{plain}

\theoremstyle{definition}

\theoremstyle{remark}

\usepackage[textsize=tiny]{todonotes}

\usepackage{tikz}
\usetikzlibrary{shapes,calc,decorations.pathreplacing,arrows.meta,tikzmark,positioning,fit,patterns}

\usepackage{amssymb}
\newcommand{\circled}[1]{\raisebox{.5pt}{\textcircled{\raisebox{-.9pt}{\small #1}}}}

\newcommand{\ModelName}{CoExVQA}
\newcommand{\ModelSpell}{Chain-of-Explanation-guided DocVQA}

\newcommand{\ImgEmb}{E_I} %
\newcommand{\GatedImgEmb}{\tilde{E_I}} %

\newcommand{\QMask}{{\hat{H}_Q}}                %
\newcommand{\QMaskPrior}{{H_Q}}    %

\newcommand{\QMaskN}{{\hat{H}_{Q,n}}}

\newcommand{\ABox}{{\hat{b}_A}}            %
\newcommand{\ABoxPrior}{b_A} %
\newcommand{\PredA}{\hat{y}}  %
\newcommand{\A}{y}  %

\newcommand{\AboxC}[1]{{\hat{c}_{n}}}

\newcommand{\Aboxw}{\hat{w}}
\newcommand{\Aboxh}{\hat{h}}
\newcommand{\AboxPriorC}[1]{{c_{n}}}

\newcommand{\AboxPriorw}{w}
\newcommand{\AboxPriorh}{h}

\newcommand{\QProj}{P_Q}      
\newcommand{\AProj}{P_A}

\newcommand{\Enc}{\mathrm{Enc}}
\newcommand{\Dec}{\mathrm{Dec}}

\newcommand{\Mask}{\mathcal{M}}

\newcounter{rq}

\newcommand{\UserQ}[1]{\hypertarget{user_eval_q:#1}{\textbf{(#1)}}}
\newcommand{\UserQref}[1]{\hyperlink{user_eval_q:#1}{\textbf{#1}}}

\newcommand{\UserEvalParticipants}{17}

\newcommand{\NumStudents}{11}
\newcommand{\NumAcademia}{1}
\newcommand{\NumIndustry}{2}
\newcommand{\NumOther}{3}

\newcommand{\NumUnfamiliar}{1} %
\newcommand{\NumBasic}{6} %
\newcommand{\NumInformed}{6} %
\newcommand{\NumAdvanced}{4} %

\newcommand{\NumVeryLittle}{3} %
\newcommand{\NumSome}{8} %
\newcommand{\NumALot}{4} %
\newcommand{\NumMost}{2} %

\pgfmathsetmacro{\StudentsPCT}{round(1000*\NumStudents/\UserEvalParticipants)/10}
\pgfmathsetmacro{\AcademiaPCT}{round(1000*\NumAcademia/\UserEvalParticipants)/10}
\pgfmathsetmacro{\IndustryPCT}{round(1000*\NumIndustry/\UserEvalParticipants)/10}
\pgfmathsetmacro{\OtherPCT}{round(1000*\NumOther/\UserEvalParticipants)/10}

\pgfmathsetmacro{\UnfamiliarPCT}{round(1000*\NumUnfamiliar/\UserEvalParticipants)/10}
\pgfmathsetmacro{\BasicPCT}{round(1000*\NumBasic/\UserEvalParticipants)/10}
\pgfmathsetmacro{\InformedPCT}{round(1000*\NumInformed/\UserEvalParticipants)/10}
\pgfmathsetmacro{\AdvancedPCT}{round(1000*\NumAdvanced/\UserEvalParticipants)/10}

\pgfmathsetmacro{\VeryLittlePCT}{round(1000*\NumVeryLittle/\UserEvalParticipants)/10}
\pgfmathsetmacro{\SomePCT}{round(1000*\NumSome/\UserEvalParticipants)/10}
\pgfmathsetmacro{\ALotPCT}{round(1000*\NumALot/\UserEvalParticipants)/10}
\pgfmathsetmacro{\MostPCT}{round(1000*\NumMost/\UserEvalParticipants)/10}

\newcommand{\BinaryGoodExamples}{4} 
\newcommand{\BinaryBadExamples}{2}
\newcommand{\BinaryExamples}{6}

\newcommand{\ExOneSufficientYes}{15}
\newcommand{\ExOneSufficientNo}{2}

\newcommand{\ExOneCorrectYes}{15}
\newcommand{\ExOneCorrectNo}{2}

\newcommand{\ExOneConfOne}{1}
\newcommand{\ExOneConfTwo}{0}
\newcommand{\ExOneConfThree}{0}
\newcommand{\ExOneConfFour}{0}
\newcommand{\ExOneConfFive}{0}
\newcommand{\ExOneConfSix}{8}
\newcommand{\ExOneConfSeven}{8}

\newcommand{\ExOneInfAnswerBox}{10}
\newcommand{\ExOneInfHeatmap}{6}
\newcommand{\ExOneInfBoth}{6}
\newcommand{\ExOneInfUnclear}{1}
\newcommand{\ExOneInfReading}{6}
\newcommand{\ExOneInfPlausible}{2}
\newcommand{\ExOneInfOther}{0}

\newcommand{\ExOneRectOne}{1}
\newcommand{\ExOneRectTwo}{0}
\newcommand{\ExOneRectThree}{0}
\newcommand{\ExOneRectFour}{1}
\newcommand{\ExOneRectFive}{3}
\newcommand{\ExOneRectSix}{1}
\newcommand{\ExOneRectSeven}{11}

\newcommand{\ExOneHeatOne}{2}
\newcommand{\ExOneHeatTwo}{2}
\newcommand{\ExOneHeatThree}{0}
\newcommand{\ExOneHeatFour}{2}
\newcommand{\ExOneHeatFive}{4}
\newcommand{\ExOneHeatSix}{6}
\newcommand{\ExOneHeatSeven}{1}

\newcommand{\ExTwoSufficientYes}{14}
\newcommand{\ExTwoSufficientNo}{3}

\newcommand{\ExTwoCorrectYes}{14}
\newcommand{\ExTwoCorrectNo}{3}

\newcommand{\ExTwoConfOne}{0}
\newcommand{\ExTwoConfTwo}{0}
\newcommand{\ExTwoConfThree}{0}
\newcommand{\ExTwoConfFour}{1}
\newcommand{\ExTwoConfFive}{2}
\newcommand{\ExTwoConfSix}{3}
\newcommand{\ExTwoConfSeven}{11}

\newcommand{\ExTwoInfAnswerBox}{12}
\newcommand{\ExTwoInfHeatmap}{0}
\newcommand{\ExTwoInfBoth}{1}
\newcommand{\ExTwoInfUnclear}{2}
\newcommand{\ExTwoInfReading}{5}
\newcommand{\ExTwoInfPlausible}{4}
\newcommand{\ExTwoInfOther}{0}

\newcommand{\ExTwoRectOne}{0}
\newcommand{\ExTwoRectTwo}{0}
\newcommand{\ExTwoRectThree}{1}
\newcommand{\ExTwoRectFour}{0}
\newcommand{\ExTwoRectFive}{0}
\newcommand{\ExTwoRectSix}{4}
\newcommand{\ExTwoRectSeven}{12}

\newcommand{\ExTwoHeatOne}{4}
\newcommand{\ExTwoHeatTwo}{3}
\newcommand{\ExTwoHeatThree}{4}
\newcommand{\ExTwoHeatFour}{1}
\newcommand{\ExTwoHeatFive}{2}
\newcommand{\ExTwoHeatSix}{2}
\newcommand{\ExTwoHeatSeven}{1}

\newcommand{\ExThreeSufficientYes}{17}
\newcommand{\ExThreeSufficientNo}{0}

\newcommand{\ExThreeCorrectYes}{17}
\newcommand{\ExThreeCorrectNo}{0}

\newcommand{\ExThreeConfOne}{0}
\newcommand{\ExThreeConfTwo}{0}
\newcommand{\ExThreeConfThree}{0}
\newcommand{\ExThreeConfFour}{0}
\newcommand{\ExThreeConfFive}{0}
\newcommand{\ExThreeConfSix}{4}
\newcommand{\ExThreeConfSeven}{13}

\newcommand{\ExThreeInfAnswerBox}{11}
\newcommand{\ExThreeInfHeatmap}{8}
\newcommand{\ExThreeInfBoth}{6}
\newcommand{\ExThreeInfUnclear}{1}
\newcommand{\ExThreeInfReading}{4}
\newcommand{\ExThreeInfPlausible}{2}
\newcommand{\ExThreeInfOther}{0}

\newcommand{\ExThreeRectOne}{0}
\newcommand{\ExThreeRectTwo}{0}
\newcommand{\ExThreeRectThree}{0}
\newcommand{\ExThreeRectFour}{0}
\newcommand{\ExThreeRectFive}{1}
\newcommand{\ExThreeRectSix}{3}
\newcommand{\ExThreeRectSeven}{13}

\newcommand{\ExThreeHeatOne}{0}
\newcommand{\ExThreeHeatTwo}{0}
\newcommand{\ExThreeHeatThree}{0}
\newcommand{\ExThreeHeatFour}{0}
\newcommand{\ExThreeHeatFive}{6}
\newcommand{\ExThreeHeatSix}{4}
\newcommand{\ExThreeHeatSeven}{7}

\newcommand{\ExFourSufficientYes}{15}
\newcommand{\ExFourSufficientNo}{2}

\newcommand{\ExFourCorrectYes}{14}
\newcommand{\ExFourCorrectNo}{3}

\newcommand{\ExFourConfOne}{0}
\newcommand{\ExFourConfTwo}{0}
\newcommand{\ExFourConfThree}{1}
\newcommand{\ExFourConfFour}{0}
\newcommand{\ExFourConfFive}{1}
\newcommand{\ExFourConfSix}{6}
\newcommand{\ExFourConfSeven}{9}

\newcommand{\ExFourInfAnswerBox}{8}
\newcommand{\ExFourInfHeatmap}{1}
\newcommand{\ExFourInfBoth}{2}
\newcommand{\ExFourInfUnclear}{3}
\newcommand{\ExFourInfReading}{10}
\newcommand{\ExFourInfPlausible}{2}
\newcommand{\ExFourInfOther}{0}

\newcommand{\ExFourRectOne}{0}
\newcommand{\ExFourRectTwo}{0}
\newcommand{\ExFourRectThree}{2}
\newcommand{\ExFourRectFour}{2}
\newcommand{\ExFourRectFive}{3}
\newcommand{\ExFourRectSix}{6}
\newcommand{\ExFourRectSeven}{4}

\newcommand{\ExFourHeatOne}{1}
\newcommand{\ExFourHeatTwo}{3}
\newcommand{\ExFourHeatThree}{4}
\newcommand{\ExFourHeatFour}{3}
\newcommand{\ExFourHeatFive}{4}
\newcommand{\ExFourHeatSix}{1}
\newcommand{\ExFourHeatSeven}{1}

\newcommand{\ExFiveSufficientYes}{3}
\newcommand{\ExFiveSufficientNo}{14}

\newcommand{\ExFiveCorrectYes}{0}
\newcommand{\ExFiveCorrectNo}{17}

\newcommand{\ExFiveConfOne}{1}
\newcommand{\ExFiveConfTwo}{0}
\newcommand{\ExFiveConfThree}{0}
\newcommand{\ExFiveConfFour}{0}
\newcommand{\ExFiveConfFive}{3}
\newcommand{\ExFiveConfSix}{4}
\newcommand{\ExFiveConfSeven}{9}

\newcommand{\ExFiveInfAnswerBox}{4}
\newcommand{\ExFiveInfHeatmap}{4}
\newcommand{\ExFiveInfBoth}{3}
\newcommand{\ExFiveInfUnclear}{3}
\newcommand{\ExFiveInfReading}{13}
\newcommand{\ExFiveInfPlausible}{1}
\newcommand{\ExFiveInfOther}{0}

\newcommand{\ExFiveRectOne}{5}
\newcommand{\ExFiveRectTwo}{3}
\newcommand{\ExFiveRectThree}{2}
\newcommand{\ExFiveRectFour}{0}
\newcommand{\ExFiveRectFive}{2}
\newcommand{\ExFiveRectSix}{1}
\newcommand{\ExFiveRectSeven}{4}

\newcommand{\ExFiveHeatOne}{10}
\newcommand{\ExFiveHeatTwo}{5}
\newcommand{\ExFiveHeatThree}{1}
\newcommand{\ExFiveHeatFour}{1}
\newcommand{\ExFiveHeatFive}{0}
\newcommand{\ExFiveHeatSix}{0}
\newcommand{\ExFiveHeatSeven}{0}

\newcommand{\ExSixSufficientYes}{7}
\newcommand{\ExSixSufficientNo}{10}

\newcommand{\ExSixCorrectYes}{3}
\newcommand{\ExSixCorrectNo}{14}

\newcommand{\ExSixConfOne}{0}
\newcommand{\ExSixConfTwo}{0}
\newcommand{\ExSixConfThree}{0}
\newcommand{\ExSixConfFour}{1}
\newcommand{\ExSixConfFive}{1}
\newcommand{\ExSixConfSix}{3}
\newcommand{\ExSixConfSeven}{12}

\newcommand{\ExSixInfAnswerBox}{6}
\newcommand{\ExSixInfHeatmap}{3}
\newcommand{\ExSixInfBoth}{4}
\newcommand{\ExSixInfUnclear}{2}
\newcommand{\ExSixInfReading}{11}
\newcommand{\ExSixInfPlausible}{1}
\newcommand{\ExSixInfOther}{0}

\newcommand{\ExSixRectOne}{11}
\newcommand{\ExSixRectTwo}{2}
\newcommand{\ExSixRectThree}{1}
\newcommand{\ExSixRectFour}{0}
\newcommand{\ExSixRectFive}{0}
\newcommand{\ExSixRectSix}{2}
\newcommand{\ExSixRectSeven}{1}

\newcommand{\ExSixHeatOne}{4}
\newcommand{\ExSixHeatTwo}{6}
\newcommand{\ExSixHeatThree}{2}
\newcommand{\ExSixHeatFour}{1}
\newcommand{\ExSixHeatFive}{1}
\newcommand{\ExSixHeatSix}{2}
\newcommand{\ExSixHeatSeven}{1}

\pgfmathsetmacro{\PartOneGoodTotal}{\BinaryGoodExamples * \UserEvalParticipants}
\pgfmathsetmacro{\PartOneBadTotal}{\BinaryBadExamples * \UserEvalParticipants}

\pgfmathsetmacro{\PartOneGoodSufficientYes}{\ExOneSufficientYes+\ExTwoSufficientYes+\ExThreeSufficientYes+\ExFourSufficientYes}
\pgfmathsetmacro{\PartOneGoodSufficientNo}{\ExOneSufficientNo+\ExTwoSufficientNo+\ExThreeSufficientNo+\ExFourSufficientNo}

\pgfmathsetmacro{\PartOneBadSufficientYes}{\ExFiveSufficientYes+\ExSixSufficientYes}
\pgfmathsetmacro{\PartOneBadSufficientNo}{\ExFiveSufficientNo+\ExSixSufficientNo}

\pgfmathsetmacro{\PartOneGoodCorrectYes}{\ExOneCorrectYes+\ExTwoCorrectYes+\ExThreeCorrectYes+\ExFourCorrectYes}
\pgfmathsetmacro{\PartOneGoodCorrectNo}{\ExOneCorrectNo+\ExTwoCorrectNo+\ExThreeCorrectNo+\ExFourCorrectNo}

\pgfmathsetmacro{\PartOneBadCorrectYes}{\ExFiveCorrectYes+\ExSixCorrectYes}
\pgfmathsetmacro{\PartOneBadCorrectNo}{\ExFiveCorrectNo+\ExSixCorrectNo}

\pgfmathsetmacro{\PartOneGoodConfOne}{\ExOneConfOne+\ExTwoConfOne+\ExThreeConfOne+\ExFourConfOne}
\pgfmathsetmacro{\PartOneGoodConfTwo}{\ExOneConfTwo+\ExTwoConfTwo+\ExThreeConfTwo+\ExFourConfTwo}
\pgfmathsetmacro{\PartOneGoodConfThree}{\ExOneConfThree+\ExTwoConfThree+\ExThreeConfThree+\ExFourConfThree}
\pgfmathsetmacro{\PartOneGoodConfFour}{\ExOneConfFour+\ExTwoConfFour+\ExThreeConfFour+\ExFourConfFour}
\pgfmathsetmacro{\PartOneGoodConfFive}{\ExOneConfFive+\ExTwoConfFive+\ExThreeConfFive+\ExFourConfFive}
\pgfmathsetmacro{\PartOneGoodConfSix}{\ExOneConfSix+\ExTwoConfSix+\ExThreeConfSix+\ExFourConfSix}
\pgfmathsetmacro{\PartOneGoodConfSeven}{\ExOneConfSeven+\ExTwoConfSeven+\ExThreeConfSeven+\ExFourConfSeven}

\pgfmathsetmacro{\PartOneBadConfOne}{\ExFiveConfOne+\ExSixConfOne}
\pgfmathsetmacro{\PartOneBadConfTwo}{\ExFiveConfTwo+\ExSixConfTwo}
\pgfmathsetmacro{\PartOneBadConfThree}{\ExFiveConfThree+\ExSixConfThree}
\pgfmathsetmacro{\PartOneBadConfFour}{\ExFiveConfFour+\ExSixConfFour}
\pgfmathsetmacro{\PartOneBadConfFive}{\ExFiveConfFive+\ExSixConfFive}
\pgfmathsetmacro{\PartOneBadConfSix}{\ExFiveConfSix+\ExSixConfSix}
\pgfmathsetmacro{\PartOneBadConfSeven}{\ExFiveConfSeven+\ExSixConfSeven}

\pgfmathsetmacro{\PartOneGoodInfAnswerBox}{\ExOneInfAnswerBox+\ExTwoInfAnswerBox+\ExThreeInfAnswerBox+\ExFourInfAnswerBox}
\pgfmathsetmacro{\PartOneGoodInfHeatmap}{\ExOneInfHeatmap+\ExTwoInfHeatmap+\ExThreeInfHeatmap+\ExFourInfHeatmap}
\pgfmathsetmacro{\PartOneGoodInfBoth}{\ExOneInfBoth+\ExTwoInfBoth+\ExThreeInfBoth+\ExFourInfBoth}
\pgfmathsetmacro{\PartOneGoodInfUnclear}{\ExOneInfUnclear+\ExTwoInfUnclear+\ExThreeInfUnclear+\ExFourInfUnclear}
\pgfmathsetmacro{\PartOneGoodInfReading}{\ExOneInfReading+\ExTwoInfReading+\ExThreeInfReading+\ExFourInfReading}
\pgfmathsetmacro{\PartOneGoodInfPlausible}{\ExOneInfPlausible+\ExTwoInfPlausible+\ExThreeInfPlausible+\ExFourInfPlausible}
\pgfmathsetmacro{\PartOneGoodInfOther}{\ExOneInfOther+\ExTwoInfOther+\ExThreeInfOther+\ExFourInfOther}

\pgfmathsetmacro{\PartOneBadInfAnswerBox}{\ExFiveInfAnswerBox+\ExSixInfAnswerBox}
\pgfmathsetmacro{\PartOneBadInfHeatmap}{\ExFiveInfHeatmap+\ExSixInfHeatmap}
\pgfmathsetmacro{\PartOneBadInfBoth}{\ExFiveInfBoth+\ExSixInfBoth}
\pgfmathsetmacro{\PartOneBadInfUnclear}{\ExFiveInfUnclear+\ExSixInfUnclear}
\pgfmathsetmacro{\PartOneBadInfReading}{\ExFiveInfReading+\ExSixInfReading}
\pgfmathsetmacro{\PartOneBadInfPlausible}{\ExFiveInfPlausible+\ExSixInfPlausible}
\pgfmathsetmacro{\PartOneBadInfOther}{\ExFiveInfOther+\ExSixInfOther}

\pgfmathsetmacro{\PartOneGoodRectOne}{\ExOneRectOne+\ExTwoRectOne+\ExThreeRectOne+\ExFourRectOne}
\pgfmathsetmacro{\PartOneGoodRectTwo}{\ExOneRectTwo+\ExTwoRectTwo+\ExThreeRectTwo+\ExFourRectTwo}
\pgfmathsetmacro{\PartOneGoodRectThree}{\ExOneRectThree+\ExTwoRectThree+\ExThreeRectThree+\ExFourRectThree}
\pgfmathsetmacro{\PartOneGoodRectFour}{\ExOneRectFour+\ExTwoRectFour+\ExThreeRectFour+\ExFourRectFour}
\pgfmathsetmacro{\PartOneGoodRectFive}{\ExOneRectFive+\ExTwoRectFive+\ExThreeRectFive+\ExFourRectFive}
\pgfmathsetmacro{\PartOneGoodRectSix}{\ExOneRectSix+\ExTwoRectSix+\ExThreeRectSix+\ExFourRectSix}
\pgfmathsetmacro{\PartOneGoodRectSeven}{\ExOneRectSeven+\ExTwoRectSeven+\ExThreeRectSeven+\ExFourRectSeven}

\pgfmathsetmacro{\PartOneBadRectOne}{\ExFiveRectOne+\ExSixRectOne}
\pgfmathsetmacro{\PartOneBadRectTwo}{\ExFiveRectTwo+\ExSixRectTwo}
\pgfmathsetmacro{\PartOneBadRectThree}{\ExFiveRectThree+\ExSixRectThree}
\pgfmathsetmacro{\PartOneBadRectFour}{\ExFiveRectFour+\ExSixRectFour}
\pgfmathsetmacro{\PartOneBadRectFive}{\ExFiveRectFive+\ExSixRectFive}
\pgfmathsetmacro{\PartOneBadRectSix}{\ExFiveRectSix+\ExSixRectSix}
\pgfmathsetmacro{\PartOneBadRectSeven}{\ExFiveRectSeven+\ExSixRectSeven}

\pgfmathsetmacro{\PartOneGoodHeatOne}{\ExOneHeatOne+\ExTwoHeatOne+\ExThreeHeatOne+\ExFourHeatOne}
\pgfmathsetmacro{\PartOneGoodHeatTwo}{\ExOneHeatTwo+\ExTwoHeatTwo+\ExThreeHeatTwo+\ExFourHeatTwo}
\pgfmathsetmacro{\PartOneGoodHeatThree}{\ExOneHeatThree+\ExTwoHeatThree+\ExThreeHeatThree+\ExFourHeatThree}
\pgfmathsetmacro{\PartOneGoodHeatFour}{\ExOneHeatFour+\ExTwoHeatFour+\ExThreeHeatFour+\ExFourHeatFour}
\pgfmathsetmacro{\PartOneGoodHeatFive}{\ExOneHeatFive+\ExTwoHeatFive+\ExThreeHeatFive+\ExFourHeatFive}
\pgfmathsetmacro{\PartOneGoodHeatSix}{\ExOneHeatSix+\ExTwoHeatSix+\ExThreeHeatSix+\ExFourHeatSix}
\pgfmathsetmacro{\PartOneGoodHeatSeven}{\ExOneHeatSeven+\ExTwoHeatSeven+\ExThreeHeatSeven+\ExFourHeatSeven}

\pgfmathsetmacro{\PartOneBadHeatOne}{\ExFiveHeatOne+\ExSixHeatOne}
\pgfmathsetmacro{\PartOneBadHeatTwo}{\ExFiveHeatTwo+\ExSixHeatTwo}
\pgfmathsetmacro{\PartOneBadHeatThree}{\ExFiveHeatThree+\ExSixHeatThree}
\pgfmathsetmacro{\PartOneBadHeatFour}{\ExFiveHeatFour+\ExSixHeatFour}
\pgfmathsetmacro{\PartOneBadHeatFive}{\ExFiveHeatFive+\ExSixHeatFive}
\pgfmathsetmacro{\PartOneBadHeatSix}{\ExFiveHeatSix+\ExSixHeatSix}
\pgfmathsetmacro{\PartOneBadHeatSeven}{\ExFiveHeatSeven+\ExSixHeatSeven}

\newcommand{\ComputeLikertMean}[7]{%
    (#1*1 + #2*2 + #3*3 + #4*4 + #5*5 + #6*6 + #7*7) / (#1 + #2 + #3 + #4 + #5 + #6 + #7)%
}

\newcommand{\ComputeLikertStd}[7]{%
    sqrt(%
        (#1*1*1 + #2*2*2 + #3*3*3 + #4*4*4 + #5*5*5 + #6*6*6 + #7*7*7) / (#1 + #2 + #3 + #4 + #5 + #6 + #7)%
        - (\ComputeLikertMean{#1}{#2}{#3}{#4}{#5}{#6}{#7})^2%
    )%
}

\newcommand{\ComputeBinaryMean}[2]{%
    #1 / (#1 + #2)%
}

\newcommand{\ComputeBinaryStd}[2]{%
    sqrt((\ComputeBinaryMean{#1}{#2}) * (1 - \ComputeBinaryMean{#1}{#2}))%
}

\pgfmathsetmacro{\ExOneSufficientMean}{\ComputeBinaryMean{\ExOneSufficientYes}{\ExOneSufficientNo}}
\pgfmathsetmacro{\ExOneSufficientStd}{\ComputeBinaryStd{\ExOneSufficientYes}{\ExOneSufficientNo}}
\pgfmathsetmacro{\ExOneCorrectMean}{\ComputeBinaryMean{\ExOneCorrectYes}{\ExOneCorrectNo}}
\pgfmathsetmacro{\ExOneCorrectStd}{\ComputeBinaryStd{\ExOneCorrectYes}{\ExOneCorrectNo}}

\pgfmathsetmacro{\ExTwoSufficientMean}{\ComputeBinaryMean{\ExTwoSufficientYes}{\ExTwoSufficientNo}}
\pgfmathsetmacro{\ExTwoSufficientStd}{\ComputeBinaryStd{\ExTwoSufficientYes}{\ExTwoSufficientNo}}
\pgfmathsetmacro{\ExTwoCorrectMean}{\ComputeBinaryMean{\ExTwoCorrectYes}{\ExTwoCorrectNo}}
\pgfmathsetmacro{\ExTwoCorrectStd}{\ComputeBinaryStd{\ExTwoCorrectYes}{\ExTwoCorrectNo}}

\pgfmathsetmacro{\ExThreeSufficientMean}{\ComputeBinaryMean{\ExThreeSufficientYes}{\ExThreeSufficientNo}}
\pgfmathsetmacro{\ExThreeSufficientStd}{\ComputeBinaryStd{\ExThreeSufficientYes}{\ExThreeSufficientNo}}
\pgfmathsetmacro{\ExThreeCorrectMean}{\ComputeBinaryMean{\ExThreeCorrectYes}{\ExThreeCorrectNo}}
\pgfmathsetmacro{\ExThreeCorrectStd}{\ComputeBinaryStd{\ExThreeCorrectYes}{\ExThreeCorrectNo}}

\pgfmathsetmacro{\ExFourSufficientMean}{\ComputeBinaryMean{\ExFourSufficientYes}{\ExFourSufficientNo}}
\pgfmathsetmacro{\ExFourSufficientStd}{\ComputeBinaryStd{\ExFourSufficientYes}{\ExFourSufficientNo}}
\pgfmathsetmacro{\ExFourCorrectMean}{\ComputeBinaryMean{\ExFourCorrectYes}{\ExFourCorrectNo}}
\pgfmathsetmacro{\ExFourCorrectStd}{\ComputeBinaryStd{\ExFourCorrectYes}{\ExFourCorrectNo}}

\pgfmathsetmacro{\ExFiveSufficientMean}{\ComputeBinaryMean{\ExFiveSufficientYes}{\ExFiveSufficientNo}}
\pgfmathsetmacro{\ExFiveSufficientStd}{\ComputeBinaryStd{\ExFiveSufficientYes}{\ExFiveSufficientNo}}
\pgfmathsetmacro{\ExFiveCorrectMean}{\ComputeBinaryMean{\ExFiveCorrectYes}{\ExFiveCorrectNo}}
\pgfmathsetmacro{\ExFiveCorrectStd}{\ComputeBinaryStd{\ExFiveCorrectYes}{\ExFiveCorrectNo}}

\pgfmathsetmacro{\ExSixSufficientMean}{\ComputeBinaryMean{\ExSixSufficientYes}{\ExSixSufficientNo}}
\pgfmathsetmacro{\ExSixSufficientStd}{\ComputeBinaryStd{\ExSixSufficientYes}{\ExSixSufficientNo}}
\pgfmathsetmacro{\ExSixCorrectMean}{\ComputeBinaryMean{\ExSixCorrectYes}{\ExSixCorrectNo}}
\pgfmathsetmacro{\ExSixCorrectStd}{\ComputeBinaryStd{\ExSixCorrectYes}{\ExSixCorrectNo}}

\pgfmathsetmacro{\ExOneConfMean}{\ComputeLikertMean{\ExOneConfOne}{\ExOneConfTwo}{\ExOneConfThree}{\ExOneConfFour}{\ExOneConfFive}{\ExOneConfSix}{\ExOneConfSeven}}
\pgfmathsetmacro{\ExOneConfStd}{\ComputeLikertStd{\ExOneConfOne}{\ExOneConfTwo}{\ExOneConfThree}{\ExOneConfFour}{\ExOneConfFive}{\ExOneConfSix}{\ExOneConfSeven}}

\pgfmathsetmacro{\ExTwoConfMean}{\ComputeLikertMean{\ExTwoConfOne}{\ExTwoConfTwo}{\ExTwoConfThree}{\ExTwoConfFour}{\ExTwoConfFive}{\ExTwoConfSix}{\ExTwoConfSeven}}
\pgfmathsetmacro{\ExTwoConfStd}{\ComputeLikertStd{\ExTwoConfOne}{\ExTwoConfTwo}{\ExTwoConfThree}{\ExTwoConfFour}{\ExTwoConfFive}{\ExTwoConfSix}{\ExTwoConfSeven}}

\pgfmathsetmacro{\ExThreeConfMean}{\ComputeLikertMean{\ExThreeConfOne}{\ExThreeConfTwo}{\ExThreeConfThree}{\ExThreeConfFour}{\ExThreeConfFive}{\ExThreeConfSix}{\ExThreeConfSeven}}
\pgfmathsetmacro{\ExThreeConfStd}{\ComputeLikertStd{\ExThreeConfOne}{\ExThreeConfTwo}{\ExThreeConfThree}{\ExThreeConfFour}{\ExThreeConfFive}{\ExThreeConfSix}{\ExThreeConfSeven}}

\pgfmathsetmacro{\ExFourConfMean}{\ComputeLikertMean{\ExFourConfOne}{\ExFourConfTwo}{\ExFourConfThree}{\ExFourConfFour}{\ExFourConfFive}{\ExFourConfSix}{\ExFourConfSeven}}
\pgfmathsetmacro{\ExFourConfStd}{\ComputeLikertStd{\ExFourConfOne}{\ExFourConfTwo}{\ExFourConfThree}{\ExFourConfFour}{\ExFourConfFive}{\ExFourConfSix}{\ExFourConfSeven}}

\pgfmathsetmacro{\ExFiveConfMean}{\ComputeLikertMean{\ExFiveConfOne}{\ExFiveConfTwo}{\ExFiveConfThree}{\ExFiveConfFour}{\ExFiveConfFive}{\ExFiveConfSix}{\ExFiveConfSeven}}
\pgfmathsetmacro{\ExFiveConfStd}{\ComputeLikertStd{\ExFiveConfOne}{\ExFiveConfTwo}{\ExFiveConfThree}{\ExFiveConfFour}{\ExFiveConfFive}{\ExFiveConfSix}{\ExFiveConfSeven}}

\pgfmathsetmacro{\ExSixConfMean}{\ComputeLikertMean{\ExSixConfOne}{\ExSixConfTwo}{\ExSixConfThree}{\ExSixConfFour}{\ExSixConfFive}{\ExSixConfSix}{\ExSixConfSeven}}
\pgfmathsetmacro{\ExSixConfStd}{\ComputeLikertStd{\ExSixConfOne}{\ExSixConfTwo}{\ExSixConfThree}{\ExSixConfFour}{\ExSixConfFive}{\ExSixConfSix}{\ExSixConfSeven}}

\pgfmathsetmacro{\ExOneRectMean}{\ComputeLikertMean{\ExOneRectOne}{\ExOneRectTwo}{\ExOneRectThree}{\ExOneRectFour}{\ExOneRectFive}{\ExOneRectSix}{\ExOneRectSeven}}
\pgfmathsetmacro{\ExOneRectStd}{\ComputeLikertStd{\ExOneRectOne}{\ExOneRectTwo}{\ExOneRectThree}{\ExOneRectFour}{\ExOneRectFive}{\ExOneRectSix}{\ExOneRectSeven}}

\pgfmathsetmacro{\ExTwoRectMean}{\ComputeLikertMean{\ExTwoRectOne}{\ExTwoRectTwo}{\ExTwoRectThree}{\ExTwoRectFour}{\ExTwoRectFive}{\ExTwoRectSix}{\ExTwoRectSeven}}
\pgfmathsetmacro{\ExTwoRectStd}{\ComputeLikertStd{\ExTwoRectOne}{\ExTwoRectTwo}{\ExTwoRectThree}{\ExTwoRectFour}{\ExTwoRectFive}{\ExTwoRectSix}{\ExTwoRectSeven}}

\pgfmathsetmacro{\ExThreeRectMean}{\ComputeLikertMean{\ExThreeRectOne}{\ExThreeRectTwo}{\ExThreeRectThree}{\ExThreeRectFour}{\ExThreeRectFive}{\ExThreeRectSix}{\ExThreeRectSeven}}
\pgfmathsetmacro{\ExThreeRectStd}{\ComputeLikertStd{\ExThreeRectOne}{\ExThreeRectTwo}{\ExThreeRectThree}{\ExThreeRectFour}{\ExThreeRectFive}{\ExThreeRectSix}{\ExThreeRectSeven}}

\pgfmathsetmacro{\ExFourRectMean}{\ComputeLikertMean{\ExFourRectOne}{\ExFourRectTwo}{\ExFourRectThree}{\ExFourRectFour}{\ExFourRectFive}{\ExFourRectSix}{\ExFourRectSeven}}
\pgfmathsetmacro{\ExFourRectStd}{\ComputeLikertStd{\ExFourRectOne}{\ExFourRectTwo}{\ExFourRectThree}{\ExFourRectFour}{\ExFourRectFive}{\ExFourRectSix}{\ExFourRectSeven}}

\pgfmathsetmacro{\ExFiveRectMean}{\ComputeLikertMean{\ExFiveRectOne}{\ExFiveRectTwo}{\ExFiveRectThree}{\ExFiveRectFour}{\ExFiveRectFive}{\ExFiveRectSix}{\ExFiveRectSeven}}
\pgfmathsetmacro{\ExFiveRectStd}{\ComputeLikertStd{\ExFiveRectOne}{\ExFiveRectTwo}{\ExFiveRectThree}{\ExFiveRectFour}{\ExFiveRectFive}{\ExFiveRectSix}{\ExFiveRectSeven}}

\pgfmathsetmacro{\ExSixRectMean}{\ComputeLikertMean{\ExSixRectOne}{\ExSixRectTwo}{\ExSixRectThree}{\ExSixRectFour}{\ExSixRectFive}{\ExSixRectSix}{\ExSixRectSeven}}
\pgfmathsetmacro{\ExSixRectStd}{\ComputeLikertStd{\ExSixRectOne}{\ExSixRectTwo}{\ExSixRectThree}{\ExSixRectFour}{\ExSixRectFive}{\ExSixRectSix}{\ExSixRectSeven}}

\pgfmathsetmacro{\ExOneHeatMean}{\ComputeLikertMean{\ExOneHeatOne}{\ExOneHeatTwo}{\ExOneHeatThree}{\ExOneHeatFour}{\ExOneHeatFive}{\ExOneHeatSix}{\ExOneHeatSeven}}
\pgfmathsetmacro{\ExOneHeatStd}{\ComputeLikertStd{\ExOneHeatOne}{\ExOneHeatTwo}{\ExOneHeatThree}{\ExOneHeatFour}{\ExOneHeatFive}{\ExOneHeatSix}{\ExOneHeatSeven}}

\pgfmathsetmacro{\ExTwoHeatMean}{\ComputeLikertMean{\ExTwoHeatOne}{\ExTwoHeatTwo}{\ExTwoHeatThree}{\ExTwoHeatFour}{\ExTwoHeatFive}{\ExTwoHeatSix}{\ExTwoHeatSeven}}
\pgfmathsetmacro{\ExTwoHeatStd}{\ComputeLikertStd{\ExTwoHeatOne}{\ExTwoHeatTwo}{\ExTwoHeatThree}{\ExTwoHeatFour}{\ExTwoHeatFive}{\ExTwoHeatSix}{\ExTwoHeatSeven}}

\pgfmathsetmacro{\ExThreeHeatMean}{\ComputeLikertMean{\ExThreeHeatOne}{\ExThreeHeatTwo}{\ExThreeHeatThree}{\ExThreeHeatFour}{\ExThreeHeatFive}{\ExThreeHeatSix}{\ExThreeHeatSeven}}
\pgfmathsetmacro{\ExThreeHeatStd}{\ComputeLikertStd{\ExThreeHeatOne}{\ExThreeHeatTwo}{\ExThreeHeatThree}{\ExThreeHeatFour}{\ExThreeHeatFive}{\ExThreeHeatSix}{\ExThreeHeatSeven}}

\pgfmathsetmacro{\ExFourHeatMean}{\ComputeLikertMean{\ExFourHeatOne}{\ExFourHeatTwo}{\ExFourHeatThree}{\ExFourHeatFour}{\ExFourHeatFive}{\ExFourHeatSix}{\ExFourHeatSeven}}
\pgfmathsetmacro{\ExFourHeatStd}{\ComputeLikertStd{\ExFourHeatOne}{\ExFourHeatTwo}{\ExFourHeatThree}{\ExFourHeatFour}{\ExFourHeatFive}{\ExFourHeatSix}{\ExFourHeatSeven}}

\pgfmathsetmacro{\ExFiveHeatMean}{\ComputeLikertMean{\ExFiveHeatOne}{\ExFiveHeatTwo}{\ExFiveHeatThree}{\ExFiveHeatFour}{\ExFiveHeatFive}{\ExFiveHeatSix}{\ExFiveHeatSeven}}
\pgfmathsetmacro{\ExFiveHeatStd}{\ComputeLikertStd{\ExFiveHeatOne}{\ExFiveHeatTwo}{\ExFiveHeatThree}{\ExFiveHeatFour}{\ExFiveHeatFive}{\ExFiveHeatSix}{\ExFiveHeatSeven}}

\pgfmathsetmacro{\ExSixHeatMean}{\ComputeLikertMean{\ExSixHeatOne}{\ExSixHeatTwo}{\ExSixHeatThree}{\ExSixHeatFour}{\ExSixHeatFive}{\ExSixHeatSix}{\ExSixHeatSeven}}
\pgfmathsetmacro{\ExSixHeatStd}{\ComputeLikertStd{\ExSixHeatOne}{\ExSixHeatTwo}{\ExSixHeatThree}{\ExSixHeatFour}{\ExSixHeatFive}{\ExSixHeatSix}{\ExSixHeatSeven}}

\pgfmathsetmacro{\GoodSufficientYes}{\ExOneSufficientYes+\ExTwoSufficientYes+\ExThreeSufficientYes+\ExFourSufficientYes}
\pgfmathsetmacro{\GoodSufficientNo}{\ExOneSufficientNo+\ExTwoSufficientNo+\ExThreeSufficientNo+\ExFourSufficientNo}
\pgfmathsetmacro{\GoodSufficientMean}{\ComputeBinaryMean{\GoodSufficientYes}{\GoodSufficientNo}}
\pgfmathsetmacro{\GoodSufficientStd}{\ComputeBinaryStd{\GoodSufficientYes}{\GoodSufficientNo}}

\pgfmathsetmacro{\GoodCorrectYes}{\ExOneCorrectYes+\ExTwoCorrectYes+\ExThreeCorrectYes+\ExFourCorrectYes}
\pgfmathsetmacro{\GoodCorrectNo}{\ExOneCorrectNo+\ExTwoCorrectNo+\ExThreeCorrectNo+\ExFourCorrectNo}
\pgfmathsetmacro{\GoodCorrectMean}{\ComputeBinaryMean{\GoodCorrectYes}{\GoodCorrectNo}}
\pgfmathsetmacro{\GoodCorrectStd}{\ComputeBinaryStd{\GoodCorrectYes}{\GoodCorrectNo}}

\pgfmathsetmacro{\GoodConfOne}{\ExOneConfOne+\ExTwoConfOne+\ExThreeConfOne+\ExFourConfOne}
\pgfmathsetmacro{\GoodConfTwo}{\ExOneConfTwo+\ExTwoConfTwo+\ExThreeConfTwo+\ExFourConfTwo}
\pgfmathsetmacro{\GoodConfThree}{\ExOneConfThree+\ExTwoConfThree+\ExThreeConfThree+\ExFourConfThree}
\pgfmathsetmacro{\GoodConfFour}{\ExOneConfFour+\ExTwoConfFour+\ExThreeConfFour+\ExFourConfFour}
\pgfmathsetmacro{\GoodConfFive}{\ExOneConfFive+\ExTwoConfFive+\ExThreeConfFive+\ExFourConfFive}
\pgfmathsetmacro{\GoodConfSix}{\ExOneConfSix+\ExTwoConfSix+\ExThreeConfSix+\ExFourConfSix}
\pgfmathsetmacro{\GoodConfSeven}{\ExOneConfSeven+\ExTwoConfSeven+\ExThreeConfSeven+\ExFourConfSeven}
\pgfmathsetmacro{\GoodConfMean}{\ComputeLikertMean{\GoodConfOne}{\GoodConfTwo}{\GoodConfThree}{\GoodConfFour}{\GoodConfFive}{\GoodConfSix}{\GoodConfSeven}}
\pgfmathsetmacro{\GoodConfStd}{\ComputeLikertStd{\GoodConfOne}{\GoodConfTwo}{\GoodConfThree}{\GoodConfFour}{\GoodConfFive}{\GoodConfSix}{\GoodConfSeven}}

\pgfmathsetmacro{\GoodRectOne}{\ExOneRectOne+\ExTwoRectOne+\ExThreeRectOne+\ExFourRectOne}
\pgfmathsetmacro{\GoodRectTwo}{\ExOneRectTwo+\ExTwoRectTwo+\ExThreeRectTwo+\ExFourRectTwo}
\pgfmathsetmacro{\GoodRectThree}{\ExOneRectThree+\ExTwoRectThree+\ExThreeRectThree+\ExFourRectThree}
\pgfmathsetmacro{\GoodRectFour}{\ExOneRectFour+\ExTwoRectFour+\ExThreeRectFour+\ExFourRectFour}
\pgfmathsetmacro{\GoodRectFive}{\ExOneRectFive+\ExTwoRectFive+\ExThreeRectFive+\ExFourRectFive}
\pgfmathsetmacro{\GoodRectSix}{\ExOneRectSix+\ExTwoRectSix+\ExThreeRectSix+\ExFourRectSix}
\pgfmathsetmacro{\GoodRectSeven}{\ExOneRectSeven+\ExTwoRectSeven+\ExThreeRectSeven+\ExFourRectSeven}
\pgfmathsetmacro{\GoodRectMean}{\ComputeLikertMean{\GoodRectOne}{\GoodRectTwo}{\GoodRectThree}{\GoodRectFour}{\GoodRectFive}{\GoodRectSix}{\GoodRectSeven}}
\pgfmathsetmacro{\GoodRectStd}{\ComputeLikertStd{\GoodRectOne}{\GoodRectTwo}{\GoodRectThree}{\GoodRectFour}{\GoodRectFive}{\GoodRectSix}{\GoodRectSeven}}

\pgfmathsetmacro{\GoodHeatOne}{\ExOneHeatOne+\ExTwoHeatOne+\ExThreeHeatOne+\ExFourHeatOne}
\pgfmathsetmacro{\GoodHeatTwo}{\ExOneHeatTwo+\ExTwoHeatTwo+\ExThreeHeatTwo+\ExFourHeatTwo}
\pgfmathsetmacro{\GoodHeatThree}{\ExOneHeatThree+\ExTwoHeatThree+\ExThreeHeatThree+\ExFourHeatThree}
\pgfmathsetmacro{\GoodHeatFour}{\ExOneHeatFour+\ExTwoHeatFour+\ExThreeHeatFour+\ExFourHeatFour}
\pgfmathsetmacro{\GoodHeatFive}{\ExOneHeatFive+\ExTwoHeatFive+\ExThreeHeatFive+\ExFourHeatFive}
\pgfmathsetmacro{\GoodHeatSix}{\ExOneHeatSix+\ExTwoHeatSix+\ExThreeHeatSix+\ExFourHeatSix}
\pgfmathsetmacro{\GoodHeatSeven}{\ExOneHeatSeven+\ExTwoHeatSeven+\ExThreeHeatSeven+\ExFourHeatSeven}
\pgfmathsetmacro{\GoodHeatMean}{\ComputeLikertMean{\GoodHeatOne}{\GoodHeatTwo}{\GoodHeatThree}{\GoodHeatFour}{\GoodHeatFive}{\GoodHeatSix}{\GoodHeatSeven}}
\pgfmathsetmacro{\GoodHeatStd}{\ComputeLikertStd{\GoodHeatOne}{\GoodHeatTwo}{\GoodHeatThree}{\GoodHeatFour}{\GoodHeatFive}{\GoodHeatSix}{\GoodHeatSeven}}

\pgfmathsetmacro{\BadSufficientYes}{\ExFiveSufficientYes+\ExSixSufficientYes}
\pgfmathsetmacro{\BadSufficientNo}{\ExFiveSufficientNo+\ExSixSufficientNo}
\pgfmathsetmacro{\BadSufficientMean}{\ComputeBinaryMean{\BadSufficientYes}{\BadSufficientNo}}
\pgfmathsetmacro{\BadSufficientStd}{\ComputeBinaryStd{\BadSufficientYes}{\BadSufficientNo}}

\pgfmathsetmacro{\BadCorrectYes}{\ExFiveCorrectYes+\ExSixCorrectYes}
\pgfmathsetmacro{\BadCorrectNo}{\ExFiveCorrectNo+\ExSixCorrectNo}
\pgfmathsetmacro{\BadCorrectMean}{\ComputeBinaryMean{\BadCorrectYes}{\BadCorrectNo}}
\pgfmathsetmacro{\BadCorrectStd}{\ComputeBinaryStd{\BadCorrectYes}{\BadCorrectNo}}

\pgfmathsetmacro{\BadConfOne}{\ExFiveConfOne+\ExSixConfOne}
\pgfmathsetmacro{\BadConfTwo}{\ExFiveConfTwo+\ExSixConfTwo}
\pgfmathsetmacro{\BadConfThree}{\ExFiveConfThree+\ExSixConfThree}
\pgfmathsetmacro{\BadConfFour}{\ExFiveConfFour+\ExSixConfFour}
\pgfmathsetmacro{\BadConfFive}{\ExFiveConfFive+\ExSixConfFive}
\pgfmathsetmacro{\BadConfSix}{\ExFiveConfSix+\ExSixConfSix}
\pgfmathsetmacro{\BadConfSeven}{\ExFiveConfSeven+\ExSixConfSeven}
\pgfmathsetmacro{\BadConfMean}{\ComputeLikertMean{\BadConfOne}{\BadConfTwo}{\BadConfThree}{\BadConfFour}{\BadConfFive}{\BadConfSix}{\BadConfSeven}}
\pgfmathsetmacro{\BadConfStd}{\ComputeLikertStd{\BadConfOne}{\BadConfTwo}{\BadConfThree}{\BadConfFour}{\BadConfFive}{\BadConfSix}{\BadConfSeven}}

\pgfmathsetmacro{\BadRectOne}{\ExFiveRectOne+\ExSixRectOne}
\pgfmathsetmacro{\BadRectTwo}{\ExFiveRectTwo+\ExSixRectTwo}
\pgfmathsetmacro{\BadRectThree}{\ExFiveRectThree+\ExSixRectThree}
\pgfmathsetmacro{\BadRectFour}{\ExFiveRectFour+\ExSixRectFour}
\pgfmathsetmacro{\BadRectFive}{\ExFiveRectFive+\ExSixRectFive}
\pgfmathsetmacro{\BadRectSix}{\ExFiveRectSix+\ExSixRectSix}
\pgfmathsetmacro{\BadRectSeven}{\ExFiveRectSeven+\ExSixRectSeven}
\pgfmathsetmacro{\BadRectMean}{\ComputeLikertMean{\BadRectOne}{\BadRectTwo}{\BadRectThree}{\BadRectFour}{\BadRectFive}{\BadRectSix}{\BadRectSeven}}
\pgfmathsetmacro{\BadRectStd}{\ComputeLikertStd{\BadRectOne}{\BadRectTwo}{\BadRectThree}{\BadRectFour}{\BadRectFive}{\BadRectSix}{\BadRectSeven}}

\pgfmathsetmacro{\BadHeatOne}{\ExFiveHeatOne+\ExSixHeatOne}
\pgfmathsetmacro{\BadHeatTwo}{\ExFiveHeatTwo+\ExSixHeatTwo}
\pgfmathsetmacro{\BadHeatThree}{\ExFiveHeatThree+\ExSixHeatThree}
\pgfmathsetmacro{\BadHeatFour}{\ExFiveHeatFour+\ExSixHeatFour}
\pgfmathsetmacro{\BadHeatFive}{\ExFiveHeatFive+\ExSixHeatFive}
\pgfmathsetmacro{\BadHeatSix}{\ExFiveHeatSix+\ExSixHeatSix}
\pgfmathsetmacro{\BadHeatSeven}{\ExFiveHeatSeven+\ExSixHeatSeven}
\pgfmathsetmacro{\BadHeatMean}{\ComputeLikertMean{\BadHeatOne}{\BadHeatTwo}{\BadHeatThree}{\BadHeatFour}{\BadHeatFive}{\BadHeatSix}{\BadHeatSeven}}
\pgfmathsetmacro{\BadHeatStd}{\ComputeLikertStd{\BadHeatOne}{\BadHeatTwo}{\BadHeatThree}{\BadHeatFour}{\BadHeatFive}{\BadHeatSix}{\BadHeatSeven}}

\newcommand{\UserAnswerGoodExamples}{4} 
\newcommand{\UserAnswerBadExamples}{2}
\newcommand{\UserAnswerExamples}{6}

\newcommand{\ExOneUserCorrect}{15}
\newcommand{\ExOneUserIncorrect}{2}

\newcommand{\ExOneUserConfOne}{0}
\newcommand{\ExOneUserConfTwo}{0}
\newcommand{\ExOneUserConfThree}{0}
\newcommand{\ExOneUserConfFour}{1}
\newcommand{\ExOneUserConfFive}{3}
\newcommand{\ExOneUserConfSix}{1}
\newcommand{\ExOneUserConfSeven}{12}

\newcommand{\ExTwoUserCorrect}{14}
\newcommand{\ExTwoUserIncorrect}{3}

\newcommand{\ExTwoUserConfOne}{0}
\newcommand{\ExTwoUserConfTwo}{0}
\newcommand{\ExTwoUserConfThree}{0}
\newcommand{\ExTwoUserConfFour}{0}
\newcommand{\ExTwoUserConfFive}{3}
\newcommand{\ExTwoUserConfSix}{3}
\newcommand{\ExTwoUserConfSeven}{11}

\newcommand{\ExThreeUserCorrect}{15}
\newcommand{\ExThreeUserIncorrect}{2}

\newcommand{\ExThreeUserConfOne}{0}
\newcommand{\ExThreeUserConfTwo}{0}
\newcommand{\ExThreeUserConfThree}{2}
\newcommand{\ExThreeUserConfFour}{0}
\newcommand{\ExThreeUserConfFive}{3}
\newcommand{\ExThreeUserConfSix}{1}
\newcommand{\ExThreeUserConfSeven}{11}

\newcommand{\ExFourUserCorrect}{16}
\newcommand{\ExFourUserIncorrect}{1}

\newcommand{\ExFourUserConfOne}{0}
\newcommand{\ExFourUserConfTwo}{0}
\newcommand{\ExFourUserConfThree}{1}
\newcommand{\ExFourUserConfFour}{0}
\newcommand{\ExFourUserConfFive}{0}
\newcommand{\ExFourUserConfSix}{3}
\newcommand{\ExFourUserConfSeven}{13}

\newcommand{\ExFiveUserCorrect}{2}
\newcommand{\ExFiveUserIncorrect}{15}

\newcommand{\ExFiveUserConfOne}{8}
\newcommand{\ExFiveUserConfTwo}{1}
\newcommand{\ExFiveUserConfThree}{1}
\newcommand{\ExFiveUserConfFour}{3}
\newcommand{\ExFiveUserConfFive}{0}
\newcommand{\ExFiveUserConfSix}{0}
\newcommand{\ExFiveUserConfSeven}{4}

\newcommand{\ExSixUserCorrect}{3}
\newcommand{\ExSixUserIncorrect}{14}

\newcommand{\ExSixUserConfOne}{7}
\newcommand{\ExSixUserConfTwo}{1}
\newcommand{\ExSixUserConfThree}{2}
\newcommand{\ExSixUserConfFour}{0}
\newcommand{\ExSixUserConfFive}{1}
\newcommand{\ExSixUserConfSix}{1}
\newcommand{\ExSixUserConfSeven}{5}

\pgfmathsetmacro{\PartTwoGoodTotal}{\UserAnswerGoodExamples * \UserEvalParticipants}
\pgfmathsetmacro{\PartTwoBadTotal}{\UserAnswerBadExamples * \UserEvalParticipants}

\pgfmathsetmacro{\PartTwoGoodCorrect}{\ExOneUserCorrect+\ExTwoUserCorrect+\ExThreeUserCorrect+\ExFourUserCorrect}
\pgfmathsetmacro{\PartTwoGoodIncorrect}{\ExOneUserIncorrect+\ExTwoUserIncorrect+\ExThreeUserIncorrect+\ExFourUserIncorrect}

\pgfmathsetmacro{\PartTwoBadCorrect}{\ExFiveUserCorrect+\ExSixUserCorrect}
\pgfmathsetmacro{\PartTwoBadIncorrect}{\ExFiveUserIncorrect+\ExSixUserIncorrect}

\pgfmathsetmacro{\PartTwoGoodConfOne}{\ExOneUserConfOne+\ExTwoUserConfOne+\ExThreeUserConfOne+\ExFourUserConfOne}
\pgfmathsetmacro{\PartTwoGoodConfTwo}{\ExOneUserConfTwo+\ExTwoUserConfTwo+\ExThreeUserConfTwo+\ExFourUserConfTwo}
\pgfmathsetmacro{\PartTwoGoodConfThree}{\ExOneUserConfThree+\ExTwoUserConfThree+\ExThreeUserConfThree+\ExFourUserConfThree}
\pgfmathsetmacro{\PartTwoGoodConfFour}{\ExOneUserConfFour+\ExTwoUserConfFour+\ExThreeUserConfFour+\ExFourUserConfFour}
\pgfmathsetmacro{\PartTwoGoodConfFive}{\ExOneUserConfFive+\ExTwoUserConfFive+\ExThreeUserConfFive+\ExFourUserConfFive}
\pgfmathsetmacro{\PartTwoGoodConfSix}{\ExOneUserConfSix+\ExTwoUserConfSix+\ExThreeUserConfSix+\ExFourUserConfSix}
\pgfmathsetmacro{\PartTwoGoodConfSeven}{\ExOneUserConfSeven+\ExTwoUserConfSeven+\ExThreeUserConfSeven+\ExFourUserConfSeven}

\pgfmathsetmacro{\PartTwoBadConfOne}{\ExFiveUserConfOne+\ExSixUserConfOne}
\pgfmathsetmacro{\PartTwoBadConfTwo}{\ExFiveUserConfTwo+\ExSixUserConfTwo}
\pgfmathsetmacro{\PartTwoBadConfThree}{\ExFiveUserConfThree+\ExSixUserConfThree}
\pgfmathsetmacro{\PartTwoBadConfFour}{\ExFiveUserConfFour+\ExSixUserConfFour}
\pgfmathsetmacro{\PartTwoBadConfFive}{\ExFiveUserConfFive+\ExSixUserConfFive}
\pgfmathsetmacro{\PartTwoBadConfSix}{\ExFiveUserConfSix+\ExSixUserConfSix}
\pgfmathsetmacro{\PartTwoBadConfSeven}{\ExFiveUserConfSeven+\ExSixUserConfSeven}

\pgfmathsetmacro{\PartTwoGoodCorrectMean}{\ComputeBinaryMean{\PartTwoGoodCorrect}{\PartTwoGoodIncorrect}}
\pgfmathsetmacro{\PartTwoGoodCorrectStd}{\ComputeBinaryStd{\PartTwoGoodCorrect}{\PartTwoGoodIncorrect}}

\pgfmathsetmacro{\PartTwoGoodConfMean}{\ComputeLikertMean{\PartTwoGoodConfOne}{\PartTwoGoodConfTwo}{\PartTwoGoodConfThree}{\PartTwoGoodConfFour}{\PartTwoGoodConfFive}{\PartTwoGoodConfSix}{\PartTwoGoodConfSeven}}
\pgfmathsetmacro{\PartTwoGoodConfStd}{\ComputeLikertStd{\PartTwoGoodConfOne}{\PartTwoGoodConfTwo}{\PartTwoGoodConfThree}{\PartTwoGoodConfFour}{\PartTwoGoodConfFive}{\PartTwoGoodConfSix}{\PartTwoGoodConfSeven}}

\pgfmathsetmacro{\PartTwoBadCorrectMean}{\ComputeBinaryMean{\PartTwoBadCorrect}{\PartTwoBadIncorrect}}
\pgfmathsetmacro{\PartTwoBadCorrectStd}{\ComputeBinaryStd{\PartTwoBadCorrect}{\PartTwoBadIncorrect}}

\pgfmathsetmacro{\PartTwoBadConfMean}{\ComputeLikertMean{\PartTwoBadConfOne}{\PartTwoBadConfTwo}{\PartTwoBadConfThree}{\PartTwoBadConfFour}{\PartTwoBadConfFive}{\PartTwoBadConfSix}{\PartTwoBadConfSeven}}
\pgfmathsetmacro{\PartTwoBadConfStd}{\ComputeLikertStd{\PartTwoBadConfOne}{\PartTwoBadConfTwo}{\PartTwoBadConfThree}{\PartTwoBadConfFour}{\PartTwoBadConfFive}{\PartTwoBadConfSix}{\PartTwoBadConfSeven}}

\newcommand{\BoxPrefRect}{11}   
\newcommand{\BoxPrefMask}{5}   
\newcommand{\BoxPrefNoPref}{1}

\newcommand{\BoxRectConfOne}{0}
\newcommand{\BoxRectConfTwo}{1}
\newcommand{\BoxRectConfThree}{0}
\newcommand{\BoxRectConfFour}{1}
\newcommand{\BoxRectConfFive}{3}
\newcommand{\BoxRectConfSix}{2}
\newcommand{\BoxRectConfSeven}{4}

\newcommand{\BoxMaskConfOne}{0}
\newcommand{\BoxMaskConfTwo}{0}
\newcommand{\BoxMaskConfThree}{0}
\newcommand{\BoxMaskConfFour}{2}
\newcommand{\BoxMaskConfFive}{1}
\newcommand{\BoxMaskConfSix}{2}
\newcommand{\BoxMaskConfSeven}{0}

\newcommand{\HeatPrefHard}{4}     
\newcommand{\HeatPrefColor}{10}    
\newcommand{\HeatPrefNoPref}{3}

\newcommand{\HeatHardMisleadOne}{0}
\newcommand{\HeatHardMisleadTwo}{0}
\newcommand{\HeatHardMisleadThree}{0}
\newcommand{\HeatHardMisleadFour}{0}
\newcommand{\HeatHardMisleadFive}{1}
\newcommand{\HeatHardMisleadSix}{1}
\newcommand{\HeatHardMisleadSeven}{2}

\newcommand{\HeatColorMisleadOne}{0}
\newcommand{\HeatColorMisleadTwo}{1}
\newcommand{\HeatColorMisleadThree}{0}
\newcommand{\HeatColorMisleadFour}{6}
\newcommand{\HeatColorMisleadFive}{2}
\newcommand{\HeatColorMisleadSix}{0}
\newcommand{\HeatColorMisleadSeven}{1}

\pgfmathsetmacro{\BoxPrefTotal}{\BoxPrefRect + \BoxPrefMask + \BoxPrefNoPref}

\pgfmathsetmacro{\BoxPrefRectRatio}{\BoxPrefRect / max(1,\BoxPrefTotal)}
\pgfmathsetmacro{\BoxPrefMaskRatio}{\BoxPrefMask / max(1,\BoxPrefTotal)}
\pgfmathsetmacro{\BoxPrefNoPrefRatio}{\BoxPrefNoPref / max(1,\BoxPrefTotal)}

\pgfmathsetmacro{\BoxPrefRectPercent}{100 * \BoxPrefRectRatio}
\pgfmathsetmacro{\BoxPrefMaskPercent}{100 * \BoxPrefMaskRatio}
\pgfmathsetmacro{\BoxPrefNoPrefPercent}{100 * \BoxPrefNoPrefRatio}

\pgfmathsetmacro{\BoxRectTrustMean}{%
\ComputeLikertMean{\BoxRectConfOne}{\BoxRectConfTwo}{\BoxRectConfThree}{\BoxRectConfFour}{\BoxRectConfFive}{\BoxRectConfSix}{\BoxRectConfSeven}}

\pgfmathsetmacro{\BoxRectTrustStd}{%
\ComputeLikertStd{\BoxRectConfOne}{\BoxRectConfTwo}{\BoxRectConfThree}{\BoxRectConfFour}{\BoxRectConfFive}{\BoxRectConfSix}{\BoxRectConfSeven}}

\pgfmathsetmacro{\BoxMaskTrustMean}{%
\ComputeLikertMean{\BoxMaskConfOne}{\BoxMaskConfTwo}{\BoxMaskConfThree}{\BoxMaskConfFour}{\BoxMaskConfFive}{\BoxMaskConfSix}{\BoxMaskConfSeven}}

\pgfmathsetmacro{\BoxMaskTrustStd}{%
\ComputeLikertStd{\BoxMaskConfOne}{\BoxMaskConfTwo}{\BoxMaskConfThree}{\BoxMaskConfFour}{\BoxMaskConfFive}{\BoxMaskConfSix}{\BoxMaskConfSeven}}

\pgfmathsetmacro{\HeatPrefTotal}{\HeatPrefHard + \HeatPrefColor + \HeatPrefNoPref}

\pgfmathsetmacro{\HeatPrefHardRatio}{\HeatPrefHard / max(1,\HeatPrefTotal)}
\pgfmathsetmacro{\HeatPrefColorRatio}{\HeatPrefColor / max(1,\HeatPrefTotal)}
\pgfmathsetmacro{\HeatPrefNoPrefRatio}{\HeatPrefNoPref / max(1,\HeatPrefTotal)}

\pgfmathsetmacro{\HeatPrefHardPercent}{100 * \HeatPrefHardRatio}
\pgfmathsetmacro{\HeatPrefColorPercent}{100 * \HeatPrefColorRatio}
\pgfmathsetmacro{\HeatPrefNoPrefPercent}{100 * \HeatPrefNoPrefRatio}

\pgfmathsetmacro{\HeatHardMisleadMean}{%
\ComputeLikertMean{\HeatHardMisleadOne}{\HeatHardMisleadTwo}{\HeatHardMisleadThree}{\HeatHardMisleadFour}{\HeatHardMisleadFive}{\HeatHardMisleadSix}{\HeatHardMisleadSeven}}

\pgfmathsetmacro{\HeatHardMisleadStd}{%
\ComputeLikertStd{\HeatHardMisleadOne}{\HeatHardMisleadTwo}{\HeatHardMisleadThree}{\HeatHardMisleadFour}{\HeatHardMisleadFive}{\HeatHardMisleadSix}{\HeatHardMisleadSeven}}

\pgfmathsetmacro{\HeatColorMisleadMean}{%
\ComputeLikertMean{\HeatColorMisleadOne}{\HeatColorMisleadTwo}{\HeatColorMisleadThree}{\HeatColorMisleadFour}{\HeatColorMisleadFive}{\HeatColorMisleadSix}{\HeatColorMisleadSeven}}

\pgfmathsetmacro{\HeatColorMisleadStd}{%
\ComputeLikertStd{\HeatColorMisleadOne}{\HeatColorMisleadTwo}{\HeatColorMisleadThree}{\HeatColorMisleadFour}{\HeatColorMisleadFive}{\HeatColorMisleadSix}{\HeatColorMisleadSeven}}

\newcommand{\FALone}{1}
\newcommand{\FALtwo}{0}
\newcommand{\FALthree}{1}
\newcommand{\FALfour}{3}
\newcommand{\FALfive}{5}
\newcommand{\FALsix}{6}
\newcommand{\FALseven}{1}

\newcommand{\FBLone}{2}
\newcommand{\FBLtwo}{0}
\newcommand{\FBLthree}{2}
\newcommand{\FBLfour}{3}
\newcommand{\FBLfive}{4}
\newcommand{\FBLsix}{4}
\newcommand{\FBLseven}{2}

\newcommand{\FCLone}{1}
\newcommand{\FCLtwo}{0}
\newcommand{\FCLthree}{3}
\newcommand{\FCLfour}{1}
\newcommand{\FCLfive}{6}
\newcommand{\FCLsix}{5}
\newcommand{\FCLseven}{1}

\newcommand{\FDLone}{0}
\newcommand{\FDLtwo}{0}
\newcommand{\FDLthree}{2}
\newcommand{\FDLfour}{0}
\newcommand{\FDLfive}{3}
\newcommand{\FDLsix}{8}
\newcommand{\FDLseven}{4}

\newcommand{\FELone}{0}
\newcommand{\FELtwo}{0}
\newcommand{\FELthree}{1}
\newcommand{\FELfour}{2}
\newcommand{\FELfive}{1}
\newcommand{\FELsix}{5}
\newcommand{\FELseven}{8}

\newcommand{\FFLone}{0}
\newcommand{\FFLtwo}{1}
\newcommand{\FFLthree}{0}
\newcommand{\FFLfour}{2}
\newcommand{\FFLfive}{5}
\newcommand{\FFLsix}{6}
\newcommand{\FFLseven}{3}

\newcommand{\FGLone}{3}
\newcommand{\FGLtwo}{2}
\newcommand{\FGLthree}{3}
\newcommand{\FGLfour}{2}
\newcommand{\FGLfive}{3}
\newcommand{\FGLsix}{4}
\newcommand{\FGLseven}{0}

\newcommand{\FHLone}{2}
\newcommand{\FHLtwo}{5}
\newcommand{\FHLthree}{1}
\newcommand{\FHLfour}{3}
\newcommand{\FHLfive}{3}
\newcommand{\FHLsix}{2}
\newcommand{\FHLseven}{1}

\newcommand{\FILone}{0}
\newcommand{\FILtwo}{0}
\newcommand{\FILthree}{1}
\newcommand{\FILfour}{2}
\newcommand{\FILfive}{7}
\newcommand{\FILsix}{5}
\newcommand{\FILseven}{2}

\newcommand{\TALone}{0}
\newcommand{\TALtwo}{0}
\newcommand{\TALthree}{0}
\newcommand{\TALfour}{3}
\newcommand{\TALfive}{7}
\newcommand{\TALsix}{3}
\newcommand{\TALseven}{4}

\newcommand{\TBLone}{0}
\newcommand{\TBLtwo}{0}
\newcommand{\TBLthree}{0}
\newcommand{\TBLfour}{1}
\newcommand{\TBLfive}{3}
\newcommand{\TBLsix}{6}
\newcommand{\TBLseven}{7}

\newcommand{\TCLone}{0}
\newcommand{\TCLtwo}{0}
\newcommand{\TCLthree}{2}
\newcommand{\TCLfour}{0}
\newcommand{\TCLfive}{3}
\newcommand{\TCLsix}{7}
\newcommand{\TCLseven}{5}

\newcommand{\UALone}{1}
\newcommand{\UALtwo}{0}
\newcommand{\UALthree}{1}
\newcommand{\UALfour}{0}
\newcommand{\UALfive}{7}
\newcommand{\UALsix}{3}
\newcommand{\UALseven}{5}

\newcommand{\UBLone}{0}
\newcommand{\UBLtwo}{3}
\newcommand{\UBLthree}{3}
\newcommand{\UBLfour}{3}
\newcommand{\UBLfive}{6}
\newcommand{\UBLsix}{1}
\newcommand{\UBLseven}{1}

\newcommand{\UCLone}{1}
\newcommand{\UCLtwo}{1}
\newcommand{\UCLthree}{1}
\newcommand{\UCLfour}{0}
\newcommand{\UCLfive}{2}
\newcommand{\UCLsix}{8}
\newcommand{\UCLseven}{4}

\pgfmathsetmacro{\FAMean}{\ComputeLikertMean{\FALone}{\FALtwo}{\FALthree}{\FALfour}{\FALfive}{\FALsix}{\FALseven}}
\pgfmathsetmacro{\FAStd}{\ComputeLikertStd{\FALone}{\FALtwo}{\FALthree}{\FALfour}{\FALfive}{\FALsix}{\FALseven}}

\pgfmathsetmacro{\FBMean}{\ComputeLikertMean{\FBLone}{\FBLtwo}{\FBLthree}{\FBLfour}{\FBLfive}{\FBLsix}{\FBLseven}}
\pgfmathsetmacro{\FBStd}{\ComputeLikertStd{\FBLone}{\FBLtwo}{\FBLthree}{\FBLfour}{\FBLfive}{\FBLsix}{\FBLseven}}

\pgfmathsetmacro{\FCMean}{\ComputeLikertMean{\FCLone}{\FCLtwo}{\FCLthree}{\FCLfour}{\FCLfive}{\FCLsix}{\FCLseven}}
\pgfmathsetmacro{\FCStd}{\ComputeLikertStd{\FCLone}{\FCLtwo}{\FCLthree}{\FCLfour}{\FCLfive}{\FCLsix}{\FCLseven}}

\pgfmathsetmacro{\FDMean}{\ComputeLikertMean{\FDLone}{\FDLtwo}{\FDLthree}{\FDLfour}{\FDLfive}{\FDLsix}{\FDLseven}}
\pgfmathsetmacro{\FDStd}{\ComputeLikertStd{\FDLone}{\FDLtwo}{\FDLthree}{\FDLfour}{\FDLfive}{\FDLsix}{\FDLseven}}

\pgfmathsetmacro{\FEMean}{\ComputeLikertMean{\FELone}{\FELtwo}{\FELthree}{\FELfour}{\FELfive}{\FELsix}{\FELseven}}
\pgfmathsetmacro{\FEStd}{\ComputeLikertStd{\FELone}{\FELtwo}{\FELthree}{\FELfour}{\FELfive}{\FELsix}{\FELseven}}

\pgfmathsetmacro{\FFMean}{\ComputeLikertMean{\FFLone}{\FFLtwo}{\FFLthree}{\FFLfour}{\FFLfive}{\FFLsix}{\FFLseven}}
\pgfmathsetmacro{\FFStd}{\ComputeLikertStd{\FFLone}{\FFLtwo}{\FFLthree}{\FFLfour}{\FFLfive}{\FFLsix}{\FFLseven}}

\pgfmathsetmacro{\FGMean}{\ComputeLikertMean{\FGLone}{\FGLtwo}{\FGLthree}{\FGLfour}{\FGLfive}{\FGLsix}{\FGLseven}}
\pgfmathsetmacro{\FGStd}{\ComputeLikertStd{\FGLone}{\FGLtwo}{\FGLthree}{\FGLfour}{\FGLfive}{\FGLsix}{\FGLseven}}

\pgfmathsetmacro{\FHMean}{\ComputeLikertMean{\FHLone}{\FHLtwo}{\FHLthree}{\FHLfour}{\FHLfive}{\FHLsix}{\FHLseven}}
\pgfmathsetmacro{\FHStd}{\ComputeLikertStd{\FHLone}{\FHLtwo}{\FHLthree}{\FHLfour}{\FHLfive}{\FHLsix}{\FHLseven}}

\pgfmathsetmacro{\FIMean}{\ComputeLikertMean{\FILone}{\FILtwo}{\FILthree}{\FILfour}{\FILfive}{\FILsix}{\FILseven}}
\pgfmathsetmacro{\FIStd}{\ComputeLikertStd{\FILone}{\FILtwo}{\FILthree}{\FILfour}{\FILfive}{\FILsix}{\FILseven}}

\pgfmathsetmacro{\TAMean}{\ComputeLikertMean{\TALone}{\TALtwo}{\TALthree}{\TALfour}{\TALfive}{\TALsix}{\TALseven}}
\pgfmathsetmacro{\TAStd}{\ComputeLikertStd{\TALone}{\TALtwo}{\TALthree}{\TALfour}{\TALfive}{\TALsix}{\TALseven}}

\pgfmathsetmacro{\TBMean}{\ComputeLikertMean{\TBLone}{\TBLtwo}{\TBLthree}{\TBLfour}{\TBLfive}{\TBLsix}{\TBLseven}}
\pgfmathsetmacro{\TBStd}{\ComputeLikertStd{\TBLone}{\TBLtwo}{\TBLthree}{\TBLfour}{\TBLfive}{\TBLsix}{\TBLseven}}

\pgfmathsetmacro{\TCMean}{\ComputeLikertMean{\TCLone}{\TCLtwo}{\TCLthree}{\TCLfour}{\TCLfive}{\TCLsix}{\TCLseven}}
\pgfmathsetmacro{\TCStd}{\ComputeLikertStd{\TCLone}{\TCLtwo}{\TCLthree}{\TCLfour}{\TCLfive}{\TCLsix}{\TCLseven}}

\pgfmathsetmacro{\UAMean}{\ComputeLikertMean{\UALone}{\UALtwo}{\UALthree}{\UALfour}{\UALfive}{\UALsix}{\UALseven}}
\pgfmathsetmacro{\UAStd}{\ComputeLikertStd{\UALone}{\UALtwo}{\UALthree}{\UALfour}{\UALfive}{\UALsix}{\UALseven}}

\pgfmathsetmacro{\UBMean}{\ComputeLikertMean{\UBLone}{\UBLtwo}{\UBLthree}{\UBLfour}{\UBLfive}{\UBLsix}{\UBLseven}}
\pgfmathsetmacro{\UBStd}{\ComputeLikertStd{\UBLone}{\UBLtwo}{\UBLthree}{\UBLfour}{\UBLfive}{\UBLsix}{\UBLseven}}

\pgfmathsetmacro{\UCMean}{\ComputeLikertMean{\UCLone}{\UCLtwo}{\UCLthree}{\UCLfour}{\UCLfive}{\UCLsix}{\UCLseven}}
\pgfmathsetmacro{\UCStd}{\ComputeLikertStd{\UCLone}{\UCLtwo}{\UCLthree}{\UCLfour}{\UCLfive}{\UCLsix}{\UCLseven}}

\title{Towards Self-Explainable Document Visual Question Answering with Chain-of-Explanation Predictions}

\author{%
  Kjetil Indrehus\textsuperscript{1,2}\quad
  Adrian Duric\textsuperscript{1,2,3}\quad
  Changkyu Choi\textsuperscript{1}\quad
  Ali Ramezani-Kebrya\textsuperscript{1,2,3} \\[0.5em]
  \textsuperscript{1}Department of Informatics, University of Oslo\\
  \textsuperscript{2}Integreat -- Norwegian Centre for Knowledge-driven Machine Learning\\
  \textsuperscript{3}TRUST -- The Norwegian Centre for Trustworthy AI\\[0.3em]
  \texttt{\{kjetiki, adriandu, changkyc, ali\}@ifi.uio.no}
}
\usepackage[table]{xcolor}

\begin{document}

\maketitle

\begin{abstract}
Document Visual Question Answering (DocVQA) requires vision–language models to reason not only about \emph{what} information in a document is relevant to a question, but also \emph{where} the answer is grounded on the page. Existing DocVQA models entangle question-relevant evidence and answer localization and operate largely as black boxes, offering limited means to verify how predictions depend on visual evidence. We propose \emph{CoExVQA}, a self-explainable DocVQA framework with a grounded reasoning process through a \emph{chain-of-explanation} design. CoExVQA first identifies question-relevant evidence, then explicitly localizes the answer region, and finally decodes the answer exclusively from the grounded region. Prediction via CoExVQA's chain-of-explanation enables direct inspection and verification of the reasoning process across modalities. Empirical results show that restricting decoding to grounded evidence achieves \emph{SotA explainable DocVQA performance} on PFL-DocVQA, improving ANLS by $12\%$ over the current explainable baselines while providing transparent and verifiable predictions.
\end{abstract}

\section{Introduction}

While high prediction accuracy is vital for adopting machine learning in modern systems, it is not the only requirement~\cite{KAZMIERCZAK2025103184, s25103020}. This is particularly pronounced for vision-language models (VLMs), where predictions stem from complex, often opaque interactions between text and images. In high-stakes fields such as healthcare~\cite{bose2025visualalignmentmedicalvisionlanguage, nazir2023survey, saraswat2022aiforhealthcare,  nguyen2021automatedgenerationaccurate}, finance~\cite{arsenault2025xaifinance, giudici2023safe, cao2022ai}, and autonomous driving~\cite{jiang2025surveyvisionlanguageactionmodelsautonomous, gupta2021deep, badue2021self}, transparency is equally critical. This demand has driven the development of explainable artificial intelligence (XAI)~\cite{longo2024explainable, ALI2023101805, SAEED2023110273, yang2023survey_xai, ribeiro2016should}, where users can validate predictions, understand failure modes of a model, and enhance their confidence in the model's judgments through explanations of the decisions. 

Decision transparency requires novel VLM designs in which the features used for prediction are explicitly observable. Conventional post-hoc visual explanations do not always reflect the features actually used by the model~\cite{wu2025faithfylvit}, and language rationales can be misleading about what truly drives a prediction~\cite{chen2024multiobjecthallucinationvisionlanguagemodels, turpin2023languagemodelsdontsay}. For document-based VLMs, users can directly compare highlighted regions against the source page, making spatial grounding a particularly effective mechanism for verifying predictions.

Across domains, the ability to reason over documents has become increasingly important~\cite{barboule2025surveyquestionansweringvisually, faysse2025colpaliefficientdocumentretrieval}. A representative instance of this capability is DocVQA, where a model must jointly reason over document layout, visual structure, and a textual question to produce a correct textual answer~\cite{mathew2021docvqadatasetvqadocument}. Despite strong performance, many existing DocVQA models remain opaque, offering limited insight into the evidence underlying their predictions~\cite{souibgui2025docvxqacontextawarevisualexplanations}. These methods continue to push predictive accuracy \cite{huang2022layoutlmv3, lee2023pix2structscreenshotparsingpretraining, kim2022ocrfreedocumentunderstandingtransformer}, but none inherently produce the explanations \emph{required} for high-stakes adoption.

To address this challenge, we propose \ModelName{}, \ModelSpell.
Drawing inspiration from the \textit{chain-of-explainability} (CoE) design~\cite{yu2025coechainofexplanationautomaticvisual}, our method reinterprets explainability as a structured prediction process that separates \emph{what} evidence is relevant to the question from \emph{where} the answer is grounded in the document, and decodes the answer exclusively from this grounded region. Concretely, \ModelName{} is self-explainable by design, producing \emph{two complementary, spatially grounded explanations} that make both dimensions explicit.

Our main contributions are as follows:
\begin{itemize}[noitemsep, topsep=0pt, parsep=0pt, partopsep=0pt]
    \item We propose \ModelName{}, a novel self-explainable DocVQA framework that enforces decision transparency through a sequential \emph{Chain-of-Explanations} formulation, separating question-conditioned evidence selection from answer localization.

    \item We achieve \emph{SotA explainable performance on PFL-DocVQA}~\cite{tito2024privacyawaredocumentvisualquestion},  a large-scale document understanding benchmark (${\sim}6\times$ larger than DocVQA~\cite{mathew2021docvqadatasetvqadocument}), outperforming current baselines by 12 absolute ANLS points. We evaluate on both DocVQA and PFL-DocVQA and demonstrate generalization across backbones (Donut-Base 200M, Pix2Struct-Large 1.3B).

    \item We validate the faithfulness of the explanation chain through intervention-based masking experiments, demonstrating that the model functionally relies on the predicted evidence to produce its final answer.

    \item We conduct a structured user evaluation (\UserEvalParticipants{} participants) showing that \ModelName{}'s explanations are actionable; participants reliably distinguished correct from incorrect predictions and recovered correct answers from explanations alone, while reporting that explanations support verification without inducing blind trust.
\end{itemize}

\section{Related Work}
\label{sec:related_work}

\subsection{Document visual question answering}

DocVQA requires models to jointly perform document understanding and text generation from document images that are often visually rich~\cite{mathew2021docvqadatasetvqadocument}.
Most conventional DocVQA approaches adopt an image encoder–text decoder architecture, in which a document image is encoded into visual representations and a textual answer is generated autoregressively conditioned on the encoded document and the input question~\cite{faysse2025colpaliefficientdocumentretrieval,lee2023pix2structscreenshotparsingpretraining, kim2022ocrfreedocumentunderstandingtransformer}.
Several DocVQA approaches adopt an Optical Character Recognition (OCR)-free formulation, in which the document image is encoded directly and a textual answer is generated without an explicit text recognition stage~\cite{lee2023pix2structscreenshotparsingpretraining, kim2022ocrfreedocumentunderstandingtransformer}. These end-to-end architectures simplify the processing pipeline and achieve strong task performance. However, they typically do not expose the document regions that support a given prediction.

\subsection{Late-interaction retrieval priors}

Recent multimodal document retrieval work provides question-conditioned spatial priors for evidence localization on visually rich pages. ColPali represents each page as a set of patch embeddings and scores a text question using ColBERT-style late interaction \cite{faysse2025colpaliefficientdocumentretrieval, khattab2020colbertefficienteffectivepassage}. The results are fine-grained token-to-patch similarities that can be visualized as page heatmaps. Unlike OCR string-matching heuristics, these retriever heatmaps provide question-conditioned relevance distribution over the full page, which aligns well with supervising a question–evidence map rather than only the final answer region. This late-interaction token–patch similarity mechanism has since been reused beyond ColPali \cite{souibgui2025docvxqacontextawarevisualexplanations, masry2025colmate, cui2025attentiongroundedenhancementvisual}.

\subsection{Self-explainable DocVQA}

End-to-end DocVQA models predict the final textual answer in a single step, while the underlying evidence and grounding remain implicit. As a result, it is difficult to verify whether predictions rely on semantically relevant document regions or instead exploit spurious cues. Post-hoc attribution methods such as Grad-CAM \cite{Selvaraju_2019} can be applied but provide no guarantee that the highlighted regions reflect the features actually used for prediction \cite{wu2025faithfylvit}. Retriever-derived priors naturally align with self-explainable DocVQA models, which aim to expose explanation signals that are interpretable and directly coupled with the prediction process. 

One notable example is DocVXQA, a self-explainable DocVQA framework~\cite{souibgui2025docvxqacontextawarevisualexplanations}. Built upon Pix2Struct~\cite{lee2023pix2structscreenshotparsingpretraining}, DocVXQA learns an interpretable mask designed to remain faithful to the model prediction. The framework adopts three competing loss terms that enforce minimality, sufficiency, and interactivity, providing a principled information-theoretic basis for mask learning~\cite{choi2024dib}. Nevertheless, the semantic role of the learned mask remains ambiguous, as it may capture regions broadly relevant to the question rather than precisely localizing the answer itself, thereby necessitating extensive post-processing.

\subsection{Localization-based and text-aware VQA}

Spatial localization as an intermediate reasoning step has been explored extensively in natural-image VQA, including hard attention~\cite{Malinowski_2018_ECCV}, locate-then-generate pipelines~\cite{zhu2023locate, shao2024vcot}, coordinate-based decoding~\cite{chen2023shikraunleashingmultimodalllms, peng2023kosmos2groundingmultimodallarge}, and tool-augmented region selection~\cite{hu2024visualsketchpad, wang2025pixelreasonerincentivizingpixelspace}. In text-aware VQA, copy mechanisms link answer tokens to OCR source locations~\cite{hu2020iterative}, though this is a decoding strategy rather than an architectural constraint on evidence access. These methods target natural images or scene text and do not enforce that the decoder reasons exclusively from the localized region. We provide a detailed feature-level comparison in Appendix~\ref{app:extended_related_work}.

\section{Methodology}
\label{sec:method}

To address this, we structure the prediction as a sequence of interpretable intermediate explanations, where each step produces a human-interpretable explanation that can be inspected and influence the next step. We compare \emph{predictive performance against explainable baselines,} and \emph{explanation utility against both explainable and non-explainable alternatives}. Our framework adopt a \textit{chain-of-explainability} (CoE) design in which the model first identifies question-relevant evidence, then grounds the answer spatially, and finally decodes the answer from the grounded region \cite{yu2025coechainofexplanationautomaticvisual}. This sequentially chained design encourages faithfulness by construction. The model is forced to “show its work” in intermediate outputs that are used for the final prediction, rather than attaching explanations after the answer has already been decided.

\begin{figure*}[t!]
    \centering
    \includegraphics[width=0.9\linewidth]{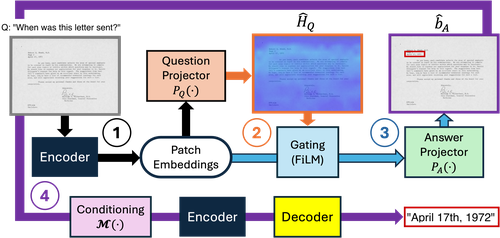}
    \caption{\textbf{Overview of the \ModelName{} prediction pipeline.} Given a document image and a question (left),
    $\circled{1}$ the image encoder produces patch-level embeddings.
    $\circled{2}$ The Question Projector $\QProj(\cdot)$ predicts a patch-level question–evidence alignment heatmap $\QMask$.
    $\circled{3}$ The predicted heatmap gates the patch embeddings via a FiLM module, yielding question-conditioned representations.
    $\circled{4}$ The Answer Projector $\AProj(\cdot)$ predicts an answer bounding box $\ABox$, which is used by the conditioning operator $\Mask(\cdot)$ to mask or crop the document; the conditioned image is re-encoded and passed to the same image encoder used in $\circled{1}$ and the text decoder to generate the final answer from the localized region only. The entire pipeline is OCR-free. Best viewed in the digital version.}
    \label{fig:architecture}
\end{figure*}

Building on this principle, we propose \ModelName{}, which instantiates a CoE formulation for DocVQA. At the level of prediction structure, \ModelName{} follows
\begin{equation}
    (\ImgEmb, q) \to \underbrace{\QMask \to \ABox}_{{\text{Learnable explanation}}} \to \PredA,
\end{equation}
which explicitly separates two explanatory roles within the prediction process.  Here, $\ImgEmb$ and $q$ respectively denote patch embeddings extracted by the image encoder and the input question, $\QMask$ is a question-evidence alignment heatmap, $\ABox$ is an answer-localization bounding box, and $\PredA$ is the decoded textual answer. The model produces two semantically distinct explanation signals: a question-evidence alignment heatmap $(\QMask)$ that highlights regions informative for the question, and an answer-localization bounding box $(\ABox)$ that identifies where the answer is derived.  This separation clarifies both what evidence supports the question and where the answer originates, improving interpretability over single-mask formulations that conflate these roles.

This design is based on the principle of self-explainability through masking introduced in DocVXQA~\cite{souibgui2025docvxqacontextawarevisualexplanations}. However, \ModelName{} differs in two key respects. First, it decomposes the explanation into two chained components with distinct semantic roles, one for question-evidence selection and one for answer grounding, where each stage constrains the next, enabling targeted failure diagnosis. Second, the predicted answer region acts as a hard information bottleneck: the decoder is re-encoded from $\ABox$ alone (Eq.~\ref{eq:mask}), so it cannot access information outside the selected region. This distinguishes \ModelName{} from prior localization-based VQA methods~\cite{zhu2023locate, shao2024vcot}, where the localization does not restrict downstream information flow, and from DocVXQA's single-mask formulation, whose semantic role is not explicitly separated. Figure~\ref{fig:architecture} shows the architecture. We provide an extended comparison in Appendix~\ref{app:extended_related_work}.

\subsection{Prediction pipeline}
\label{sec:pipeline}
Our framework builds on a pretrained vision-encoder $\Enc(\cdot)$, text-decoder $\Dec(\cdot)$ architecture~\cite{lee2023pix2structscreenshotparsingpretraining}.
Given a document image $x$ and a textual question $q$, the encoder renders the question onto the document image and produces patch-level visual embeddings $\ImgEmb = \Enc(x, q)$. A question projector $\QProj(\cdot)$ then predicts a patch-level question-evidence alignment heatmap $\QMask \in [0,1]^P$, where $P$ is the number of patches and $\sigma(\cdot)$ is the element-wise sigmoid:
\begin{equation}
    \ImgEmb = \Enc(x, q), \qquad
    \QMask = \sigma\!\bigg(\QProj(\ImgEmb, q)\bigg).
\end{equation}
We gate the image embeddings with $\QMask$ using Feature-wise Linear Modulation (FiLM)~\cite{perez2017filmvisualreasoninggeneral}, a lightweight conditioning mechanism that preserves the patch grid structure. Two MLPs $g_{\gamma}(\cdot)$ and $g_{\beta}(\cdot)$ map the mask to feature-wise scale and shift parameters $\boldsymbol{\gamma}=g_{\gamma}(\QMask)$ and $\boldsymbol{\beta}=g_{\beta}(\QMask)$ (ablations in Appendix~\ref{app:gating}). An answer projector $\AProj(\cdot)$ then predicts the answer bounding box from the gated embeddings:
\begin{equation}
    \GatedImgEmb = \ImgEmb \odot (1+\boldsymbol{\gamma}) + \boldsymbol{\beta}, \qquad
    \ABox = \AProj(\GatedImgEmb, q).
\end{equation}
To ensure that the decoded text is grounded in $\ABox$, we restrict the visual evidence using the predicted answer region. We consider two strategies: (1)~mask re-encoding, masking all pixels outside the answer region (Figure~\ref{fig:mask_reencode}), and (2)~crop re-encoding, cropping to the answer region (Figure~\ref{fig:crop_reencode}). Denoting either operation by $\Mask(\cdot)$, the conditioned image is re-encoded and decoded to produce the final answer $\PredA$:
\begin{equation}
    x_{\ABox} = \Mask(x;\, \ABox), \qquad
    {\ImgEmb}_{\ABox} = \Enc(x_{\ABox},\, q), \qquad
    \PredA = \Dec({\ImgEmb}_{\ABox},\, q).
    \label{eq:mask}
\end{equation}
Conditioning on \emph{ground-truth} answer regions confirms that localization-guided decoding is a viable strategy. Cropping achieves 
ANLS of 0.87, substantially above backbone's default of 0.77, 
while masking yields 0.54. We evaluate both variants throughout 
our experiments. Full upper bound results and alternative conditioning 
strategies are in Appendix~\ref{app:decoder_input}.

\newcommand{\bboxlegend}{\textcolor{red}{\fboxrule=0.6pt\fboxsep=1pt\fbox{\phantom{xx}}}}
\begin{figure}[ht!]
    \centering
    \begin{subfigure}[t]{0.42\textwidth}
        \centering
        \begin{tikzpicture}[x=1cm,y=1cm, font=\scriptsize, >=Latex]
          \def\W{2.6}\def\H{2.0}
          \def\x{0.75}\def\y{0.55}\def\bw{1.2}\def\bh{0.85}

          \draw[rounded corners=1pt, fill=gray!10] (0,0) rectangle (\W,\H);
          \draw[step=0.4, gray!35, line width=0.2pt] (0,0) grid (\W,\H);

          \draw[very thick, red] (\x,\y) rectangle (\x+\bw,\y+\bh);

          \fill[pattern=north east lines, pattern color=black!35, opacity=0.4] (0,0) rectangle (\W,\y);
          \fill[pattern=north east lines, pattern color=black!35, opacity=0.4] (0,\y+\bh) rectangle (\W,\H);
          \fill[pattern=north east lines, pattern color=black!35, opacity=0.4] (0,\y) rectangle (\x,\y+\bh);
          \fill[pattern=north east lines, pattern color=black!35, opacity=0.4] (\x+\bw,\y) rectangle (\W,\y+\bh);

          \draw[->, thick] (\W+0.05, \H/2) -- (\W+0.55, \H/2);
          \draw[rounded corners=1pt, fill=gray!15] (\W+0.6,\H/2-0.35) rectangle (\W+1.75,\H/2+0.35);
          \node at (\W+1.175,\H/2) {Encoder};
        \end{tikzpicture}
        \caption{\textbf{Mask re-encoding.} The predicted answer box (\bboxlegend) keeps pixels inside the region and \emph{masks out} everything else while preserving the original image size.}
        \label{fig:mask_reencode}
    \end{subfigure}%
    \hfill
    \begin{subfigure}[t]{0.56\textwidth}
        \centering
        \begin{tikzpicture}[x=1cm,y=1cm, font=\scriptsize, >=Latex]
          \def\W{2.6}\def\H{2.0}
          \def\x{0.75}\def\y{0.55}\def\bw{1.2}\def\bh{0.85}
          \def\gap{0.6}

          \draw[rounded corners=1pt, draw=gray!55, fill=gray!6] (0,0) rectangle (\W,\H);
          \draw[step=0.4, gray!30, line width=0.2pt] (0,0) grid (\W,\H);
          \draw[very thick, red] (\x,\y) rectangle (\x+\bw,\y+\bh);

          \draw[->, thick] (\W+0.05, \H/2) -- (\W+\gap-0.05, \H/2);

          \begin{scope}[shift={(\W+\gap,0)}]
            \draw[rounded corners=1pt, draw=gray!55, fill=gray!10] (0,0) rectangle (\W,\H);
            \draw[step=0.8, gray!35, line width=0.2pt] (0,0) grid (\W,\H);
            \draw[very thick, red] (0,0) rectangle (\W,\H);
            \draw[red, very thick] (0,0) -- (0,0.25) (0,0) -- (0.25,0);
            \draw[red, very thick] (\W,\H) -- (\W-0.25,\H) (\W,\H) -- (\W,\H-0.25);
          \end{scope}

          \draw[->, thick] (2*\W+\gap+0.05, \H/2) -- (2*\W+\gap+0.55, \H/2);
          \draw[rounded corners=1pt, fill=gray!15]
            (2*\W+\gap+0.6,\H/2-0.35) rectangle (2*\W+\gap+1.75,\H/2+0.35);
          \node at (2*\W+\gap+1.175,\H/2) {Encoder};
        \end{tikzpicture}
        \caption{\textbf{Crop re-encoding.} The predicted answer box (\bboxlegend) defines a cropped view that is \emph{zoomed} to a full-sized image and re-encoded, so the encoder processes only the selected region.}
        \label{fig:crop_reencode}
    \end{subfigure}
    \caption{\textbf{Re-encoding variants.} The two re-encoding strategies used to refine the model's focus on the predicted answer region.}
    \label{fig:reencode_strategies}
\end{figure}
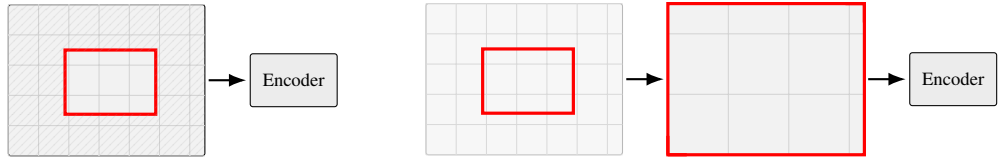

\subsection{Weakly supervised question–evidence alignment}
\label{sec:colpali}

\ModelName{} proposes to supervise the question-evidence alignment heatmaps $\QMask$ to overlap with ColPali-based priors $\QMaskPrior \in [0,1]^P$ \cite{faysse2025colpaliefficientdocumentretrieval}. These priors are \emph{question-conditioned patch relevance scores} obtained from the late-interaction similarity between the question-tokens and image-patches, spatially aligned to the backbone's patch grid (512 patches by default). Following DocVXQA, we treat the priors $\QMaskPrior$ as weak spatial supervision for learning the question–evidence alignment heatmaps $\QMask$.The question supervision loss $\mathcal{L}_Q$ is a MSE loss, weighted by the hyperparameter $\lambda_{prior}$ 
\begin{equation}
    \mathcal{L}_Q = \lambda_{prior} \;  \mathrm{MSE} \left( \QMask, \QMaskPrior \right).
\end{equation}
We further describe the prior construction and post-processing in Appendix~\ref{app:question_priors}.

\subsection{Answer localization supervision}
\label{sec:answer_loc_supervision}

We train an answer projector $\AProj(\cdot)$ to predict a bounding box $\ABox = (\hat{c}_x, \hat{c}_y, \hat{w}, \hat{h}) \in [0,1]^4$ in relative page coordinates. Ground-truth answer location priors $\ABoxPrior = (c_x, c_y, w, h)$ are obtained by matching the answer text to OCR lines (Appendix~\ref{app:answer_priors}). We supervise localization with three complementary losses: a $\mathrm{GIoU}$ overlap loss~\cite{rezatofighi2019generalizedintersectionunionmetric} that provides useful gradients even for non-overlapping boxes, a centre regression loss that stabilizes training, and a scale-invariant area loss inspired by YOLO-style square-root box regression~\cite{yolov13, redmon2016lookonceunifiedrealtime},
\begin{equation}
    \mathcal{L}_{\mathrm{GIoU}} = 1 - \mathrm{GIoU}(\ABox, \ABoxPrior), \qquad
    \mathcal{L}_{\text{centre}} = \left\| \hat{c} - c \right\|_2, \qquad
    \mathcal{L}_{\text{area}} = \left(\sqrt{\frac{\Aboxw \Aboxh}{\AboxPriorw \AboxPriorh}} - 1\right)^{\!2}.
    \label{eq:losses}
\end{equation}
The final answer localization objective is $\mathcal{L}_{\text{proj}} = \lambda_{\mathrm{GIoU}}\mathcal{L}_{\mathrm{GIoU}} + \lambda_{\text{centre}}\mathcal{L}_{\text{centre}} + \lambda_{\text{area}}\mathcal{L}_{\text{area}}$.

\subsection{Training objective}

All losses are computed per sample and averaged over the mini-batch during training. The overall objective combines supervision losses for heatmap alignment and answer location alignment, with the standard decoder cross-entropy loss for answer generation $\text{CE}_{\mathrm{Dec}}(\cdot)$. The decoder is regularized  with $\lambda_{\text{dec}}$, to balance the importance of predicting the correct answer location and decoding the correct answer. The overall training objective is denoted as 

\begin{equation}
    \min \mathcal{L}_{\text{\ModelName{}}} = \mathcal{L}_Q + \mathcal{L}_{\text{proj}} + \lambda_{\mathrm{Dec}} \; \text{CE}_{\mathrm{Dec}}(\PredA, \A).
\end{equation}

\subsection{Answer localization metrics}
\label{sec:answer_localization_metrics}
We evaluate answer localization with two overlap metrics and one auxiliary size metric over $N$ validation examples, where $b_n$ and $\hat{b}_n$ denote the ground-truth and predicted answer boxes for example $n$. $\mathrm{IoU}$ measures how tightly the predicted box overlaps the target, penalizing misalignment and oversized predictions. Since $\mathrm{IoU}$ can be small even when the prediction fully contains the answer, we additionally report $\mathrm{Coverage}$, the fraction of the target area covered by the predicted box. The area ratio $\mathrm{AR}$ serves as a diagnostic for size bias. A large $\mathrm{AR}$ is not necessarily undesirable, as a semantically coherent context region can yield $\mathrm{AR}\gg 1$. We substantiate this with an analysis of high-AR explanations in Appendix~\ref{app:ar_examples}.

\begin{equation}
    \mathrm{IoU}_{\text{m}} = \frac{1}{N}\!\sum_{n}\frac{|\hat{b}_n \cap b_n|}{|\hat{b}_n \cup b_n|}, \qquad
    \mathrm{Cov}_{\text{m}} = \frac{1}{N}\!\sum_{n}\frac{|\hat{b}_n \cap b_n|}{|b_n|}, \qquad
    \mathrm{AR}_{\text{m}} = \frac{1}{N}\!\sum_{n}\frac{|\hat{b}_n|}{|b_n|}.
\end{equation}

\section{Experiments and results}

\subsection{Experimental setup}

We conduct all experiments on the DocVQA dataset \cite{mathew2021docvqadatasetvqadocument} and adopt the pretrained \texttt{pix2struct-docvqa-base}\footnote{\url{https://huggingface.co/google/pix2struct-docvqa-base}} model as the backbone \cite{lee2023pix2structscreenshotparsingpretraining}.
During training and validation, we augment the data with two forms of weak spatial supervision: (1) answer location priors $\ABoxPrior$ introduced in \Cref{sec:answer_loc_supervision}, and (2) question relevance priors $\QMaskPrior$ introduced in \Cref{sec:colpali}, with full details provided in Appendix~\ref{app:question_priors}. All components are modular and configured via preset model configurations. We provide architectural and configuration details in Appendix \ref{app:architecture_details}.

We train \ModelName{} using a two-stage schedule. In the first stage, we linearly warm up the decoder loss weight from $0$ to $\lambda_{\text{dec}}$ over the first 10 epochs to stabilize answer localization. This allows the model to improve localization before learning to decode on a region that does not cover the ground truth answer location. In the second stage, we train with decoder loss at full weight $\lambda_{\text{dec}}$. Training details and a fine-tuning table can be found in Appendix \ref{app:training_details}. 

\subsection{Baselines}
\label{sec:baselines}

We compare with existing publications on the DocVQA \cite{lee2023pix2structscreenshotparsingpretraining, kim2022ocrfreedocumentunderstandingtransformer, souibgui2025docvxqacontextawarevisualexplanations}, and PFL-DocVQA dataset \cite{tito2024privacyawaredocumentvisualquestion} report on existing DocVQA metrics \cite{mathew2021docvqadatasetvqadocument} in addition to our proposed answer localization metrics. We perform backbone ablation, and show the framework is not restricted to a single backbone.

\newcommand{\gray}[1]{\textcolor{gray}{#1}}
\newcommand{\best}[1]{\cellcolor{blue!12}\textbf{#1}}
\newcommand{\gapclosed}[2]{\cellcolor{blue!12}\textbf{#1}\;\scriptsize\textcolor{green!50!black}{(\,$\uparrow$\!#2)}}

\begin{table*}[t!]
    \centering
    \small
    \caption{\textbf{Task performance compared to baselines.} Baseline comparison results on DocVQA by \cite{mathew2021docvqadatasetvqadocument} and PFL-DocVQA by \cite{tito2024privacyawaredocumentvisualquestion}. Our Crop Variant achieves \textbf{SotA among explainable methods on PFL-DocVQA}, closing \textbf{49\%} of the ACC and \textbf{46\%} of the ANLS gap to the non-explainable upper bound. \best{Best} explainable result per metric is highlighted.}
    \setlength{\tabcolsep}{4.5pt}
    \resizebox{\textwidth}{!}{%
    \begin{tabular}{l c ccccc @{\hspace{6pt}\vrule width 0.4pt\hspace{6pt}} ccccc}
        \toprule
        & &
        \multicolumn{5}{c}{\textbf{DocVQA} \cite{mathew2021docvqadatasetvqadocument}} &
        \multicolumn{5}{c}{\textbf{PFL-DocVQA} \cite{tito2024privacyawaredocumentvisualquestion}} \\
        \cmidrule(lr){3-7}\cmidrule(l){8-12}
        \textbf{Model} &
        \textbf{Expl.} &
        ACC $\uparrow$ & ANLS $\uparrow$ &
        $\mathrm{IoU}_{\text{m}}$ $\uparrow$ & $\mathrm{Cov}_{\text{m}}$ $\uparrow$ & $\mathrm{AR}_{\text{m}}$ &
        ACC $\uparrow$ & ANLS $\uparrow$ &
        $\mathrm{IoU}_{\text{m}}$ $\uparrow$ & $\mathrm{Cov}_{\text{m}}$ $\uparrow$ & $\mathrm{AR}_{\text{m}}$ \\
        \midrule
        \multicolumn{12}{l}{\gray{\textit{Non-explainable}}} \\
        \addlinespace[1pt]
        Pix2Struct$^{\dagger}$ \cite{lee2023pix2structscreenshotparsingpretraining}
          & \xmark & -- & 0.77 & -- & -- & --
                   & 0.80 & 0.92 & -- & -- & -- \\
        Donut \cite{kim2022ocrfreedocumentunderstandingtransformer}
          & \xmark & -- & 0.68 & -- & -- & --
                   & -- & -- & -- & -- & -- \\
        \midrule
        \multicolumn{12}{l}{\gray{\textit{Explainable}}} \\
        \addlinespace[1pt]
        DocVXQA$^{\ddagger}$ \cite{souibgui2025docvxqacontextawarevisualexplanations}
          & \cmark & \best{0.38} & \best{0.54} & -- & -- & --
                   & 0.43 & 0.66 & -- & -- & -- \\
        \addlinespace[3pt]
        \multicolumn{12}{l}{\textbf{\ModelName{} (Ours)}$^{\S}$} \\
        \addlinespace[2pt]
        \quad Mask Variant
          & \cmark & 0.10 & 0.24 & \best{0.16} & 0.28 & 2.13
                   & 0.35 & 0.63 & 0.43 & 0.69 & 2.85 \\
        \addlinespace[2pt]
        \quad Crop Variant
          & \cmark & 0.34 & 0.43 & 0.06 & \best{0.37} & \best{19.53}
                   & \best{0.61} & \best{0.78} & \best{0.47} & \best{0.76} & \best{4.92} \\
        \bottomrule
        \addlinespace[2pt]
        \multicolumn{12}{l}{\footnotesize $^{\dagger}$\,4{,}096 patches. \quad $^{\ddagger}$\,1{,}024 patches. \quad $^{\S}$\,512 patches.} \\
    \end{tabular}%
    }
    \label{tab:baseline_comparison}
\end{table*}

Table \ref{tab:baseline_comparison} summarises the comparison between \ModelName{} and existing baselines on DocVQA and PFL-DocVQA, and Table \ref{tab:backbone_summary} presents the backbone ablation results. On PFL-DocVQA, our Crop variant achieves \emph{SotA performance among explainable methods,} closing approximately 49\% of the ACC gap and 46\% of the ANLS gap to the non-explainable Pix2Struct. Compared to DocVXQA, the Crop variant improves ANLS by 12 absolute points (0.66 to 0.78), while using $2\times$ fewer input patches and $8\times$ fewer input patches than the non-explainable Pix2Struct baseline. On DocVQA, DocVXQA leads by 11 ANLS points (0.54 vs.\ 0.43); however, it does not provide explicit answer localisation, precluding a full explainability comparison. These results suggest that the explainability–performance trade-off improves with dataset scale.

\begin{figure*}[t!]
  \centering
  \begin{subfigure}[t]{0.32\textwidth}
    \centering
    \includegraphics[width=\linewidth]{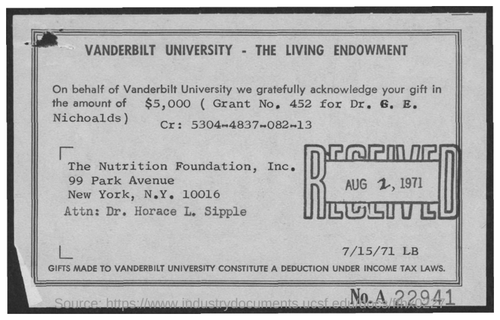}
    \caption{Original document given to the model.}
    \label{fig:example_3_a}
  \end{subfigure}\hfill
  \begin{subfigure}[t]{0.32\textwidth}
    \centering
    \includegraphics[width=\linewidth]{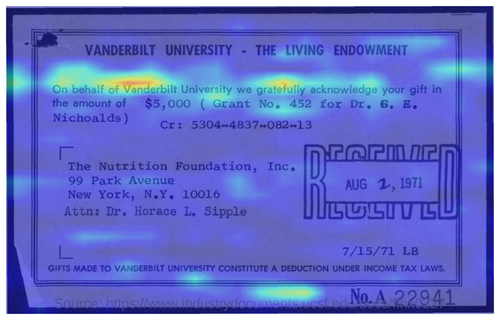}
    \caption{Question-heatmap ($\QMask$), visualized with a jet colormap (\textcolor{blue}{low} $\rightarrow$ \textcolor{orange}{high} relevance).}
    \label{fig:example_3_b}
  \end{subfigure}\hfill
  \begin{subfigure}[t]{0.32\textwidth}
    \centering
    \includegraphics[width=\linewidth]{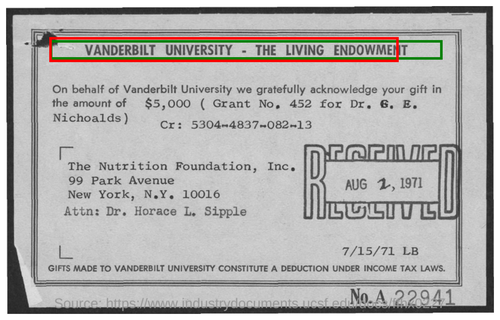}
    \caption{Predicts the answer region as a bounding box ($\ABox$, \textcolor{red}{red}), and the ground truth location ($\ABoxPrior$, \textcolor{ForestGreen}{green}).}
    \label{fig:example_3_c}
  \end{subfigure}

  \caption{\textbf{\ModelName{} Example Prediction.} Question given to the model: ``What is the name of the University?". \ref{fig:example_3_a} shows the original document given to the document. \ref{fig:example_3_b} shows the question heatmap overlay predicted by the model over the document. The model highlights ``Vanderbilt" with high correlation. \ref{fig:example_3_c} shows how the predicted answer location is correctly aligned with the ground truth answer location. From the answer region prediction, the model correctly predicts in text format the answer: ``Vanderbilt University". For more prediction examples, see Appendix \ref{app:analysis_of_predictions}.}
  \label{fig:example_3}
\end{figure*}

A grouped analysis by prediction accuracy (Appendix~\ref{app:ar_examples}, Table~\ref{tab:category_analysis}) confirms that localization quality correlates with answer quality: correct predictions achieve $5.4\times$ higher Coverage than incorrect ones (0.70 vs.\ 0.13) while selecting tighter regions (AR 12.9 vs.\ 24.3). We note that despite high AR relative to the ground-truth box, the predicted regions cover less than 3\% of the total page area, as ground-truth boxes typically span a single word or short phrase. $\mathcal{L}_{\text{area}}$ explicitly prevents degenerate full-document selection (Eq. \ref{eq:losses}). Figure \ref{fig:example_3} and Appendix \ref{app:analysis_of_predictions} provide qualitative examples of how predicted explanations align with semantically relevant document regions. Lastly, our backbone ablation study shows that \ModelName{} is not restricted to a single backbone (Appendix \ref{app:backbone_abl_study}). 

\begin{table}[t!]
    \centering
    \small
    \caption{\textbf{Backbone comparison on DocVQA and PFL-DocVQA.} We report the best performing crop configuration per backbone. The results show that the framework is not restricted to a singular backbone. Full hyperparameter sweeps in Appendix \ref{app:backbone_abl_study}.}
    \setlength{\tabcolsep}{4pt}
    \renewcommand{\arraystretch}{1.15}
    \resizebox{\linewidth}{!}{%
    \begin{tabular}{l r ccccc @{\hspace{6pt}\vrule width 0.4pt\hspace{6pt}} ccccc}
        \toprule
        & &
        \multicolumn{5}{c}{\textbf{DocVQA} \cite{mathew2021docvqadatasetvqadocument}} &
        \multicolumn{5}{c}{\textbf{PFL-DocVQA} \cite{tito2024privacyawaredocumentvisualquestion}} \\
        \cmidrule(lr){3-7}\cmidrule(l){8-12}
        \textbf{Backbone} &
        \textbf{Params} &
        ACC $\uparrow$ & ANLS $\uparrow$ &
        $\mathrm{IoU}_{\text{m}}$ $\uparrow$ & $\mathrm{Cov}_{\text{m}}$ $\uparrow$ & $\mathrm{AR}_{\text{m}}$ &
        ACC $\uparrow$ & ANLS $\uparrow$ &
        $\mathrm{IoU}_{\text{m}}$ $\uparrow$ & $\mathrm{Cov}_{\text{m}}$ $\uparrow$ & $\mathrm{AR}_{\text{m}}$ \\
        \midrule
        Pix2Struct-Large & 1.3B &
        \textbf{0.3801} & \textbf{0.4821} & \textbf{0.1144} & \textbf{0.3889} & 15.29 &
        \textbf{0.5845} & \textbf{0.7514} & \textbf{0.2649} & \textbf{0.7157} & 9.17 \\
        Donut-Base & 200M &
        0.1759 & 0.2677 & 0.0968 & 0.3804 & 19.22 &
        0.4217 & 0.5274 & 0.1742 & 0.6731 & 13.78 \\
        \bottomrule
    \end{tabular}%
    }
    \label{tab:backbone_summary}
\end{table}

\FloatBarrier
\subsection{Faithfulness and robustness evaluation}
\label{sec:faitfulness_and_robustness_eval}

We perform masking experiments with the predicted question alignment heatmap to evaluate its robustness and faithfulness. To obtain an estimate, we pass images from a validation dataset through a fully trained \ModelName{} model. Once we have predicted a question alignment heatmap within the forward pass, we mask its patches that are either overlapping or non-overlapping with the ColPali-based question prior, and continue the forward pass with the masked question heatmap. We either mask all overlapping patches or randomly mask each patch with a certain probability. For additional details and visualizations of the masking operations, see Appendix \ref{app:faithfulness_details}. 

\begin{table}[t!]
    \centering
    \small
    \caption{\textbf{Results from intervention based experiment.} All of the masked patches were either selected from the area of the question heatmap that overlapped with the question prior, or the non-overlapping part (denoted ``Non-QP"). The percentages denote the probabilities that a patch in the relevant area would be masked. Baseline corresponds to no masking (Table \ref{tab:baseline_comparison}). The right table additionally masks the embeddings of the corresponding patches. Bold indicates the top two scores per column.}
    \label{tab:results_faithfulness_robustness}
    \begin{minipage}[t]{0.48\textwidth}
        \centering
        \subcaption{Patches only.}
        \label{tab:results_faithfulness_robustness_patches_only}
        \begin{tabular}{lcc}
            \toprule
            \textbf{Overlapping Area} & \textbf{ACC} $\uparrow$ & \textbf{ANLS} $\uparrow$ \\
            \midrule
            Baseline (no masking) & \textbf{0.3352} & \textbf{0.4328} \\
            \midrule
            Question Prior (10\%) & 0.3257 & 0.4279 \\
            Question Prior (50\%) & 0.3277 & 0.4242 \\
            Question Prior (90\%) & 0.2892 & 0.3874 \\
            Question Prior (Full) & 0.2813 & 0.3748 \\
            \midrule
            Non-QP (10\%) & 0.3338 & 0.4320 \\
            Non-QP (50\%) & \textbf{0.3340} & \textbf{0.4324} \\
            Non-QP (90\%) & 0.3265 & 0.4226 \\
            Non-QP (Full) & 0.3231 & 0.4233 \\
            \bottomrule
        \end{tabular}
    \end{minipage}%
    \hfill
    \begin{minipage}[t]{0.48\textwidth}
        \centering
        \subcaption{Patches and embeddings.}
        \label{tab:results_faithfulness_robustness_patches_and_embeddings}
        \begin{tabular}{lcc}
            \toprule
            \textbf{Overlapping Area} & \textbf{ACC} $\uparrow$ & \textbf{ANLS} $\uparrow$ \\
            \midrule
            Baseline (no masking) & {0.3352} & {0.4328} \\
            \midrule
            Question Prior (10\%) & 0.3326 & 0.4339 \\
            Question Prior (50\%) & 0.3280 & 0.4264 \\
            Question Prior (90\%) & 0.2565 & 0.3456 \\
            Question Prior (Full) & 0.1748 & 0.2544 \\
            \midrule
            Non-QP (10\%) & 0.3403 & 0.4363 \\
            Non-QP (50\%) & 0.3434 & 0.4513 \\
            Non-QP (90\%) & \textbf{0.3560} & \textbf{0.4546} \\
            Non-QP (Full) & \textbf{0.3566} & \textbf{0.4565} \\
            \bottomrule
        \end{tabular}
    \end{minipage}
\end{table}

To evaluate the faithfulness of \ModelName{} to the question alignment heatmap, we measure how masking the question heatmap affects the quality of the final decoder output by measuring the difference in accuracy and ANLS compared to the variant without masking. If the model is faithful to the heatmap, masking its prior-aligned regions should progressively degrade performance. To evaluate the robustness of \ModelName{} explanations, we also perform experiments in which we mask areas that are considered unimportant for the DocVQA task (i.e., some of or all patches that do not overlap with the question prior).

From our results, we see that masking selected overlapping patches reduces performance. Masking the question heatmap alone reduces ACC from 0.3352 to 0.2813 ($-0.0539$, $\approx 16\%$ relative) and ANLS from 0.4328 to 0.3748 ($-0.0580$, $\approx 13\%$ relative) in Table~\ref{tab:results_faithfulness_robustness_patches_only}, demonstrating that the decoder actively relies on the heatmap's spatial signal even though the underlying embeddings remain available. When both the heatmap and the corresponding embeddings are masked, Table~\ref{tab:results_faithfulness_robustness_patches_and_embeddings} shows that the drop increases to 0.1748 ACC ($\approx 48\%$ relative) and 0.2544 ANLS ($\approx 41\%$ relative), confirming that the heatmap is also an effective proxy for identifying answer-relevant patches. The FiLM gating mechanism does not enforce a strict information bottleneck, but the model has learned to depend on the heatmap's spatial signal. This is further supported by the Non-QP results in Table~\ref{tab:results_faithfulness_robustness_patches_and_embeddings}, where removing patches \textit{outside} the heatmap's focus improves performance from 0.3352 to 0.3566 ACC.

For non-overlapping patches, the results are broadly consistent with the robustness claim. We observed small changes with respect to performance for the question heatmap only masking in Table \ref{tab:results_faithfulness_robustness_patches_only}. When we masked the non-overlapping embeddings, we observed that the performance did not degrade, but in fact increased. Table \ref{tab:results_faithfulness_robustness_patches_and_embeddings} shows that ACC changed from 0.3352 to 0.3566, and ANLS changed from 0.4328 to 0.4565 when we mask non-overlapping patches with 100\% probability. The results indicate that removing patches that do not align with the question prior is not harmful to model performance. Crucially, if the model had not learned to separate relevant from irrelevant regions, removing non-aligned patches would degrade performance. The consistent monotonic improvement confirms that the model correctly identifies these patches as uninformative.

The results of these interventions provide evidence that (i) the model is faithful to the question heatmap, (ii) it is therefore dependent on the heatmap overlapping with the question prior to make accurate predictions, and (iii) both the question-relevance aligned heatmap and the model's subsequent predictions are robust to removal of unimportant features. Together, the results demonstrate that the model has learned effective information separation between question-relevant and irrelevant regions. 

\FloatBarrier
\subsection{Qualitative user evaluation}
\label{sec:user_evaluation}

\pgfmathsetmacro{\BadIdentification}{1 - \BadCorrectMean}
\pgfmathsetmacro{\DeltaRecovery}{0.73}
\pgfmathsetmacro{\BraceLow}{\PartTwoBadCorrectMean}
\pgfmathsetmacro{\BraceHigh}{\PartTwoGoodCorrectMean}
\pgfmathtruncatemacro{\TotalEvals}{\UserEvalParticipants * 12}

We conducted a structured user evaluation with \UserEvalParticipants{} participants (\TotalEvals{} examples evaluated) to assess whether \ModelName{}'s explanations are actionable, enable verification of predictions, and make model failures apparent. The evaluation consisted of four parts: (1) answer justification, where participants judged whether predictions were supported by explanations; (2) answering with explanations only, where participants attempted to answer questions using only the model's visual explanations; (3) visualisation preferences, comparing localization and heatmap variants; and (4) a post-questionnaire assessing perceived faithfulness, trust, and usability on 7-point Likert scales. The complete protocol and results are given in Appendix~\ref{app:user_evaluation}.

As shown in Figure~\ref{fig:user_eval_combined}a, each participant evaluated correct and incorrect model predictions (identification rates of $\pgfmathprintnumber[precision=2]{\GoodCorrectMean}$ and $\pgfmathprintnumber[precision=2]{\BadIdentification}$). The key finding emerges in answer recovery. Participants recovered the correct answer from explanations alone at a rate of $\pgfmathprintnumber[precision=2]{\PartTwoGoodCorrectMean}$ for correct predictions, dropping to $\pgfmathprintnumber[precision=2]{\PartTwoBadCorrectMean}$ for incorrect ones ($\Delta = \pgfmathprintnumber[precision=2]{\DeltaRecovery}$), confirming that explanations are actionable without inducing false confidence. Post-questionnaire ratings (Figure~\ref{fig:user_eval_combined}b) further support this: participants agreed that the answer rectangle aided verification (\UserQref{FE}: $\pgfmathprintnumber[precision=2]{\FEMean}/7$) and helped decide when to trust predictions (\UserQref{TA}: $\pgfmathprintnumber[precision=2]{\TAMean}/7$), while also reporting they would double-check answers even when evidence appears strong (\UserQref{TC}: $\pgfmathprintnumber[precision=2]{\TCMean}/7$).

\definecolor{correctblue}{HTML}{2563EB}
\definecolor{incorrectred}{HTML}{DC2626}
\definecolor{correctbluefill}{HTML}{93C5FD}
\definecolor{incorrectredfill}{HTML}{FCA5A5}

\definecolor{faithcolor}{HTML}{2563EB}
\definecolor{trustcolor}{HTML}{DC2626}
\definecolor{usecolor}{HTML}{059669}

\begin{figure}[ht!]
    \centering
    \vspace{-0.3em}
    \resizebox{\textwidth}{!}{%
    \begin{minipage}[c]{0.44\textwidth}
        \centering
        \begin{tikzpicture}

        \begin{axis}[
            width=0.95\linewidth,
            height=4.6cm,
            ybar,
            bar width=11pt,
            ylabel={Rate},
            ylabel style={font=\small},
            ymin=0, ymax=1.08,
            ytick={0,0.5,1.0},
            y tick label style={font=\small},
            xtick={1,2},
            xticklabels={Identification,Recovery},
            x tick label style={font=\scriptsize\bfseries},
            xmin=0.45, xmax=2.55,
            axis line style={gray!50},
            tick style={gray!50},
            ymajorgrids=true,
            grid style={gray!12},
            clip=false,
        ]

        \addplot[
            fill=correctbluefill,
            draw=correctblue,
            line width=0.6pt,
            bar shift=-8pt,
            nodes near coords={\pgfmathprintnumber[precision=2]{\pgfplotspointmeta}},
            point meta=y,
            every node near coord/.append style={
                font=\scriptsize\bfseries,
                text=black!70,
                anchor=south,
                yshift=1pt
            }
        ] coordinates {
            (1,\GoodCorrectMean)
            (2,\PartTwoGoodCorrectMean)
        };

        \addplot[
            fill=incorrectredfill,
            draw=incorrectred,
            line width=0.6pt,
            bar shift=8pt,
            nodes near coords={\pgfmathprintnumber[precision=2]{\pgfplotspointmeta}},
            point meta=y,
            every node near coord/.append style={
                font=\scriptsize\bfseries,
                text=black!70,
                anchor=south,
                yshift=1pt
            }
        ] coordinates {
            (1,\BadIdentification)
            (2,\PartTwoBadCorrectMean)
        };

        \draw[
            decorate,
            decoration={brace, amplitude=3pt, mirror},
            thick,
            black
        ]
        ([xshift=20pt]axis cs:2,\BraceLow) --
        ([xshift=20pt]axis cs:2,\BraceHigh)
        node[
            midway,
            right=3pt,
            font=\scriptsize\bfseries,
            text=black
        ]
        {$\Delta=\pgfmathprintnumber[precision=2]{\DeltaRecovery}$};

        \end{axis}
        \end{tikzpicture}

        \vspace{0.15em}
        {\scriptsize
        \tikz{\draw[fill=correctbluefill, draw=correctblue] (0,0) rectangle (0.12,0.12);}
        Correct predictions
        \quad
        \tikz{\draw[fill=incorrectredfill, draw=incorrectred] (0,0) rectangle (0.12,0.12);}
        Incorrect predictions
        }

        \vspace{0.2em}
        \textbf{\small (a) Task-based}
    \end{minipage}%
    \begin{minipage}[c]{0.50\textwidth}
        \centering
        \begin{tikzpicture}
        \begin{axis}[
            width=0.87\linewidth,
            height=5.8cm,
            xmin=1, xmax=7.2,
            xtick={1,2,3,4,5,6,7},
            xticklabels={1,2,3,\textbf{4},5,6,7},
            xlabel={Likert},
            xlabel style={font=\scriptsize},
            x tick label style={font=\scriptsize},
            ymin=-0.8, ymax=14,
            ytick={0,1.25,2.5,4.5,5.75,7,8.75,10,11.25,12.5},
            yticklabels={
                {\scriptsize User-friendly},
                {\scriptsize Cluttered$^\dagger$},
                {\scriptsize Easy to interpret},
                {\scriptsize Double-check},
                {\scriptsize Rely when strong},
                {\scriptsize Improves trust},
                {\scriptsize Helps verify},
                {\scriptsize Matches expected},
                {\scriptsize Errors apparent},
                {\scriptsize Relevant evidence}
            },
            y tick label style={font=\scriptsize, text width=2.2cm, align=right},
            axis line style={gray!50},
            tick style={gray!50},
            xmajorgrids=true,
            ymajorgrids=true,
            grid style={gray!45, dotted, line width=0.35pt},
            extra x ticks={4},
            extra x tick labels={},
            extra x tick style={grid style={black!45, dashed, line width=0.7pt}},
            clip=false,
        ]

        \fill[faithcolor!5] (axis cs:1,8.15) rectangle (axis cs:7,13.1);
        \fill[trustcolor!5] (axis cs:1,3.85) rectangle (axis cs:7,7.55);
        \fill[usecolor!5] (axis cs:1,-0.45) rectangle (axis cs:7,3.05);

        \node[font=\tiny\bfseries, text=faithcolor!70, anchor=north west] at (axis cs:1.05,12.9) {FAITHFULNESS};
        \node[font=\tiny\bfseries, text=trustcolor!70, anchor=north west] at (axis cs:1.05,7.35) {TRUST};
        \node[font=\tiny\bfseries, text=usecolor!70, anchor=north west] at (axis cs:1.05,2.85) {USABILITY};

        \node[
            font=\tiny,
            fill=white,
            inner sep=1pt,
            anchor=south,
            yshift=2pt
        ] at (axis cs:4,13.1) {Neutral};

        \addplot[
            only marks, mark=*, mark size=1.7pt, color=faithcolor,
            error bars/.cd, x dir=both, x explicit,
            error bar style={line width=0.5pt, faithcolor!35},
            error mark options={rotate=90, mark size=1.4pt, faithcolor!35},
        ] coordinates {
            (\FAMean,12.5)+-(\FAStd,0)
            (\FBMean,11.25)+-(\FBStd,0)
            (\FDMean,10)+-(\FDStd,0)
            (\FEMean,8.75)+-(\FEStd,0)
        };

        \addplot[
            only marks, mark=*, mark size=1.7pt, color=trustcolor,
            error bars/.cd, x dir=both, x explicit,
            error bar style={line width=0.5pt, trustcolor!35},
            error mark options={rotate=90, mark size=1.4pt, trustcolor!35},
        ] coordinates {
            (\TAMean,7)+-(\TAStd,0)
            (\TBMean,5.75)+-(\TBStd,0)
            (\TCMean,4.5)+-(\TCStd,0)
        };

        \addplot[
            only marks, mark=*, mark size=1.7pt, color=usecolor,
            error bars/.cd, x dir=both, x explicit,
            error bar style={line width=0.5pt, usecolor!35},
            error mark options={rotate=90, mark size=1.4pt, usecolor!35},
        ] coordinates {
            (\UAMean,2.5)+-(\UAStd,0)
            (\UBMean,1.25)+-(\UBStd,0)
            (\UCMean,0)+-(\UCStd,0)
        };

        \end{axis}
        \end{tikzpicture}

        \vspace{0.1em}
        \textbf{\small (b) Questionnaire}
    \end{minipage}%
    }
    \vspace{-0.2em}
    \caption{\textbf{User evaluation results.} (a) Identification and answer recovery rates for correct and incorrect model predictions. (b) Perceived faithfulness, trust, and usability (7-point Likert). $^\dagger$ denotes a negatively coded statement. Explanations enable participants to distinguish correct from incorrect predictions and recover the model's answer, while perceived faithfulness and usability score favourably.}
   \label{fig:user_eval_combined}
\end{figure}
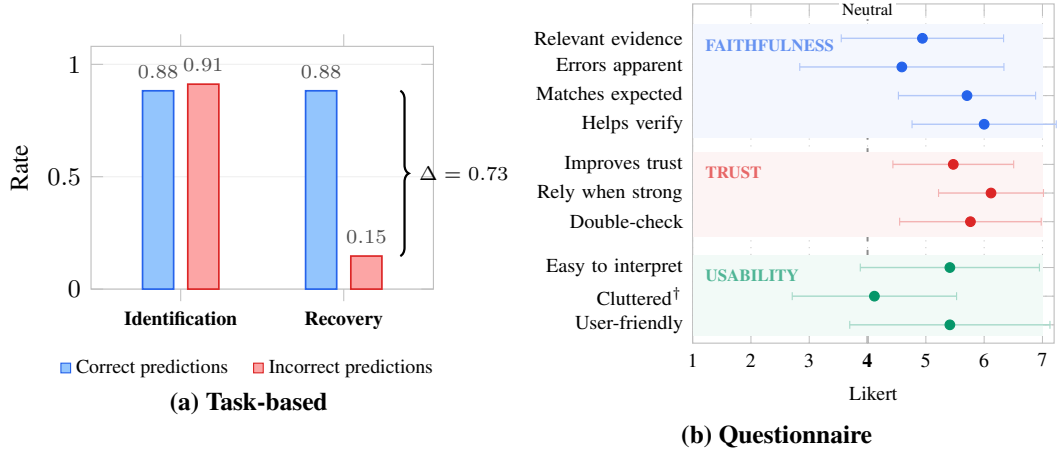

\FloatBarrier
\section{Conclusion}

\ModelName{} introduces a self-explainable DocVQA framework that makes the model’s evidence explicit at two levels: a question-relevance heatmap over document patches and a predicted answer region that can be directly inspected and evaluated. Our experiments demonstrate three key findings. First, \ModelName{} achieves SotA performance among explainable methods on PFL-DocVQA, closing the gap to non-explainable baselines while using significantly fewer input patches (Section~\ref{sec:baselines}, Table~\ref{tab:baseline_comparison}). Second, the model's question heatmap is faithful to relevant evidence and robust to the masking of task-irrelevant patches (Section~\ref{sec:faitfulness_and_robustness_eval}, Tables~\ref{tab:results_faithfulness_robustness_patches_only} and~\ref{tab:results_faithfulness_robustness_patches_and_embeddings}). The predicted textual answer is grounded in the last predicted explanation. Third, our user evaluation confirms that the explanations are actionable, which aids both error identification and answer verification (Section~\ref{sec:user_evaluation}, Figure~\ref{fig:user_eval_combined}). Taken together, these results support restricting the decoder to visually grounded evidence as a practical path toward more transparent and verifiable document question answering.

\section*{Limitations}

Answer localization relies on OCR-derived pseudo-boxes for training supervision. A manual audit of 200 validation examples found 86\% to be correct (Appendix~\ref{app:answer_priors}, Table~\ref{tab:val-audit-30}), and retraining with 300 human-annotated answer boxes yields an improvement of +0.05 ANLS (Table~\ref{tab:human_annotation}), suggesting that annotation quality has a limited effect at this scale. Whether these gains would compound with full-scale human annotation remains an open question. However, the result indicates that the architectural bottleneck of answer localization is a more immediate limiting factor than supervision noise alone. Our results on PFL-DocVQA show that performance scales substantially with dataset size. Since self-explainable DocVQA with spatial grounding is a novel task, no existing dataset provides native answer-region annotations or question-relevancy heatmaps, and adapting current benchmarks requires the pseudo-label pipeline we describe.

We foresee no direct negative social impact from this work, as improving model transparency is broadly aligned with responsible AI practices. The most plausible concern, over-trust in explanations, is not supported by our user evaluation (Appendix \ref{app:user_evaluation}).

\newpage

\begin{ack}

This project was completed as part of an MSc thesis conducted at the University of Oslo \cite{indrehus2026thesis}. 

The authors would like to thank the INTEGREAT and TRUST centres for 
valuable feedback on earlier versions of this work. Computational resources were provided by the Digital Signal Processing  and Image Analysis (DSB) research group at the University of Oslo, and the Norwegian Research Infrastructure Services (NRIS).

This work was supported by the Research Council of Norway through the FRIPRO Grant under project number 356103 and its Centres of Excellence scheme, Integreat - Norwegian Centre for knowledge-driven machine learning under project number 332645.

\end{ack}

\bibliography{ref}
\bibliographystyle{plain}

\newpage
\appendix
\onecolumn

\section*{\Large Appendix}

The appendix contains the following supplementary material:

\begin{itemize}[leftmargin=*, itemsep=0.1em]
  \item Appendix~\ref{app:extended_related_work}: \textbf{Detailed Comparison with Related Approaches.} Extended literature review comparing \ModelName{}  against related work across visually grounded reasoning, text-aware VQA, visual chain-of-thought, and self-explainable models, with a structured comparison table.
  \item Appendix~\ref{app:answer_priors}: \textbf{Generating Answer Location Priors.} Procedure for generating OCR-based answer bounding boxes used as supervision. We provide OCR-based audit, and additionally evaluate the effect on human annotation priors. 
  \item Appendix~\ref{app:question_priors}: \textbf{Generating Question Priors.} Construction of question-relevance heatmap priors, comparison of three prior sources, and quantitative evaluation with and without post-processing.
  \item Appendix~\ref{app:architecture_details}: \textbf{Architecture Details.} Additional model details, including projector modules, gating mechanisms, configuration presets, and decoder conditioning with predicted answer locations.
  \item Appendix~\ref{app:training_details}: \textbf{Training Details.} Training setup, optimization choices, and the full hyperparameter sweep table for DocVQA and PFL-DocVQA.
  \item Appendix~\ref{app:backbone_abl_study}: \textbf{Backbone Ablation Study.} Additional training with Donut-base (200M) and Pix2Struct-Large (1.3B) backbone models. Training was conducted on DocVQA and PFl-DocVQA. 
  \item Appendix~\ref{app:ar_examples}: \textbf{Qualitative Analysis of Area Ratio.} Examples illustrating the effect of predicting larger answer regions, and a grouped performance analysis linking explanation quality to prediction accuracy.
  \item Appendix~\ref{app:faithfulness_details}: \textbf{Faithfulness and Robustness Experiment Details.} Full protocol for the experiments in Section~\ref{sec:faitfulness_and_robustness_eval}, including visual examples of masking-based interventions on the question heatmap.
  \item Appendix~\ref{app:analysis_of_predictions}: \textbf{Qualitative Analysis .} Example model output including question mask, predicted answer box, and decoded answers. We also show representative decoding failures and localization errors.
  \item Appendix~\ref{app:user_evaluation}: \textbf{Qualitative User Evaluation. } A structured evaluation with \UserEvalParticipants{} participants (204 evaluations) assessing whether \ModelName's explanations are actionable, enable verification of predictions, and make model failures apparent. We include the instructions given to participants, and all results. 
  
\end{itemize}

\clearpage

\section{Detailed Comparison with Related Approaches}
\label{app:extended_related_work}

We extend the literature review from Section~\ref{sec:related_work} to 
provide a comprehensive comparison between \ModelName{} and related 
approaches. We organize the discussion along four lines of work and 
summarize key distinctions. First, we compare against work focusing on localization within natural-image VQA in Appendix \ref{sec:loc_natural_images_vqa_compare}. Second, we compare the text-aware VQA mechanism in Appendix \ref{sec:text_aware_vqa_compare}. Lastly, we compare explainability within DocVQA in Appendix \ref{sec:exp_and_doc_vqa_compare}. 

\subsection{Localization in Natural-Image VQA}
\label{sec:loc_natural_images_vqa_compare}

Using spatial localization as an intermediate reasoning step has been widely explored in natural-image VQA. We review representative approaches and contrast them with \ModelName{} in Table~\ref{tab:comp_nat_vqa}.

\ModelName{} differs from these approaches by enforcing faithfulness by construction. Prior localisation-based VQA methods share the intuition of focusing on relevant regions before answering. They differ from \ModelName{} in several key respects. Hard spatial selection~\cite{Malinowski_2018_ECCV} discards unselected feature-map cells but operates on coarse grid cells without producing an interpretable bounding box. LTG~\cite{zhu2023locate} adopt an explicit locate-then-answer pipeline with bounding boxes, yet does not constrain the decoder to use only the localized region. Visual CoT~\cite{shao2024vcot} goes further by cropping the image to the predicted box before answering, which does restrict downstream information flow. Shikra~\cite{chen2023shikraunleashingmultimodalllms} and Kosmos-2~\cite{peng2023kosmos2groundingmultimodallarge} embed spatial coordinates into the generation process, but the grounding is a byproduct of autoregressive decoding rather than an architectural bottleneck. Furthermore, all listed methods target natural images or scene text, where reasoning centres on object recognition rather than the layout-sensitive, text-rich understanding required in document VQA. None produce a dual rationale that separates question-evidence localization from answer grounding.

\begin{table}[t!]
    \centering
    \caption{Comparison of localization-centric VQA/grounding methods. ``Faithful by construction" indicates localization is a hard bottleneck for downstream answering. ``Verifiable box output" denotes explicit coordinate boxes that can be directly evaluated.}
    \resizebox{\textwidth}{!}{
    \begin{tabular}{lcccccc}
        \toprule
        Method & Domain & Localization & Faithful by & Dual & Verifiable \\
               &        & Type         & Construction & Rationale & Box Output \\
        \midrule
        HAN \cite{Malinowski_2018_ECCV} & Natural & Hard spatial selection        & \cmark & \xmark & \xmark \\ %
        LTG \cite{zhu2023locate}       & Scene Text & Locate then generate      & Partial & \xmark & \cmark \\ %
        Shikra \cite{chen2023shikraunleashingmultimodalllms}  & Natural & Coordinate I/O  & \xmark & \xmark & \cmark \\ %
        Kosmos-2 \cite{peng2023kosmos2groundingmultimodallarge}       & Natural & Location tokens  & \xmark & \xmark & \cmark \\ %
        Visual Sketchpad \cite{hu2024visualsketchpad}       & Natural & Sketching with tools   & \xmark & \xmark & \xmark \\ %
        VisCoT \cite{shao2024vcot}       & Natural & Box, crop, answer & \cmark & \xmark & \cmark \\ %
        Pixel Reasoner \cite{wang2025pixelreasonerincentivizingpixelspace}   & Natural & Zoom-in / select-frame & \xmark & \xmark & \xmark \\ %
        \midrule
        \ModelName{} (Ours)      & Document & Gated visual selection       & \cmark & \cmark & \cmark \\
        \bottomrule
    \end{tabular}}
    \label{tab:comp_nat_vqa}
\end{table}

\subsection{Text-Aware VQA}
\label{sec:text_aware_vqa_compare}

Text-aware VQA methods aim to answer questions that require reading and reasoning over text embedded in images, spanning both natural scene text and structured documents. Early approaches rely on OCR pipelines to extract text tokens, which are then consumed by multimodal transformers. A common design pattern is the copy mechanism, where the decoder can directly copy OCR tokens into the answer rather than generating from a fixed vocabulary. More recent OCR-free models bypass explicit text extraction entirely, learning to decode answers directly from the image. We review representative methods across this spectrum and contrast them with \ModelName{} in Table~\ref{tab:text_vqa_compare}.

M4C~\cite{hu2020iterative} fuses OCR tokens, visual objects, and the question through a multimodal transformer and decodes the answer via a multi-step pointer network that selects from either a fixed vocabulary or the detected OCR tokens. TAP~\cite{yang2021tap} extends this line by pre-training on large-scale OCR-aware data, improving text-image alignment without an explicit copy mechanism at inference. LayoutLMv3~\cite{huang2022layoutlmv3} jointly pre-trains on text, layout, and image modalities using masked language and region modeling, encoding OCR tokens with their spatial positions for document understanding tasks. Donut~\cite{kim2022ocrfreedocumentunderstandingtransformer} and Pix2Struct~\cite{lee2023pix2structscreenshotparsingpretraining} completely eliminate the OCR dependency, training end-to-end image-to-text models that generate answers directly from document images.

\ModelName{} shares the OCR-free, generative design of Donut and Pix2Struct but differs fundamentally in its self-explainability. While copy-based methods like M4C provide an indirect link between copied tokens and their source locations, this link is a decoding strategy rather than an architectural constraint. The model is not required to localize the evidence region before answering. Similarly, OCR-free baselines treat the full image as input without isolating the relevant region. In contrast, \ModelName{} architecturally restricts the decoder to reason only from the gated evidence region, producing a verifiable bounding box as an explicit intermediate output. None of the methods in this category provide self-explainable predictions.

\begin{table}[ht!]
    \centering
    \small
    \caption{Comparison of text-aware VQA models spanning scene-text VQA and DocQA. \emph{OCR-based} indicates OCR tokens are explicit model inputs. \emph{Copy mechanism} denotes a pointer/copy decoder over OCR tokens. \emph{Generative} denotes open-vocabulary decoding rather than classification/extraction.}
    \begin{tabular}{l c c c c c}
        \toprule
        Method & DocVQA & OCR-based & Copy Mechanism & Generative & Self-Explainable \\
        \midrule
        M4C \cite{hu2020iterative}              & \xmark & \cmark & \cmark & \cmark & \xmark \\
        TAP \cite{yang2021tap}                 & \xmark & \cmark & \xmark & \cmark & \xmark \\
        LayoutLMv3 \cite{huang2022layoutlmv3}  & \cmark & \cmark & \xmark & \xmark & \xmark \\
        Donut \cite{kim2022ocrfreedocumentunderstandingtransformer}              & \cmark & \xmark & \xmark & \cmark & \xmark \\
        Pix2Struct \cite{lee2023pix2structscreenshotparsingpretraining}    & \cmark & \xmark & \xmark & \cmark & \xmark \\
        \midrule
        \ModelName{} (Ours)                          & \cmark & \xmark & \xmark & \cmark & \cmark \\
        \bottomrule
    \end{tabular}
    \label{tab:text_vqa_compare}
\end{table}

\subsection{Explainability in Multimodal and Document VQA}
\label{sec:exp_and_doc_vqa_compare}

Explainability in VQA includes methods, such as post-hoc attribution applied to opaque models, and inherently interpretable architectures that produce explanations as part of their prediction process. We organize this section in three directions: (1) post-hoc explanation methods, (2) multimodal explanation models, and (3) self-explainable document VQA systems. We contrast representative approaches with \ModelName{} in Table~\ref{tab:explain_vqa_compare}.

Post-hoc methods such as LIME~\cite{ribeiro2016should} and SHAP~\cite{lundberg2018shap} generate input-level feature attributions for arbitrary models and are model-agnostic. They offer no guarantee that the highlighted features reflect the model's actual reasoning process. Grad-CAM~\cite{Selvaraju_2019} extends gradient-based attribution to convolutional networks by producing visual saliency maps that highlight discriminative image regions. These maps are post-hoc projections of gradient flow rather than architectural constraints on reasoning. PJ-X~\cite{park2018multimodal} goes beyond attribution by generating multimodal explanations with both textual justifications and visual pointing. PJ-X operates on natural images and does not enforce that the explanation constrains the answer.

LayoutXLM~\cite{xu2021layoutxlmmultimodalpretrainingmultilingual}, Donut~\cite{kim2022ocrfreedocumentunderstandingtransformer}, and Pix2Struct~\cite{lee2023pix2structscreenshotparsingpretraining} achieve strong DocVQA performance but are fully opaque. They provide no mechanism for localizing or explaining the evidence used to generate an answer. LaBo~\cite{yang2023languagebottlelanguagemodel} introduces a concept bottleneck based on the language that produces intermediate interpretable representations but is designed for image classification rather than VQA or document understanding. NOTICE~\cite{golovanevsky2025vlmsnoticemechanisticinterpretability} provides mechanistic interpretability analysis of vision-language models, offering insight into internal representations but not producing user-facing explanations at inference time.

The most closely related work is DocVXQA~\cite{souibgui2025docvxqacontextawarevisualexplanations}, which generates context-aware visual explanations for DocVQA by highlighting relevant document regions. DocVXQA is self-explainable and operates in the document domain but produces a single explanation rationale.

\ModelName{} is, to the best of our knowledge, the only method that combines all four properties. It operates on document images, processes multimodal inputs, is self-explainable by construction, and produces two visual explanations in the form of both a question-evidence heatmap and an answer-grounding bounding box. \ModelName{} does not require a post-hoc method to offer explanation. We instead restrict the architecture to produce explanations that the model is faithful to the model's internal reasoning process. The decoder cannot access information outside the selected region, making the dual rationale a faithful reflection of the model's reasoning process.

\begin{table}[ht!]
    \centering
    \small
    \caption{Comparison of post-hoc explanation methods, multimodal explanation models, and DocVQA architectures. \emph{DocVQA} indicates the method is designed for document understanding. \emph{Multimodal} indicates the model processes more than one input modality. \emph{Self-Explainable} indicates explanations are produced as part of the model's inference rather than applied post-hoc. \emph{Dual Visual Explanation} indicates the model produces two 
complementary spatial rationales (e.g.\ evidence localization and answer grounding).}
    \begin{tabular}{l c c c c}
        \toprule
        Method & DocVQA & Multimodal & Self-Explainable & Dual Visual Expl. \\
        \midrule
        LIME \cite{ribeiro2016should}          & \xmark & \xmark & \xmark & \xmark \\
        SHAP \cite{lundberg2018shap}        & \xmark & \xmark & \xmark & \xmark \\
        Grad-CAM \cite{Selvaraju_2019}      & \xmark & \cmark & \xmark & \xmark \\
        
        PJ-X \cite{park2018multimodal}         & \xmark & \cmark & \xmark & \cmark \\
        
        LayoutXLM \cite{xu2021layoutxlmmultimodalpretrainingmultilingual}       & \cmark & \cmark & \xmark & \xmark \\
        Donut \cite{kim2022ocrfreedocumentunderstandingtransformer}              & \cmark & \cmark & \xmark & \xmark \\
        Pix2Struct \cite{lee2023pix2structscreenshotparsingpretraining}    & \cmark & \cmark & \xmark & \xmark \\
        
        LaBo \cite{yang2023languagebottlelanguagemodel}           & \xmark & \xmark & \cmark & \xmark \\
        NOTICE \cite{golovanevsky2025vlmsnoticemechanisticinterpretability}   & \xmark & \cmark & \xmark & \xmark \\
        
        DocVXQA \cite{souibgui2025docvxqacontextawarevisualexplanations}     & \cmark & \cmark & \cmark & \xmark \\
        \midrule
        \ModelName{} (Ours)                           & \cmark & \cmark & \cmark & \cmark \\
        \bottomrule
    \end{tabular}
    \label{tab:explain_vqa_compare}
    
\end{table}

\FloatBarrier
\section{Generating Answer Location Priors}
\label{app:answer_priors}

We compute the answer prior to obtain a bounding box that covers the ground-truth answer. We extend the original DocVQA dataset \cite{mathew2021docvqadatasetvqadocument} using the OCR provided by Amazon Textract\footnote{\url{https://aws.amazon.com/textract/}}. As a backup OCR engine, we propose using PaddleOCR \cite{du2020ppocrp, cui2025paddleocr} to handle handwritten texts. Although this does not substitute for full human annotation, it is a scalable and cost-effective alternative that yields cleaner answer priors than training with missing or unverified priors, and therefore provides a stronger supervision signal for downstream training. 

\paragraph{Text Matching.} We match the answer $a$ to OCR lines. OCR output is noisy and inconsistent. The same answer may appear with different punctuation, spacing, or minor character errors. Relying on a single exact string match can miss valid matches. We therefore use a small set of complementary matching rules that handle common OCR variations. First, we apply text normalization $\mathrm{norm}(\cdot)$\footnote{NFKC, lowercasing, and replacing non-alphanumeric characters with white-space.} to handle general text answers. For numeric answers, we remove all non-digits, making matching robust to OCR formatting differences such as commas, spaces, currency symbols or parentheses (e.g., “1,234.00” vs. “1234”). Digit-only matching reduces sensitivity to OCR formatting, so numerically equivalent strings match even when punctuation differs. We prioritize exact and substring matches under these transforms, and if none match, we fall back to fuzzy matching using Levenshtein similarity, defined as $\mathrm{sim}(a,b)=1-\frac{d_{\mathrm{lev}}(a,b)}{\max(|a|,|b|,1)}$, where $d_{\mathrm{lev}}$ is the Levenshtein edit distance \cite{levenshtein1966}. We use $\tau_{\text{text}}=0.82$ for $\mathrm{norm}(\cdot)$ because word-level OCR typically contains small character errors and a moderate threshold preserves recall. For digit-only strings we use a stricter threshold $\tau_{\text{dig}}=0.95$, since numeric strings are short and less redundant, and allowing larger deviations increases the risk of matching the wrong number.

\paragraph{Pipeline.}
\begin{enumerate}
    \item Use Amazon Textract OCR to collect line texts with bounding boxes.
    \item For each OCR line $t_i$, compute $\mathrm{norm}(t_i)$ and $\mathrm{dig}(t_i)$ and compare them to the answer $a$. We select the first matching line according to the priority: (1) exact match on $\mathrm{norm}$, (2) exact match on $\mathrm{dig}$, (3) substring match on $\mathrm{norm}$, (4) substring match on $\mathrm{dig}$, (5) fuzzy match on $\mathrm{norm}$ (score $\ge \tau_{\text{text}}$), and (6) fuzzy match on $\mathrm{dig}$ (score $\ge \tau_{\text{dig}}$).
    \item If no acceptable match is found, run PaddleOCR and repeat step 2. If this also fails, set the prior to \texttt{None}.
    \item Convert the selected box to normalized coordinates $[x_1,y_1,x_2,y_2]\in[0,1]^4$, expand it by $+10\%$ in $x$ and $+15\%$ in $y$, and clip to $[0,1]$.
\end{enumerate}

To better understand how the pipeline behaves in practice, we record which rule in Step 2 produced the selected match and report the distribution of match reasons per OCR engine. Table \ref{tab:reason-dist-textract} shows the Amazon Textract OCR answer-match distribution, and Table \ref{tab:reason-dist-paddleocr} for PaddleOCR answer-match distribution. Both OCR engines used mostly exact matches and substring contributions. This suggests that the pipeline typically succeeds without resorting to aggressive fuzziness, and that digits-only matching is particularly helpful for numeric fields. 

\begin{figure}[ht!]
    \centering
    \caption[Reason distribution by OCR engine]{
    \textbf{Reason distribution by OCR engine.}
    Distribution of selected match reasons for \textit{(Left)} PaddleOCR and \textit{(Right)} Amazon Textract OCR on train and validation splits. Each table shows the distribution of the used answer-match methods against each other.}

    \vspace{2mm}

    \begin{subfigure}[t]{0.48\textwidth}
        \centering
        \small
        \begin{tabular}{lcc}
            \toprule
            Reason & Train & Validation \\
            \midrule
            exact\_digits     & 147 (18.1\%) & 21 (18.6\%) \\
            exact\_norm       & 431 (53.1\%) & 67 (59.3\%) \\
            fuzzy\_norm       & 101 (12.4\%) & 15 (13.3\%) \\
            substring\_digits & 19 (2.3\%)   & 1 (0.9\%)   \\
            substring\_norm   & 114 (14.0\%) & 9 (8.0\%)   \\
            \midrule
            \textbf{TOTAL}    & 812 (100\%)  & 113 (100\%) \\
            \bottomrule
        \end{tabular}
        \caption{PaddleOCR.}
        \label{tab:reason-dist-paddleocr}
    \end{subfigure}
    \hfill
    \begin{subfigure}[t]{0.48\textwidth}
        \centering
        \small
        \begin{tabular}{lcc}
            \toprule
            Reason & Train & Validation \\
            \midrule
            exact\_digits     & 4,881 (13.4\%)  & 631 (13.0\%) \\
            exact\_norm       & 23,806 (65.5\%) & 2,969 (61.3\%) \\
            fuzzy\_norm       & 723 (2.0\%)     & 82 (1.7\%) \\
            substring\_digits & 183 (0.5\%)     & 22 (0.5\%) \\
            substring\_norm   & 6,777 (18.6\%)  & 1,138 (23.5\%) \\
            \midrule
            \textbf{TOTAL}    & 36,370 (100\%)  & 4,842 (100\%) \\
            \bottomrule
        \end{tabular}
        \caption{Amazon Textract OCR.}
        \label{tab:reason-dist-textract}
    \end{subfigure}

    \label{fig:reason-dist-by-ocr}
\end{figure}

\begin{figure}[ht!]
  \centering
  \begin{subfigure}[t]{0.32\textwidth}
    \centering
    \includegraphics[width=\linewidth]{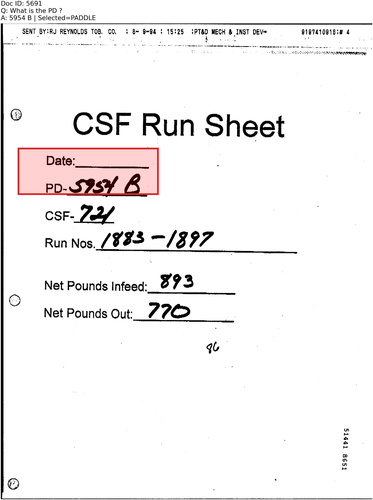}
    \caption{Paddle correctly finds the location of the answer. The predicted location cuts off the bottom of the letter ``B". The example was marked as ``partial". }
    \label{fig:paddle_a}
  \end{subfigure}\hfill
  \begin{subfigure}[t]{0.32\textwidth}
    \centering
    \includegraphics[width=\linewidth]{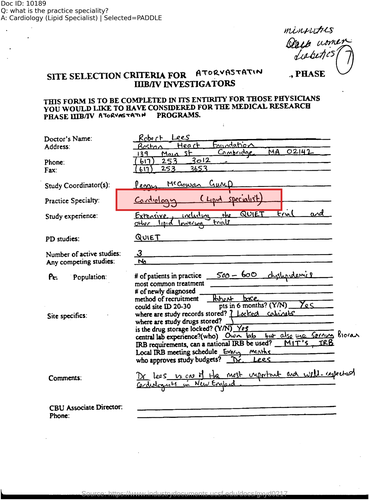}
    \caption{Paddle finds the correct location, and no text is left outside the bounding box.}
    \label{fig:paddle_b}
  \end{subfigure}\hfill
  \begin{subfigure}[t]{0.32\textwidth}
    \centering
    \includegraphics[width=\linewidth]{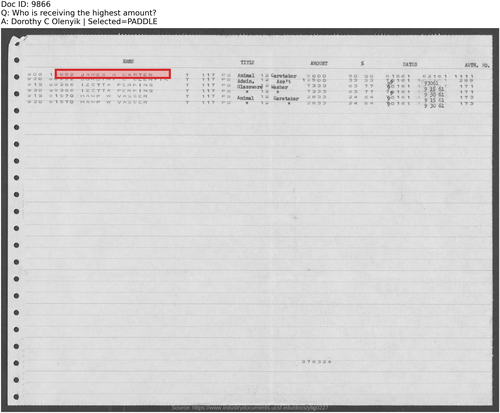}
    \caption{Paddle failed to locate the correct location. The correct location is on the line below. Text quality of the image is very low. }
    \label{fig:paddle_c}
  \end{subfigure}

  \caption{Paddle examples where the red bounding box denotes the predicted location. Document information is appended as a header to the document (ID, Question, Answer, Selected method for answer location).}
  \label{fig:paddle_examples}
\end{figure}

\begin{figure}[ht!]
  \centering
  \begin{subfigure}[t]{0.32\textwidth}
    \centering
    \includegraphics[width=\linewidth]{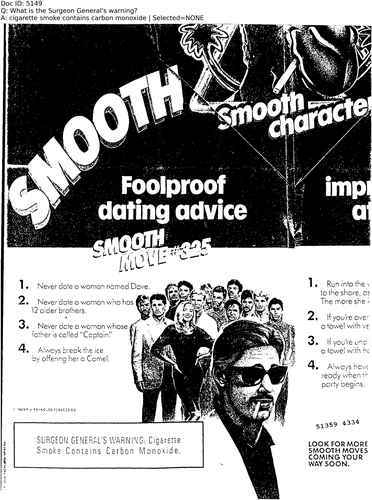}
    \caption{Correct location is in the box below. The text is missing connections, and would require manual annotation to locate.}
    \label{fig:none_a}
  \end{subfigure}\hfill
  \begin{subfigure}[t]{0.32\textwidth}
    \centering
    \includegraphics[width=\linewidth]{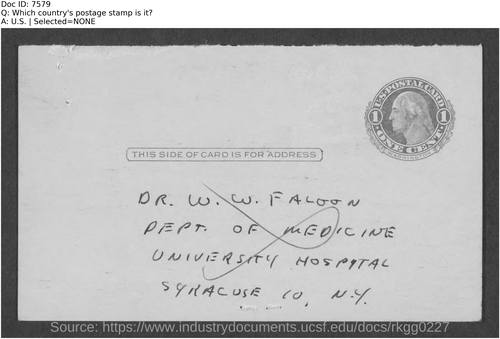}
    \caption{Correct answer is ``U.S.". This information is only available on the stamp, around the edge.}
    \label{fig:none_b}
  \end{subfigure}\hfill
  \begin{subfigure}[t]{0.32\textwidth}
    \centering
    \includegraphics[width=\linewidth]{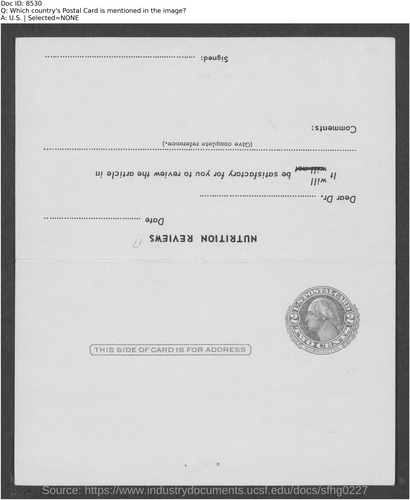}
    \caption{Another example of a letter where the correct answer location is on the stamp. Here, majority of the text content is upside down.}
    \label{fig:none_c}
  \end{subfigure}

  \caption{Examples where the pipeline the failed to locate the answer location. Document information is appended as a header to the document (ID, Question, Answer, Selected method for answer location). }
  \label{fig:none_examples}
\end{figure}

\paragraph{OCR utilization and audit.} The distribution of the actual OCR engine used is shown in Table \ref{tab:ocr-source-dist}. In case both OCR engines are unable to find a match for the ground truth answer, it is marked as ``None" and no answer bounding box is generated. We filter out such examples for our training setup due to not providing any supervision for our task. To validate this approach, we do a quantitative sanity check of the results by randomly choosing 200 examples from the validation split and validate how correct they are. Table \ref{tab:val-audit-30} shows the validation results with about 86\% of the examples being acceptable. We observed that the failed cases were often due to low quality document images and unreadable texts. Amazon Textract OCR was able to correctly predict the location in the majority of the examples. Figure \ref{fig:paddle_examples} shows examples where Paddle OCR obtained the location of the answer. Figure \ref{fig:none_examples} shows some of the sampled cases where the answer location pipeline was unable to find the answer location. 

\begin{figure}[ht]
    \centering

    \caption[OCR utilization and audit]{
    \textbf{OCR utilization and audit} \textit{(Left)} Distribution of which OCR engine produced an acceptable match for answer localization in the train and validation splits. \textit{(Right)} Quantitative outcome of a manual visual audit of 200 randomly sampled validation examples, labelled as \emph{Correct} (answer fully inside the predicted box), \emph{Partial} (answer inside but box misses some tokens), or \emph{Incorrect} (answer not covered or wrong region).}

    \vspace{2mm}
    
    \begin{subfigure}[t]{0.48\textwidth}
        \centering
        \small
        \begin{tabular}{lcc}
            \toprule
            \textbf{Source} & \textbf{Train} & \textbf{Validation} \\
            \midrule
            Textract  & 92.2\% & 90.5\% \\
            PaddleOCR & 2.1\%  & 2.1\%  \\
            None      & 5.8\%  & 7.4\%  \\
            \bottomrule
        \end{tabular}
        \caption{OCR source used to produce answer priors.}
        \label{tab:ocr-source-dist}
    \end{subfigure}
    \hfill
    \begin{subfigure}[t]{0.48\textwidth}
        \centering
        \small
        \begin{tabular}{lccc}
            \toprule
            \textbf{Source} & \textbf{Correct} & \textbf{Partial} & \textbf{Incorrect} \\
            \midrule
            Textract & 168 & 0 & 9 \\
            PaddleOCR & 2 & 2 & 5 \\
            None & 0 & 0 & 14 \\
            \bottomrule
        \end{tabular}
        \caption{Visual audit on 200 validation examples.}
        \label{tab:val-audit-30}
    \end{subfigure}
    \label{fig:answer_prior_table}
\end{figure}

\FloatBarrier
\paragraph{Effect of human annotation priors.} In addition to the OCR audit, we manually annotated 300 examples in the DocVQA examples to measure the performance change under re-training manually annotating examples. We annotate 200 examples in the train split and 100 examples in the validation split. Table \ref{tab:human_annotation} shows that replacing OCR-based priors with manual annotations under the same settings and seed yields only marginal performance gains (+0.04 ACC, +0.05 ANLS), indicating that supervision quality is not the primary bottleneck.

\begin{table}[ht!]
    \centering
    \caption{\textbf{Effect of human annotation priors}. We conduct training on the humanly annotated dataset (DocVQA, \cite{mathew2021docvqadatasetvqadocument}) under the same setting and seed. Marginal performance improvement was observed.}
    \begin{tabular}{ccc}
        \toprule
        \textbf{Human annotation} & \textbf{ACC} & \textbf{ANLS} \\
        \midrule
        \xmark & 0.34 & 0.43 \\
        \cmark & \textbf{0.38 (+0.04)} & \textbf{0.48 (+0.05)} \\
        \bottomrule
    \end{tabular}
    \label{tab:human_annotation}
\end{table}

\FloatBarrier
\section{Generating Question Priors}
\label{app:question_priors}

\paragraph{Compute Priors.} We propose to compute the question priors using two late-interaction retrievers (ColSmol-500M\footnote{\url{https://huggingface.co/vidore/colSmol-500M}}, ColQwen2.5\footnote{\url{https://huggingface.co/vidore/colqwen2.5-v0.2}}) and cross-attention-based salient maps from the pretrained Pix2Struct-base model\footnote{\url{https://huggingface.co/google/pix2struct-docvqa-base}}. For the late-interaction retrievers, we compute a token-level similarity matrix between all question tokens and image patches, then aggregate to a single relevance score per patch by taking the maximum across question tokens. The late-interaction retrievers operate on a fixed patch grid, whereas Pix2Struct uses a variable-resolution grid that adapts to each document's aspect ratio. To align the two, we first upsample the retriever's patch scores to pixel level using bicubic interpolation, then downsample to the backbone's patch grid (512 patches by default) using bilinear interpolation, where the target grid dimensions are computed from the image's aspect ratio and the configured patch budget.

\begin{figure}[ht!]
    \centering
    \includegraphics[width=\linewidth]{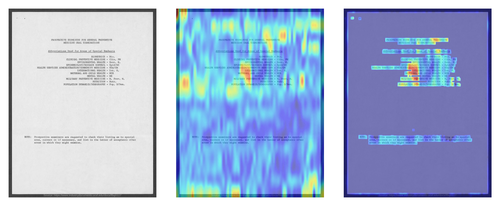}
    \caption{Question prior from ColSmol-500M: document (left), raw prior (middle), post-processed prior (right).}
    \label{fig:q_prior_colsmol}
\end{figure}

\begin{figure}[ht!]
    \centering
    \includegraphics[width=\linewidth]{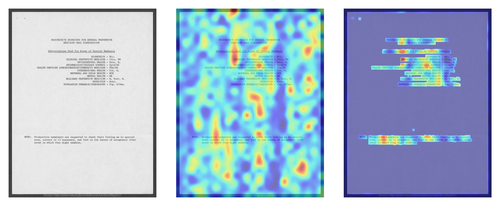}
    \caption{Question prior from ColQwen2.5: document (left), raw prior (middle), post-processed prior (right).}
    \label{fig:q_prior_colqwen}
\end{figure}

\begin{figure}[ht!]
    \centering
    \includegraphics[width=\linewidth]{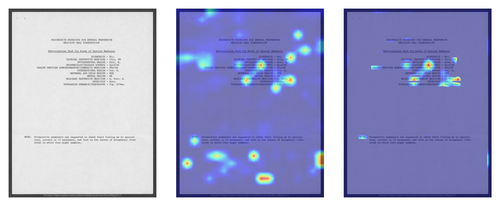}
    \caption{Question prior from Pix2Struct cross-attention: document (left), raw prior (middle), post-processed prior (right).}
    \label{fig:q_prior_pix2struct}
\end{figure}

For Pix2Struct, we propose using the cross-attention-based relevance signal and convert it to a patch-level map by aggregating attention across heads and layers. We apply a post-processing step before we save the question prior as patch level priors. Raw question priors are often noisy and highlight uninformative information for supervision. Post-processing consists of three steps. First, we compute a local appearance-variance score in 15×15 pixel windows and down-weight windows with low variance. Low-variance regions typically include uniform regions with less question-relevant information. Next, we apply spatial normalization to further suppress noise. In the final step, we suppress a thin border to avoid the high activation caused by the frame artefacts around the edge. We set the outer 7\% of the border to zero. We chose 7\% empirically: it removed most edge artefacts while preserving the content regions of the document. Figures \ref{fig:q_prior_colsmol}, \ref{fig:q_prior_colqwen} and \ref{fig:q_prior_pix2struct} show the original document (left), the raw question prior (middle) and the post-processed question priors (right), using the same validation example. 

\paragraph{Question Prior Evaluation.} We evaluated the different generated outputs based on two key perspectives: (1) whether the question prior places mass on the ground-truth answer region, and (2) how much irrelevant information the prior able to suppress. A question-relevant heatmap should be selective and highlight only a portion of the page. We emphasise that pure question-answer overlap alone are not a complete metric to evaluate question-relevancy. Question-relevant information can span multiple lines within the document. Therefore, we propose a set of overlap metrics between the question-answer as a lower-bound sanity check. To the best of our knowledge, there is no established protocol for evaluating question-relevance heatmaps. We also evaluate before and after post-processing ($\textbf{PP}$) to quantify the post-processing effect, and compare between prior sources. The metrics proposed are: soft Intersection of Union ($\mathrm{IoU}_{\text{soft}}$), precision at threshold $K$ ($P@K$), recall at threshold ($R@K$), sparsity $S$ and Jensen-Shannon Divergence ($JSD$). Here, $P@K$ and $R@K$ threshold the top-$K$ percentage of the question prior. $S$ measures the fraction of activation over a small threshold. Table \ref{tab:q-prior-metrics-result} shows the evaluation results for all models across the train and validation splits, with and without post-processing. 

\begin{table}[ht!]
  \centering
  \small
  \caption[Question prior evaluation metrics.]{Question prior evaluation metrics on train and validation splits with and without post-processing (PP). P@K = Precision@K, R@K = Recall@K, S = Sparsity, JSD = Jensen-Shannon Divergence. Here, K for precision and recall is set to 30\%. Sparsity assumes that any activation under 0.01 is negligible. Arrows indicate desired direction ($\uparrow$ higher is better, $\downarrow$ lower is better).}
  \label{tab:q-prior-metrics-result}
  \resizebox{\linewidth}{!}{
  \begin{tabular}{@{} l c c c c c c c c c c c c @{}}
    \toprule
    \multicolumn{1}{@{}l}{\textbf{Model}} &
    \multicolumn{6}{c}{\textbf{Train}} &
    \multicolumn{6}{c}{\textbf{Validation}} \\
    \cmidrule(lr){2-7} \cmidrule(l){8-13}
    & PP & $\mathrm{IoU}_{\text{soft}}\,\uparrow$ & $\mathrm{P@K}\,\uparrow$ & $\mathrm{R@K}\,\uparrow$ & $\mathrm{S}$ & $\mathrm{JSD}\,\downarrow$
    & PP & $\mathrm{IoU}_{\text{soft}}\,\uparrow$ & $\mathrm{P@K}\,\uparrow$ & $\mathrm{R@K}\,\uparrow$ & $\mathrm{S}$ & $\mathrm{JSD}\,\downarrow$ \\
    \midrule
    \multirow{2}{*}{\texttt{vidore/colqwen2.5-v0.2}} 
    & \xmark & 0.0062 & 0.0062 & 0.2969 & 0.0013 & 0.0792 & \xmark & 0.0069 & 0.0068 & 0.2918 & 0.0013 & 0.0800 \\
    & \cmark & 0.0145 & 0.0119 & 0.7442 & 0.5729 & 0.0864 & \cmark & 0.0158 & 0.0130 & 0.7359 & 0.5731 & 0.0871 \\
    \addlinespace[0.5em]
    \multirow{2}{*}{\texttt{vidore/colSmol-500M}} 
    & \xmark & 0.0071 & 0.0089 & 0.4159 & 0.0016 & 0.0791 & \xmark & 0.0078 & 0.0096 & 0.4082 & 0.0016 & 0.0800 \\
    & \cmark & 0.0164 & 0.0128 & 0.7843 & 0.5851 & 0.0863 & \cmark & 0.0176 & 0.0140 & 0.7726 & 0.5855 & 0.0871 \\
    \addlinespace[0.5em]
    \multirow{2}{*}{\texttt{google/pix2struct-docvqa-base}} 
    & \xmark & 0.0079 & 0.0104 & 0.4466 & 0.2580 & 0.0880 & \xmark & 0.0079 & 0.0108 & 0.4187 & 0.2411 & 0.0889 \\
    & \cmark & 0.0165 & 0.0065 & 0.9989 & 0.9366 & 0.1145 & \cmark & 0.0153 & 0.0073 & 0.9988 & 0.9337 & 0.1157 \\
    \bottomrule
  \end{tabular}}
\end{table}

\paragraph{Interpretation of the Prior Evaluation.} Pix2Struct showed the highest recall at the 30\% threshold (0.9989), which indicates the cross-attention based prior almost always highlights the answer ground truth region. This behaviour is expected because cross-attention is directly tied to highlighting tokens that are relevant for the decoder. However, our objective is a question-relevancy heatmap that highlights \textit{a broader relevancy region}. Between the evaluated question-prior sources, ColSmol-500M had the best trade-off between answer-coverage ( $R@30\% = 0.7843$) and a concentration of ($S =0.58$). In practice, ColSmol-500M reduces the number of patches activated by $\geq50\%$ while highlighting a high degree of answer-relevant regions. Thus, \textit{we used ColSmol-500M question priors to supervise the question projector}.

\section{Architecture Details}
\label{app:architecture_details}

Our method (Section \ref{sec:method}) describes the model design at a conceptual level. We provide more details about the model architecture and ablation settings in the following section. As shown in Figure \ref{fig:architecture}, our approach uses a Pix2Struct-base encoder-decoder backbone\footnote{\url{https://huggingface.co/google/pix2struct-docvqa-base}} together with (i) a question projector that predicts a relevance mask, (ii) an answer projector that predicts the answer box, (iii) a gating module that applies the mask to visual tokens, and (iv) a decoder conditioning strategy for answer generation. We provide component-level details and alternatives for the projectors (Appendix~\ref{app:projectors}), gating (Appendix~\ref{app:gating}), decoder conditioning (Appendix~\ref{app:decoder_input}), and preset configurations (Appendix~\ref{app:configurations}).

\FloatBarrier
\subsection{Projectors}
\label{app:projectors}

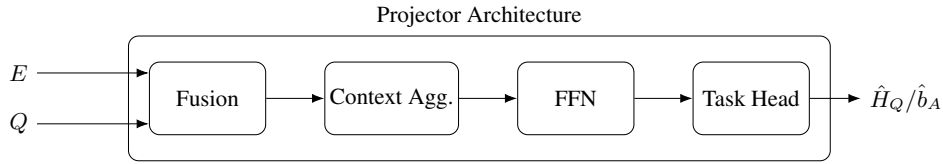
\begin{figure}[ht!]
    \centering
    \resizebox{0.9\linewidth}{!}{%
        \begin{tikzpicture}[>=Latex, node distance=8mm, font=\small]
          \tikzset{
            op/.style = {draw, rounded corners, minimum width=16mm, minimum height=10mm, inner sep=2pt, align=center}
          }
        
          \node (E) at (0,0.35) {$E$};
          \node (Q) at (0,-0.35) {$Q$};
        
          \node[op] (fusion) at (2.6,0) {Fusion};
          \node[op, right=8mm of fusion] (agg) {Context Agg.};
          \node[op, right=8mm of agg] (ffn) {FFN};
          \node[op, right=8mm of ffn] (head) {Task Head};
        
          \node[
            draw, rounded corners,
            fit=(fusion)(agg)(ffn)(head),
            inner xsep=8pt, inner ysep=10pt,
            label={above:Projector Architecture}
          ] (box) {};
        
          \draw[->] (E) -- (fusion.west |- E);
          \draw[->] (Q) -- (fusion.west |- Q);
          \draw[->] (fusion) -- (agg);
          \draw[->] (agg) -- (ffn);
          \draw[->] (ffn) -- (head);
          \draw[->] (head.east) -- ++(7mm,0) node[right] {$\QMask/\ABox$};
        \end{tikzpicture}%
    }
    \caption{End-to-end projector flow. Inputs $E, Q$ (embeddings) enter the Projector, and the output is the predicted question heatmap $\QMask$, or the predicted answer bounding box $\ABox$. The Fusion, Context Aggregation and Feed-forward Network (FFN) blocks keep the shape of the original embeddings $E$. The task-head reduces the prediction to answer localization bounding box, or keeps it to a per-patch level with the question projector. }
    \label{fig:projector}
\end{figure}

Both the answer projector and the question projector share the same modular design. The first stage is the \textit{Fusion block}. Here, we combine patch embeddings with question representation to get embeddings that have been fused with question-conditioned features. By default, Pix2Struct adds the question rendered as the header of the document image \cite{lee2023pix2structscreenshotparsingpretraining}. Since our framework is not restricted to a single backbone variant and requires such a representation for the Fusion block, we handle this for Pix2Struct by tokenizing the question using the model's own decoder embedding matrix, reusing existing parameters without introducing a separate text encoder. We use explicit question–image fusion to provide a stronger question-conditioned signal for localization. The embeddings are then fed through the \textit{Context Aggregation block}, which mixes information across patches to enable a wider document context. Next, we use a lightweight Feed Forward Network (FFN) before the task head. Both projector variants have their own task-head for predicting the question heatmap or the answer bounding box. For the question heatmap, we predict the relevancy per patch. For the answer bounding box, we predict the answer location (four values). Figure \ref{fig:projector} illustrates the overall flow within each projector. 

Each block is built on interchangeable blocks. During development, we implemented and tested multiple variants. Unless otherwise stated, we use FiLM \cite{perez2017filmvisualreasoninggeneral}, and cross-attention for the fusion block, transformer encoder for context aggregation. For the task-head, we use attention pooling before a lightweight MLP for the final task-specific prediction. 

\FloatBarrier
\subsection{Gating Mechanisms}
\label{app:gating}

We propose four different gating mechanisms that were used in our ablation study. All variants gate the image embedding using the predicted question mask before the features are passed to the answer projector. Let $\ImgEmb\in\mathbb{R}^{B\times N\times d}$ denote the image embeddings, where $B$ is the batch size, $N=H\!\cdot\!W$ the number of spatial tokens, and $d$ the embedding dimension. Let $\QMask\in\mathbb{R}^{B\times N\times 1}$ denote the predicted question mask. Each gating mechanism produces gated embeddings $\GatedImgEmb$ with the same shape as $\ImgEmb$.

\paragraph{Linear interpolation gating} applies a learnable global interpolation strength $\alpha$ to scale features according to the mask. In our implementation, $\alpha$ is a learned scalar initialized to a small value (default $0.1$), and the mask is applied multiplicatively:

\begin{equation}
    \GatedImgEmb = \ImgEmb\odot\big(\alpha\,\QMask + (1-\alpha)\big),
\end{equation}

where $\odot$ denotes element-wise multiplication and $\QMask$ is broadcast to $\mathbb{R}^{B\times N\times d}$.

\paragraph{Residual gating} learns a feature transformation network and uses the mask to interpolate between the original features and the transformed features \cite{he2015deepresiduallearningimage}. Specifically, a two-layer MLP $f_{\theta}:\mathbb{R}^{d}\rightarrow\mathbb{R}^{d}$, with GELU and dropout, produces $\mathbf{T}=f_{\theta}(\ImgEmb)$. The mask then controls how much of the transformation is applied:

\begin{equation}
    \GatedImgEmb = \QMask\odot \mathbf{T} + (1-\QMask)\odot \ImgEmb.
\end{equation}

\paragraph{Spatial attention gating} interprets the predicted mask as spatial attention weights and uses them to modulate the image embeddings, highlighting spatial regions that are relevant to the question. We first normalize the mask across tokens to obtain attention weights

\begin{equation}
    \mathbf{a} = \frac{\QMask}{\sum_{n=1}^{N}\QMaskN + \varepsilon},
\end{equation}

where the sum is taken over the token dimension and $\varepsilon$ is a small constant for numerical stability. With a learnable scalar $\alpha$, the gating is applied as:

\begin{equation}
    \GatedImgEmb = \ImgEmb\odot \big(1 + \alpha\,\mathbf{a}\big)
\end{equation}

Note that for typical sequence lengths ($N \geq 512$), the normalized weights are small, which limits the effective dynamic range of the gate. This likely contributes to the weaker performance of this variant in Table~\ref{tab:gating_eval_exp}.

\paragraph{FiLM gating} performs the feature-wise affine modulation conditioned on the mask \cite{perez2017filmvisualreasoninggeneral}. Two MLPs map the mask value at each token to a scale per-feature and shift parameters. Concretely, let $g_{\gamma}:\mathbb{R}^{1}\rightarrow\mathbb{R}^{d}$ and $g_{\beta}:\mathbb{R}^{1}\rightarrow\mathbb{R}^{d}$ be two-layer MLPs (with GELU and dropout), producing $\boldsymbol{\gamma}=g_{\gamma}(\QMask)$ and $\boldsymbol{\beta}=g_{\beta}(\QMask)$. FiLM modulation is then:

\begin{equation}
    \GatedImgEmb = \ImgEmb\odot (1+\boldsymbol{\gamma}) + \boldsymbol{\beta}
\end{equation}

\paragraph{Evaluation of gating mechanism.} To evaluate each gating mechanism, we compare their answer location metrics. We isolate the gating mechanism by keeping all answer localization weights fixed and apply no decoder loss. For each gating mechanism, we perform a sweep of question prior weight strength $\lambda_{\text{prior}} \in \{0.0, 0.2, 0.4, 0.6, 1.0 \}$.  We train with early stopping (patience = 5). We report the number of epochs and the best validation loss of the saved model. Table \ref{tab:gating_eval_exp} shows the final results. 

\begin{table}[ht!]
    \centering
    \small
    \caption{Experiment results from the Gating mechanism. The $w_0, h_0$ is the bias given to the answer projector to stabilize training.}
    \label{tab:gating_eval_exp} 
    \begin{tabular}{l c c c r r r r}
        \toprule
        \textbf{Gate} & $\boldsymbol{\lambda_{\text{prior}}}$ &
        $\boldsymbol{w_0}$ & $\boldsymbol{h_0}$ &
        \textbf{Ep.} & \textbf{Val.} &
        \textbf{IoU$_{\text{mean}}$} $\uparrow$ &
        \textbf{Cov$_{\text{mean}}$} $\uparrow$ \\
        \midrule
        Linear      & 0.00 & 0.55 & 0.19 & 42 & 0.8234 & 0.0609 & 0.1407 \\
        Linear      & 0.20 & 0.55 & 0.19 & 45 & 0.8575 & 0.0470 & 0.1110 \\
        Linear      & 0.40 & 0.55 & 0.19 & 45 & 0.8522 & 0.0530 & 0.1272 \\
        Linear      & 0.60 & 0.45 & 0.15 & 34 & 0.8545 & 0.0544 & 0.1283 \\
        Linear      & 1.00 & 0.65 & 0.22 & 34 & 0.8629 & 0.0519 & 0.1234 \\
        \midrule
        Residual    & 0.00 & 0.55 & 0.19 & 42 & 0.8411 & 0.0502 & 0.1177 \\
        Residual    & 0.20 & 0.55 & 0.19 & 45 & 0.8328 & 0.0588 & 0.1354 \\
        Residual    & 0.40 & 0.55 & 0.19 & 36 & 0.8304 & 0.0657 & 0.1473 \\
        Residual    & 0.60 & 0.45 & 0.15 & 36 & 0.8312 & 0.0637 & 0.1389 \\
        Residual    & 1.00 & 0.65 & 0.22 & 37 & 0.8536 & 0.0581 & 0.1320 \\
        \midrule
        SpatialAttn & 0.00 & 0.55 & 0.19 & 34 & 0.8173 & 0.0632 & 0.1443 \\
        SpatialAttn & 0.20 & 0.55 & 0.19 & 36 & 0.8212 & 0.0647 & 0.1490 \\
        SpatialAttn & 0.40 & 0.55 & 0.19 & 38 & 0.8322 & 0.0622 & 0.1423 \\
        SpatialAttn & 0.60 & 0.45 & 0.15 & 35 & 0.8412 & 0.0581 & 0.1405 \\
        SpatialAttn & 1.00 & 0.65 & 0.22 & 45 & 0.8445 & 0.0611 & 0.1420 \\
        \midrule
        FiLM        & 0.00 & 0.55 & 0.19 & 36 & 0.8156 & 0.0651 & 0.1439 \\
        FiLM        & 0.20 & 0.55 & 0.19 & 49 & 0.8213 & 0.0639 & 0.1428 \\
        FiLM        & 0.40 & 0.55 & 0.19 & 36 & 0.8265 & 0.0673 & 0.1507 \\
        FiLM        & 0.60 & 0.45 & 0.15 & 34 & 0.8338 & 0.0629 & 0.1430 \\
        FiLM        & 1.00 & 0.65 & 0.22 & 46 & 0.8344 & 0.0710 & 0.1557 \\
        \bottomrule
    \end{tabular}
\end{table}

The results indicate that the impact is not uniform between the gating variants. Table \ref{tab:gating_eval_exp} shows that \textit{Residual}- and \textit{FiLM}-based gating yielded on average the most gains, while \textit{Linear-} and \textit{Spatial-Attention-}based gating did not perform well. The best performing gating mechanism was FiLM, which achieved $\mathrm{IoU}_{\text{mean}} = 0.0710$, $\mathrm{Coverage}_{\text{mean}} = 15.57\%$ with $\lambda_{\text{prior}} = 1.0$. Overall, these results suggest that question-prior supervision can help, with its effectiveness depends on how the model integrates the prior through the gating mechanism.

\FloatBarrier
\subsection{Decoder Conditioning with Answer Localization}
\label{app:decoder_input}

Given the predicted answer location, we condition the decoder on this region to generate the final text answer. To further test the empirical upper-bound of the decoder,  we evaluate two re-encoding strategies that modify the input pixels and re-run the encoder. The mask approach masks out all pixels in the image that are not within the answer bounding box. The second re-encoding strategy crops the image to fit just the answer region before re-encoding. We also propose two additional methods for using the predicted answer location to generate the text answer that does not re-encode the image: attention mask and token prune.  The attention-mask approach masks the decoder attention to answer bounding box tokens. The token-pruning variant computes a soft weight per patch using Gaussian decay from the predicted box centre, binarizes at the per-sample median, and passes the resulting mask to the decoder's cross-attention. For the latter two variants, we compare the effect of giving the gated embeddings from the gating module. 

\paragraph{Decoder Conditioning Experiment.} We used the Pix2Struct-decoder for all variants  \cite{lee2023pix2structscreenshotparsingpretraining}. Each variant is given the ground truth answer location, and no answer localization loss ($ \lambda_{\mathrm{GIoU}} = 0, \lambda_{\text{centre}} = 0,  \lambda_{\text{area}} = 0$) and no question loss is applied ($\lambda_{\text{prior}} = 0$). Only decoder loss ($\lambda_{\text{dec}} > 0$) is applied.  We also compare the effect of decoder-unfrozen only, and the entire backbone unfrozen. The results strongly favoured the re-encoding strategies, improving the empirical upper-bound by $\sim0.3-0.5$ ANLS compared to the non re-encoding strategies. By re-encoding on the answer region, we can guarantee that the predicted answer is grounded in the answer location. Due to the large difference in performance, we chose the re-encoding variants for large scale training and further experiments. 

\begin{table}[ht!]
    \centering
    \caption{Decoder fine-tuning strategy ablation with ground truth answer location given. Each variant was trained on a single NVIDIA GH100 GPU. Here, ``Freeze Mode" indicate whether the backbone model will be updated with gradients during training. Frozen means the entire backbone model is set to evaluation mode (no gradient update). ``Decoder-unfrozen" means we allow weight updates for the decoder only). ``Unfrozen" means we allow the entire backbone model to update the weights. ``Emb." is the embedding type given to a non re-encoded strategy. Raw embeddings are from the encoder, while gated embeddings are conditioned on the gating module. Input type is the decoder condition strategy. We report the number of epochs trained before early stopping was triggers, and the final decoder loss. We evaluate using the main DocVQA metrics \cite{mathew2021docvqadatasetvqadocument}.}
    \label{tab:decoder_ablation}
    \small
    \begin{tabular}{l l l r r r r}
        \toprule
        \multirow{2}{*}{\textbf{Freeze mode}} &
        \multirow{2}{*}{\textbf{Emb.}} &
        \multirow{2}{*}{\textbf{Input Type}} &
        \multicolumn{2}{c}{\textbf{Convergence}} &
        \multicolumn{2}{c}{\textbf{DocVQA}} \\
        \cmidrule(lr){4-5}\cmidrule(lr){6-7}
        & & &
        Epochs & Dec.\ loss &
        ANLS $\uparrow$ & ACC $\uparrow$ \\
        \midrule
        frozen           & raw   & attention\_mask &  6 & 4.1230 & 0.1227 & 0.0539 \\
        decoder-unfrozen & raw   & attention\_mask & 18 & 2.1450 & 0.2147 & 0.0914 \\
        decoder-unfrozen & gated & attention\_mask & 13 & 2.1295 & 0.2184 & 0.0926 \\
        unfrozen         & raw   & attention\_mask & 13 & 1.8440 & 0.3267 & 0.1578 \\
        decoder-unfrozen & raw   & token\_prune    & 16 & 1.9710 & 0.2524 & 0.1197 \\
        decoder-unfrozen & gated & token\_prune    & 13 & 1.9729 & 0.2488 & 0.1189 \\
        unfrozen         & raw   & token\_prune    & 11 & 1.6900 & 0.3786 & 0.1939 \\
        decoder-unfrozen & --    & mask            & 13 & 1.9641 & 0.2733 & 0.1152 \\
        unfrozen         & --    & mask            &  9 & 1.2774 & 0.5404 & 0.2765 \\
        decoder-unfrozen & --    & crop            & 13 & 0.6538 & 0.7874 & 0.6428 \\
        unfrozen         & --    & crop            & 11 & 0.1176 & 0.8713 & 0.7756 \\
        \bottomrule
    \end{tabular}
\end{table}

\paragraph{Wall-Clock Time.} We report wall-clock training time per epoch and validation inference time per validation pass for each decoder input strategy. Because these measurements depend on hardware, the absolute times are not intended as a universal benchmark, but instead the relative differences to the Pix2Struct baseline. As shown in Table~\ref{tab:timing_by_input_type}, the attention mask and the token prune incur only a small overhead during training (+2–3\%) and slightly reduce validation time (about 1\%), consistent with these strategies operating on existing encoder features without introducing an additional full-image forward pass. In contrast, the mask and crop variants increase both training and validation time by roughly a factor of two (+105–116\% train, +94–104\% val). This increase is expected because these strategies perform an additional re-encoding step on the masked/cropped document region, adding a second encoder pass to the pipeline. Despite the extra compute from the second encoder pass, this overhead can be justified by the minimal parameter overhead of our projectors and by operating at a much smaller model scale while providing explicit localization and more verifiable predictions.

\begin{table}[ht!]
    \centering
    \small
    \caption{Mean training epoch time and validation inference time by input type. Percentages are relative to the baseline for each column. Each variant was trained on a 1x NVIDIA GH100 (120GB) GPU.}
    \label{tab:timing_by_input_type}
    \begin{tabular}{lrr}
        \toprule
        \textbf{Input type} & \textbf{Training Time (s)} & \textbf{Validation Time (s)} \\
        \midrule
        Pix2Struct (baseline) & 6217.9     & 1731.2 \\
        \midrule
        attention\_mask       & 6372.7 (+2.5\%)   & 1715.2 (-0.9\%) \\
        token\_prune          & 6355.5 (+2.2\%)   & 1713.6 (-1.0\%) \\
        mask                  & 13398.5 (+115.5\%) & 3537.0 (+104.3\%) \\
        crop                  & 12729.1 (+104.7\%) & 3360.3 (+94.1\%) \\
        \hline
    \end{tabular}
\end{table}

\subsection{Configurations}
\label{app:configurations}

We propose three configurations: \textit{base}, \textit{medium}, and \textit{large}. For each configuration, we only scale the answer projector. Across all configurations, we keep the same backbone, question projector, and gating mechanism. During development, we saw that the answer localization is the primary bottleneck. We change the number of layers, attention heads, and attention pooling with $k$ learned queries. Table \ref{tab:model_configs} shows how each configuration scales. Table \ref{tab:param_counts_presets} shows the number of parameters per configuration. To keep the approach lightweight, we constrain the number of parameters added to remain below the backbone parameter count. 

\begin{table*}[ht!]
    \centering
    \small
    \caption{\ModelName{} configuration presets. Only the answer projector is scaled across configurations. The backbone, question projector, and gating module are shared.}
    \label{tab:model_configs}
    \setlength{\tabcolsep}{7pt}
    \renewcommand{\arraystretch}{1.15}
    \begin{tabular}{lccc}
    \toprule
    \textbf{Answer projector setting} & \textbf{Base} & \textbf{Medium} & \textbf{Large} \\
    \midrule
    Fusion heads ($h$) & 8 & 12 & 16 \\
    Context layers ($L$) & 2 & 4 & 8 \\
    Context heads ($h$) & 8 & 12 & 16 \\
    Attention pooling ($k$) & 4 & 8 & 16 \\
    \bottomrule
    \end{tabular}
\end{table*}

\begin{table}[ht!]
    \centering
    \small
    \caption{Parameter count per configuration preset. The backbone is \texttt{pix2struct-docvqa-base}. Added modules include the question projector, answer projector, and gating module.}
    \label{tab:param_counts_presets}
    \begin{tabular}{lrrr}
        \toprule
        \textbf{Preset} & \textbf{Backbone} & \textbf{Added} & \textbf{Total} \\
        \midrule
        Base   & 282.3M & 48.4M (+17.1\%) & 330.7M \\
        Medium & 282.3M & 62.5M (+22.1\%) & 344.8M \\
        Large  & 282.3M & 90.9M (+32.2\%) & 373.2M \\
        \bottomrule
    \end{tabular}
\end{table}

\section{Training Details}
\label{app:training_details}

We train using 4x NVIDIA GH100 (120GB) GPUs per run\footnote{All experiments are feasible on a single GPU, but using four GPUs reduces wall-clock training time by approximately 4x in our setup.}. We use early stopping to reduce the number of epochs required to run with patience of 5-10 epochs. We train with mixed precision using autocast (BF16 when supported). We keep the seed and generation arguments for the decoder fixed for all runs. We use deterministic generation arguments with maximum 32 new tokens, one beam, no repeat ngram size of 2 and repetition penalty set to 1.2\footnote{Majority of the answer for the DocVQA dataset require few tokens to accurately generate \cite{mathew2021docvqadatasetvqadocument}, and we observed that by setting a small repetition penalty without beam-search, we got correct decoding.}. The Pix2Struct backbone and projectors (with gating) are separated into two different learning rate groups, where the backbone starts with a smaller learning rate. We implemented the training script to allow different schedulers, optimizers, and other hyper-parameters to be adjusted by CLI. Table \ref{tab:mask_hparams_and_metrics} shows the fine-tuning table for different configurations with different loss weights and optimizations on DocVQA. In our experiments, these larger configuration presets did not consistently improve performance over the base configuration, suggesting that localization quality is not primarily limited by projector capacity in this regime. Table \ref{tab:pfl_train_table} shows training results on PFL-DocVQA \cite{tito2024privacyawaredocumentvisualquestion}. 

\begin{table}[ht!]
    \centering
    \small
    \setlength{\tabcolsep}{4pt}
    \renewcommand{\arraystretch}{1.15}
    \caption{Training setup and evaluation metrics for each mask configuration on DocVQA. Each run with a fixed seed, and deterministic generation arguments for the decoder.}
    \resizebox{\linewidth}{!}{%
    \begin{tabular}{l c c c c c c c c c c c c c}
        \toprule
        \multirow{2}{*}{\textbf{Mask type}} &
        \multicolumn{5}{c}{\textbf{Regression loss weights}} &
        \multicolumn{3}{c}{\textbf{Optimization}} &
        \multicolumn{5}{c}{\textbf{Evaluation}} \\
        \cmidrule(lr){2-6}\cmidrule(lr){7-9}\cmidrule(lr){10-14}
        & $\lambda_{\text{IoU}}$ & $\lambda_{\text{centre}}$ & $\lambda_{\text{area}}$ & $\lambda_{\text{prior}}$ & $\lambda_{\text{dec}}$
        & Optimizer & LR & Scheduler
        & ACC $\uparrow$ & ANLS $\uparrow$ & $\mathrm{IoU}_{\text{mean}}$ $\uparrow$ & $\mathrm{Coverage}_{\text{mean}}$ $\uparrow$ & $\mathrm{AR}_{\text{mean}}$ \\
        \midrule
        Mask (base)        & 1.0 & 1.5 & 0.1  & 1.0  & 1.0 & AdamW & 3e-05 & ReduceLR. & 0.10 & 0.24 & 0.16 & 0.28 & 2.13 \\
        Mask (base)        & 1.5 & 1.5 & 0.05 & 1.0  & 1.0 & AdamW & 3e-05 & Cosine    & 0.10 & 0.24 & 0.12 & 0.26  & 4.19 \\ %
        Mask (base)        & 2.0 & 1.5 & 0.10 & 1.0  & 1.0 & AdamW & 2e-05 & Cosine    & 0.07 & 0.18 & 0.08 & 0.18  & 4.13 \\ %
        Mask (base)        & 2.0 & 1.5 & 0.20 & 2.0  & 1.0 & AdamW & 2e-05 & Cosine    & 0.06 & 0.15 & 0.06 & 0.13  & 3.23 \\ %
        \addlinespace
        Mask (medium)      & 1.0 & 1.5 & 0.1  & 1.0  & 1.0 & AdamW & 3e-05 & ReduceLR. & 0.09 & 0.23 & 0.15 & 0.26 & 2.25 \\
        Mask (medium)      & 1.5 & 1.5 & 0.05 & 1.0  & 1.0 & AdamW & 2e-05 & Cosine    & 0.09 & 0.20 & 0.09 & 0.24 & 5.51 \\ %
        \addlinespace
        Mask (large)       & 1.0 & 1.5 & 0.1  & 1.0  & 1.0 & AdamW & 3e-05 & ReduceLR. & 0.09 & 0.22 & 0.15 & 0.25 & 2.14 \\
        Mask (large)       & 1.5 & 1.5 & 0.05 & 1.0  & 1.0 & AdamW & 1e-05 & Cosine    & 0.06 & 0.16 & 0.05 & 0.15 & 7.58 \\ %
        \midrule
        Crop (base)        & 1.00 & 1.50 & 0.10  & 1.00  & 1.00 & AdamW & 3e-05 & ReduceLR. & 0.10 & 0.19 & 0.06 & 0.10 & 2.29 \\
        Crop (base)        & 2.00 & 2.00 & 0.05 & 0.5  & 0.50 & AdamW & 3e-05 & Cosine    & 0.26 & 0.37 & 0.08 & 0.27 & 7.48  \\ %
        Crop (base)        & 3.00 & 2.00 & 0.00 & 0.25 & 0.25& AdamW & 3e-05 & Cosine    & 0.27 & 0.40 & 0.02 & 0.87 & 334.19 \\  %
        Crop (base)        & 2.50 & 2.00 & 0.02 & 0.50  & 0.25& AdamW & 3e-05 & Cosine    & 0.34 & 0.43 & 0.06 & 0.37 & 19.53 \\ %
        Crop (base)        & 2.00 & 1.50 & 0.20 & 2.00  & 1.00 & AdamW & 3e-05 & Cosine    & 0.14   & 0.25   & 0.06  & 0.14   & 2.64 \\ %
        \addlinespace
        Crop (medium)      & 1.00 & 1.50 & 0.10  & 1.00  & 1.00 & AdamW & 3e-05 & ReduceLR. & 0.09 & 0.18 & 0.05 & 0.08 & 2.18 \\
        Crop (medium)      & 2.00 & 2.00 & 0.05 & 0.50  & 0.50 & AdamW & 2e-05 & Cosine    & 0.25 & 0.36 & 0.07 & 0.27 & 9.19 \\ %
        \addlinespace
        Crop (large)       & 1.00 & 1.50 & 0.10  & 1.00  & 1.00 & AdamW & 3e-05 & ReduceLR. & 0.08 & 0.17 & 0.07 & 0.18 & 2.13 \\
        Crop (large)       & 2.00 & 2.00 & 0.05 & 0.50  & 0.50 & AdamW & 1e-05 & Cosine    & 0.22 & 0.32 & 0.05 & 0.22 & 9.76 \\ %
        \bottomrule
    \end{tabular}%
    }
    \label{tab:mask_hparams_and_metrics}
\end{table}

\begin{table}[ht!]
    \centering
    \small
    \setlength{\tabcolsep}{4pt}
    \renewcommand{\arraystretch}{1.15}
    \caption{Training setup and evaluation metrics for each mask configuration on PFL-DocVQA. Each run with a fixed seed, and deterministic generation arguments for the decoder.}
    \resizebox{\linewidth}{!}{%
    \begin{tabular}{l c c c c c c c c c c c c c}
        \toprule
        \multirow{2}{*}{\textbf{Mask type}} &
        \multicolumn{5}{c}{\textbf{Regression loss weights}} &
        \multicolumn{3}{c}{\textbf{Optimization}} &
        \multicolumn{5}{c}{\textbf{Evaluation}} \\
        \cmidrule(lr){2-6}\cmidrule(lr){7-9}\cmidrule(lr){10-14}
        & $\lambda_{\text{IoU}}$ & $\lambda_{\text{centre}}$ & $\lambda_{\text{area}}$ & $\lambda_{\text{prior}}$ & $\lambda_{\text{dec}}$
        & Optimizer & LR & Scheduler
        & ACC $\uparrow$ & ANLS $\uparrow$ & $\mathrm{IoU}_{\text{mean}}$ $\uparrow$ & $\mathrm{Coverage}_{\text{mean}}$ $\uparrow$ & $\mathrm{AR}_{\text{mean}}$ \\
        \midrule
        Mask (base) & 1.50 & 1.50 & 0.05 & 1.00 & 1.00 & AdamW & 3e-05 & Cosine    & 0.34 & 0.63 & 0.43 & 0.69 & 2.85 \\ %
        Mask (base) & 2.00 & 1.50 & 0.10 & 1.00 & 1.00 & AdamW & 2e-05 & Cosine    & 0.31 & 0.59 & 0.42 & 0.66 & 2.54 \\ %
        Mask (base) & 2.00 & 1.50 & 0.20 & 2.00 & 1.00 & AdamW & 2e-05 & Cosine    & 0.30 & 0.56 & 0.42 & 0.63 & 2.00 \\ %
        Mask (base) & 1.25 & 1.50 & 0.10 & 1.00 & 1.00 & AdamW & 3e-05 & ReduceLR. & 0.30 & 0.58 & 0.44 & 0.63 & 2.05 \\ %
        Mask (base) & 1.00 & 1.50 & 0.05 & 1.00 & 1.00 & AdamW & 3e-05 & ReduceLR. & 0.30 & 0.59 & 0.44 & 0.65 & 2.34 \\ %
        \midrule
        Crop (base) & 2.50 & 2.00 & 0.02 & 0.50 & 0.25 & AdamW & 3e-05 & Cosine    & 0.60 & 0.77 & 0.47 & 0.75 & 4.68 \\ %
        Crop (base) & 3.00 & 2.00 & 0.05 & 0.25 & 0.25 & AdamW & 3e-05 & Cosine    & 0.60 & 0.77 & 0.44 & 0.71 & 3.58 \\ %
        Crop (base) & 2.50 & 2.00 & 0.02 & 0.50 & 0.25 & AdamW & 3e-05 & ReduceLR. & 0.61 & 0.78 & 0.47 & 0.76 & 4.92 \\ %
        Crop (base) & 2.00 & 2.00 & 0.05 & 0.50 & 0.50 & AdamW & 3e-05 & ReduceLR. & 0.58 & 0.75 & 0.43 & 0.69 & 3.11 \\ %
        \bottomrule
    \end{tabular}%
    }
    \label{tab:pfl_train_table}
\end{table}

\paragraph{Loss Plots.} Figure~\ref{fig:loss_figure} shows the loss plot of the best-performing model from Table \ref{tab:mask_hparams_and_metrics}. We plot the total loss training (Figure \ref{fig:training_loss}), and the total validation (Figure \ref{fig:validation_loss}). We also plot individual training losses. Figure \ref{fig:projector_loss} shows the projector loss $\mathcal{L}_{\text{proj}}$, and Figure \ref{fig:decoder_loss} show the decoder loss $\mathcal{L}_{\text{dec}}$. The vertical marker in the decoder plot indicates the end of the decoder-loss warmup period. After warmup, the decoder objective is applied at full weight (i.e $\lambda_{\text{dec}} > 0$) while the projector objectives continue to provide localization supervision.

\pgfplotstableread[col sep=space]{%
epoch train_total val_total train_proj val_proj train_giou val_giou train_dec val_dec train_qprior val_qprior
1 5.4013 1.1569 5.3706 1.1532 1.3502 0.3426 0.0000 0.0000 0.0307 0.0037
2 4.6095 1.0685 4.5236 1.0498 1.4265 0.3615 0.0720 0.0152 0.0139 0.0035
3 4.4056 1.0572 4.2814 1.0258 1.4100 0.3501 0.1107 0.0279 0.0135 0.0036
4 4.2966 1.0554 4.1364 1.0121 1.3657 0.3478 0.1470 0.0399 0.0133 0.0034
5 4.2424 1.0482 4.0498 0.9963 1.3326 0.3330 0.1792 0.0484 0.0135 0.0035
6 4.1616 1.0468 3.9423 0.9805 1.3078 0.3423 0.2051 0.0622 0.0143 0.0041
7 4.0554 1.0072 3.8205 0.9387 1.2831 0.3289 0.2190 0.0644 0.0159 0.0040
8 3.8795 1.0051 3.6371 0.9277 1.2544 0.3275 0.2269 0.0733 0.0155 0.0041
9 3.7825 1.0117 3.5229 0.9236 1.2338 0.3291 0.2439 0.0841 0.0156 0.0040
10 3.7544 1.0186 3.4740 0.9203 1.2242 0.3267 0.2645 0.0945 0.0160 0.0038
11 3.7100 1.0217 3.4125 0.9147 1.2081 0.3253 0.2817 0.1029 0.0158 0.0041
12 3.6712 1.0262 3.3835 0.9180 1.2003 0.3254 0.2716 0.1044 0.0161 0.0039
13 3.6143 1.0192 3.3369 0.9189 1.1888 0.3131 0.2613 0.0963 0.0161 0.0040
14 3.5753 1.0123 3.3049 0.9132 1.1795 0.3118 0.2543 0.0950 0.0161 0.0041
15 3.5290 1.0137 3.2683 0.9123 1.1705 0.3162 0.2447 0.0975 0.0160 0.0040
16 3.4871 1.0041 3.2298 0.9013 1.1607 0.3174 0.2413 0.0988 0.0160 0.0040
17 3.4539 0.9973 3.2005 0.8989 1.1536 0.3105 0.2374 0.0944 0.0161 0.0039
18 3.3954 0.9911 3.1513 0.8888 1.1419 0.3171 0.2281 0.0984 0.0160 0.0040
19 3.3495 0.9891 3.1111 0.8908 1.1324 0.3111 0.2224 0.0943 0.0160 0.0040
20 3.3120 0.9939 3.0795 0.8936 1.1242 0.3120 0.2164 0.0963 0.0161 0.0040
21 3.2819 0.9870 3.0536 0.8869 1.1176 0.3139 0.2122 0.0960 0.0160 0.0042
22 3.2454 0.9854 3.0232 0.8818 1.1094 0.3183 0.2064 0.0996 0.0159 0.0040
23 3.2196 0.9898 2.9993 0.8855 1.1031 0.3176 0.2044 0.1003 0.0160 0.0041
24 3.1848 0.9890 2.9709 0.8859 1.0949 0.3161 0.1978 0.0992 0.0161 0.0040
25 3.1575 0.9795 2.9474 0.8784 1.0889 0.3144 0.1941 0.0971 0.0160 0.0040
26 3.1274 0.9832 2.9207 0.8789 1.0814 0.3160 0.1907 0.1003 0.0160 0.0040
27 3.1094 0.9791 2.9046 0.8787 1.0776 0.3122 0.1888 0.0964 0.0160 0.0040
28 3.0810 0.9797 2.8811 0.8784 1.0709 0.3131 0.1840 0.0973 0.0159 0.0040
29 3.0521 0.9817 2.8571 0.8756 1.0633 0.3173 0.1791 0.1021 0.0159 0.0040
30 3.0339 0.9764 2.8412 0.8764 1.0592 0.3100 0.1769 0.0959 0.0158 0.0040
31 3.0110 0.9802 2.8201 0.8769 1.0530 0.3134 0.1750 0.0993 0.0160 0.0040
32 2.9960 0.9804 2.8057 0.8769 1.0496 0.3147 0.1743 0.0995 0.0159 0.0040
33 2.9754 0.9756 2.7880 0.8727 1.0443 0.3135 0.1714 0.0989 0.0159 0.0040
34 2.9561 0.9796 2.7712 0.8733 1.0391 0.3170 0.1691 0.1022 0.0159 0.0040
35 2.9393 0.9764 2.7561 0.8706 1.0347 0.3152 0.1673 0.1018 0.0159 0.0040
36 2.9234 0.9757 2.7427 0.8705 1.0309 0.3152 0.1648 0.1011 0.0159 0.0041
37 2.9081 0.9802 2.7285 0.8735 1.0262 0.3165 0.1636 0.1027 0.0159 0.0040
38 2.8964 0.9781 2.7189 0.8718 1.0241 0.3161 0.1618 0.1023 0.0158 0.0040
39 2.8826 0.9796 2.7063 0.8730 1.0199 0.3167 0.1604 0.1027 0.0159 0.0040
40 2.8707 0.9778 2.6964 0.8712 1.0168 0.3162 0.1583 0.1026 0.0160 0.0040
41 2.8649 0.9813 2.6900 0.8738 1.0145 0.3178 0.1589 0.1035 0.0159 0.0040
42 2.8568 0.9793 2.6813 0.8720 1.0123 0.3174 0.1597 0.1033 0.0159 0.0040
43 2.8482 0.9782 2.6750 0.8709 1.0103 0.3176 0.1573 0.1033 0.0159 0.0040
}\lossdata

\def\warmupepoch{10}     %
\def\earlystopepoch{43}  %

\begin{figure}[ht!]
    \centering
    \begin{subfigure}[t]{0.495\textwidth}
        \centering
        \begin{tikzpicture}
        \begin{axis}[
          width=1.03\linewidth, height=6.2cm,
          xlabel={Epoch}, ylabel={Loss},
          tick label style={font=\small},
          label style={font=\small},
          grid=both,
          clip=false,
          xmin=1, xmax=\earlystopepoch,
          ymin=2.77, ymax=5.48,
          xtick distance=10,
        ]
        \addplot[ultra thick, blue!75!black, mark=none]
          table[x=epoch,y=train_total]{\lossdata};
        \end{axis}
        \end{tikzpicture}
        \subcaption{Training loss.}
        \label{fig:training_loss}
    \end{subfigure}
    \hfill
    \begin{subfigure}[t]{0.495\textwidth}
        \centering
        \begin{tikzpicture}
        \begin{axis}[
          width=1.03\linewidth, height=6.2cm,
          xlabel={Epoch}, ylabel={Loss},
          tick label style={font=\small},
          label style={font=\small},
          grid=both,
          clip=false,
          xmin=1, xmax=\earlystopepoch,
          ymin=0.97, ymax=1.17,
          xtick distance=10,
        ]
        \addplot[ultra thick, orange!85!black, mark=none]
          table[x=epoch,y=val_total]{\lossdata};
        \end{axis}
        \end{tikzpicture}
        \subcaption{Validation loss.}
        \label{fig:validation_loss}
    \end{subfigure}
    \vspace{0.9em}
    \begin{subfigure}[t]{0.495\textwidth}
        \centering
        \begin{tikzpicture}
        \begin{axis}[
          width=1.03\linewidth, height=6.2cm,
          xlabel={Epoch}, ylabel={Loss},
          tick label style={font=\small},
          label style={font=\small},
          grid=both,
          clip=false,
          xmin=1, xmax=\earlystopepoch,
          ymin=2.59, ymax=5.46,
          xtick distance=10,
        ]
        \addplot[ultra thick, teal!75!black, mark=none]
          table[x=epoch,y=train_proj]{\lossdata};
        \end{axis}
        \end{tikzpicture}
        \subcaption{Projector loss (train).}
        \label{fig:projector_loss}
    \end{subfigure}
    \hfill
    \begin{subfigure}[t]{0.495\textwidth}
        \centering
        \begin{tikzpicture}
        \begin{axis}[
          width=1.03\linewidth, height=6.2cm,
          xlabel={Epoch}, ylabel={Loss},
          tick label style={font=\small},
          label style={font=\small},
          grid=both,
          clip=false,
          xmin=1, xmax=\earlystopepoch,
          ymin=0.00, ymax=0.30,
          xtick distance=10,
        ]
        \addplot[ultra thick, purple!80!black, mark=none]
          table[x=epoch,y=train_dec]{\lossdata};
        
        \addplot[warmupline] coordinates {
          (\warmupepoch,\pgfkeysvalueof{/pgfplots/ymin})
          (\warmupepoch,\pgfkeysvalueof{/pgfplots/ymax})
        };
        \node[font=\scriptsize, anchor=north west, xshift=2pt, yshift=-2pt,
              fill=white, fill opacity=0.7, text opacity=1, inner sep=1pt]
          at (axis cs:\warmupepoch,\pgfkeysvalueof{/pgfplots/ymax}) {Warmup end};
        \end{axis}
        \end{tikzpicture}
        \subcaption{Decoder loss (train).}
        \label{fig:decoder_loss}
    \end{subfigure}
    
    \caption{\textbf{Training plots of \ModelName{}.} \textit{Top-left:} total training loss. \textit{Top-right:} total validation loss. \textit{Bottom-left:} training projector loss (localization/prior objectives). \textit{Bottom-right:} training decoder loss (text generation objective). The vertical line marks the end of the decoder-loss warmup, after which the decoder loss is applied at full weight. Curves are shown up to the early-stopping epoch (\earlystopepoch).}
    \label{fig:loss_figure}
\end{figure}
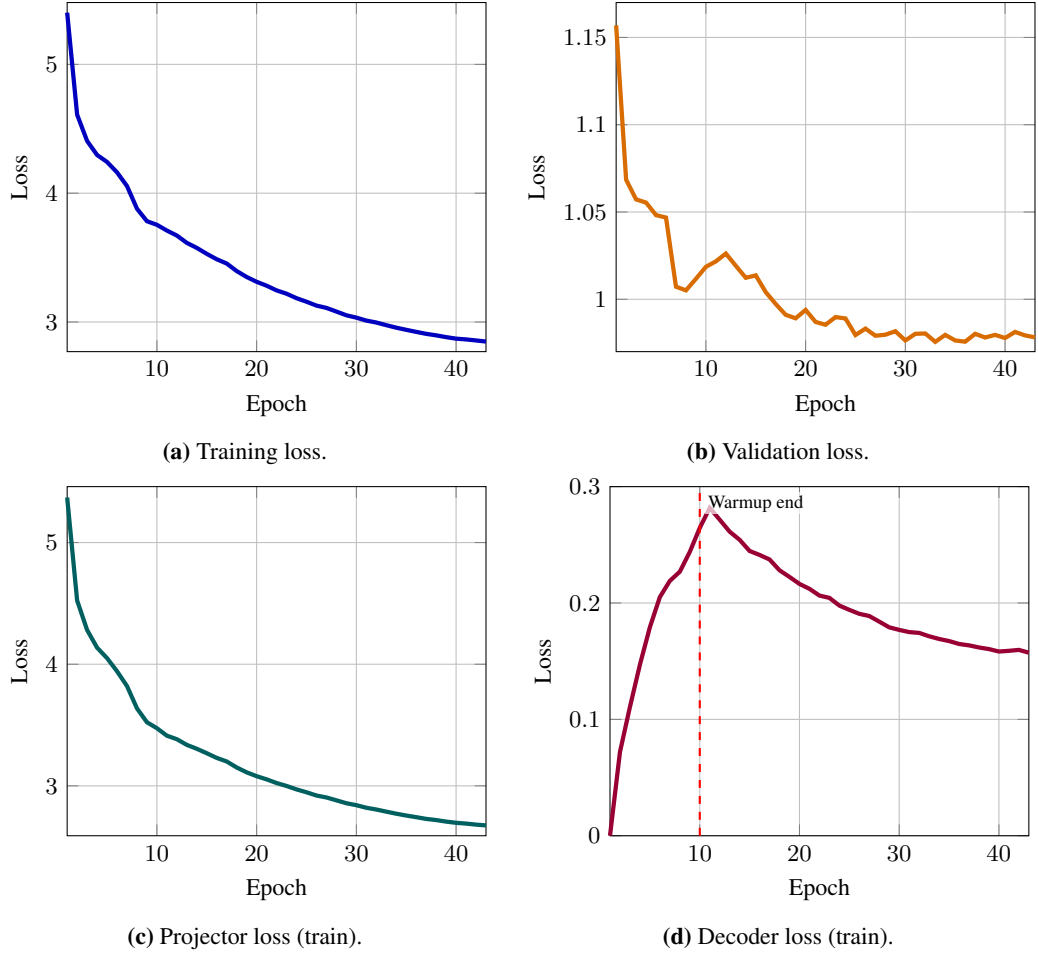

\FloatBarrier
\section{Backbone Ablation Study}
\label{app:backbone_abl_study}

We conduct additional training using two additional backbones; Donut and Pix2Struct \cite{kim2022ocrfreedocumentunderstandingtransformer, lee2023pix2structscreenshotparsingpretraining}. For Donut, we use the DocVQA finetuned base configuration with 200M trainable parameters (\texttt{naver-clova-ix/donut-base-finetuned-docvqa}\footnote{\url{https://huggingface.co/naver-clova-ix/donut-base-finetuned-docvqa}}). We keep the encoder frozen and train only the decoder. Training with Donut allows us to show that the framework can be applied to other ViT based image-encoder-text-decoder. For Pix2Struct, we use the large variant with 1.3B trainable parameters (\texttt{google/pix2struct-large}\footnote{\url{https://huggingface.co/google/pix2struct-large}}).

Tables \ref{tab:backbone_docvqa_res} and \ref{tab:backbone_pfl_docvqa_res} show the results for two additional backbone on DocVQA and PFL-DocVQA respectively \cite{mathew2021docvqadatasetvqadocument, tito2024privacyawaredocumentvisualquestion}. The crop variant consistently outperforms the mask variant across all backbones and datasets. Notably, Donut-Base improves from 0.18 ACC on DocVQA to 0.42 ACC on PFL-DocVQA with only 20\% of training data, indicating that lower performance reflects data limitations rather than architectural incompatibility. These results confirm that \ModelName{} is not tied to a specific backbone architecture.

\begin{table}[ht!]
    \centering
    \small
    \caption{Experimental configurations and evaluation results on DocVQA across backbone and mask variants.}
    \setlength{\tabcolsep}{4pt}
    \renewcommand{\arraystretch}{1.15}
    \resizebox{\linewidth}{!}{%
    \begin{tabular}{l l c c c c c c c c c c c c c c}
        \toprule
        \textbf{Backbone} &
        \textbf{Mask type} &
        $\lambda_{\text{IoU}}$ &
        $\lambda_{\text{centre}}$ &
        $\lambda_{\text{area}}$ &
        $\lambda_{\text{prior}}$ &
        $\lambda_{\text{dec}}$ &
        \textbf{Batch size} &
        \textbf{LR} &
        \textbf{Optimizer} &
        \textbf{Scheduler} &
        \textbf{ACC} $\uparrow$ &
        \textbf{ANLS} $\uparrow$ &
        $\mathrm{IoU}_{\text{mean}}$ $\uparrow$ &
        $\mathrm{Coverage}_{\text{mean}}$ $\uparrow$ &
        $\mathrm{AR}_{\text{mean}}$ \\
        \midrule
        pix2struct-large & crop & 2.5 & 2.0 & 0.02 & 0.50 & 0.25 & 16 & 3e-5 & AdamW & Cosine & 0.3801 & 0.4821 & 0.1144 & 0.3889 & 15.2934 \\
        pix2struct-large & crop & 3.0 & 2.0 & 0.02 & 0.25 & 0.25 & 16 & 3e-5 & AdamW & Cosine & 0.3567 & 0.4603 & 0.1161 & 0.3828 & 18.5991 \\
        pix2struct-large & crop & 2.5 & 2.0 & 0.05 & 0.50 & 0.25 & 16 & 3e-5 & AdamW & Cosine & 0.3567 & 0.4636 & 0.1139 & 0.3610 & 8.1743 \\
        pix2struct-large & mask & 2.5 & 2.0 & 0.05 & 0.50 & 0.25 & 8  & 3e-5 & AdamW & Cosine & 0.1340 & 0.2632 & 0.1405 & 0.4019 & 6.5148 \\
        pix2struct-large & mask & 1.5 & 1.5 & 0.05 & 0.50 & 0.25 & 8  & 3e-5 & AdamW & Cosine & 0.1235 & 0.2561 & 0.1465 & 0.3824 & 5.4718 \\
        pix2struct-large & mask & 2.0 & 2.0 & 0.02 & 0.50 & 0.25 & 8  & 3e-5 & AdamW & Cosine & 0.1566 & 0.3134 & 0.1997 & 0.4684 & 8.5179 \\
        donut-base & mask & 2.5 & 2.0 & 0.05 & 0.50 & 0.25 & 8  & 3e-5 & AdamW & Cosine & 0.1792 & 0.2629 & 0.1171 & 0.2895 & 6.7555 \\
        donut-base & mask & 2.0 & 2.0 & 0.02 & 0.50 & 0.25 & 8  & 3e-5 & AdamW & Cosine & 0.1776 & 0.2579 & 0.1006 & 0.3470 & 14.6427 \\
        donut-base & crop & 2.5 & 2.0 & 0.02 & 0.50 & 0.25 & 16 & 3e-5 & AdamW & Cosine & 0.1759 & 0.2677 & 0.0968 & 0.3804 & 19.2161 \\
        \bottomrule
    \end{tabular}
    }
    \label{tab:backbone_docvqa_res}
\end{table}

\begin{table}[ht!]
    \centering
    \small
    \caption{Experimental configurations and evaluation results across backbone and mask variants. Trained on a 20\% random subset of the PFL-DocVQA training set.}
    \setlength{\tabcolsep}{4pt}
    \renewcommand{\arraystretch}{1.15}
    \resizebox{\linewidth}{!}{%
    \begin{tabular}{l l c c c c c c c c c c c c c c}
        \toprule
        \textbf{Backbone} &
        \textbf{Mask type} &
        $\lambda_{\text{IoU}}$ &
        $\lambda_{\text{centre}}$ &
        $\lambda_{\text{area}}$ &
        $\lambda_{\text{prior}}$ &
        $\lambda_{\text{dec}}$ &
        \textbf{Batch size} &
        \textbf{LR} &
        \textbf{Optimizer} &
        \textbf{Scheduler} &
        \textbf{ACC} $\uparrow$ &
        \textbf{ANLS} $\uparrow$ &
        $\mathrm{IoU}_{\text{mean}}$ $\uparrow$ &
        $\mathrm{Coverage}_{\text{mean}}$ $\uparrow$ &
        $\mathrm{AR}_{\text{mean}}$ \\
        \midrule
        pix2struct-large & mask & 1.5 & 1.5 & 0.05 & 1.0  & 1.0  & 8  & 3e-5 & AdamW & Cosine   & 0.1538 & 0.3680 & 0.2197 & 0.5616 & 4.8337 \\
        pix2struct-large & mask & 2.0 & 1.5 & 0.10 & 1.0  & 1.0  & 8  & 3e-5 & AdamW & Cosine   & 0.1126 & 0.2798 & 0.2023 & 0.4338 & 3.3406 \\
        pix2struct-large & mask & 1.5 & 1.5 & 0.05 & 0.50 & 0.25 & 8  & 3e-5 & AdamW & Cosine   & 0.1164 & 0.2902 & 0.2295 & 0.4969 & 4.4745 \\
        pix2struct-large & mask & 2.5 & 2.0 & 0.05 & 0.50 & 0.25 & 8  & 3e-5 & AdamW & Cosine   & 0.1210 & 0.3008 & 0.2299 & 0.5568 & 5.7185 \\
        pix2struct-large & crop & 2.5 & 2.0 & 0.02 & 0.50 & 0.25 & 16 & 3e-5 & AdamW & Cosine   & 0.5845 & 0.7514 & 0.2649 & 0.7157 & 9.1719 \\
        pix2struct-large & crop & 2.5 & 2.0 & 0.02 & 0.50 & 0.25 & 16 & 3e-5 & AdamW & ReduceLR & 0.5766 & 0.7443 & 0.2652 & 0.6573 & 8.2659 \\
        pix2struct-large & crop & 3.0 & 2.0 & 0.02 & 0.25 & 0.25 & 16 & 3e-5 & AdamW & Cosine   & 0.5795 & 0.7503 & 0.2796 & 0.6981 & 8.9311 \\
        pix2struct-large & crop & 2.5 & 2.0 & 0.05 & 0.50 & 0.25 & 16 & 3e-5 & AdamW & Cosine   & 0.5144 & 0.6794 & 0.2722 & 0.5506 & 4.5411 \\
        donut-base & mask & 2.0 & 2.0 & 0.02 & 0.50 & 0.25 & 8  & 3e-5 & AdamW & Cosine   & 0.3817 & 0.5175 & 0.1932 & 0.6377 & 9.7402 \\
        donut-base & crop & 2.5 & 2.0 & 0.02 & 0.50 & 0.25 & 16 & 3e-5 & AdamW & Cosine   & 0.4217 & 0.5274 & 0.1742 & 0.6731 & 13.7836 \\
        donut-base & crop & 2.5 & 2.0 & 0.05 & 0.50 & 0.25 & 16 & 3e-5 & AdamW & Cosine   & 0.3037 & 0.3895 & 0.1164 & 0.4626 & 6.7448 \\
        \bottomrule
    \end{tabular}
    }
    \label{tab:backbone_pfl_docvqa_res}
\end{table}

\FloatBarrier
\section{Qualitative Analysis of Area Ratio}
\label{app:ar_examples}

\begin{figure}[htbp]
  \centering
  \begin{subfigure}[t]{0.32\textwidth}
    \centering
    \includegraphics[width=\linewidth]{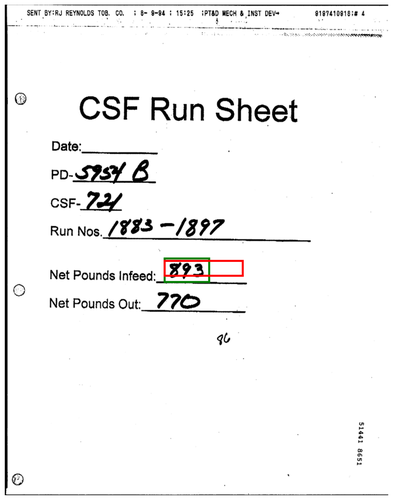}
    \caption{AR = 2.1164, ANLS = 0.2393, ACC = 0.1025} %
    \label{fig:ar_low}
  \end{subfigure}\hfill
  \begin{subfigure}[t]{0.32\textwidth}
    \centering
    \includegraphics[width=\linewidth]{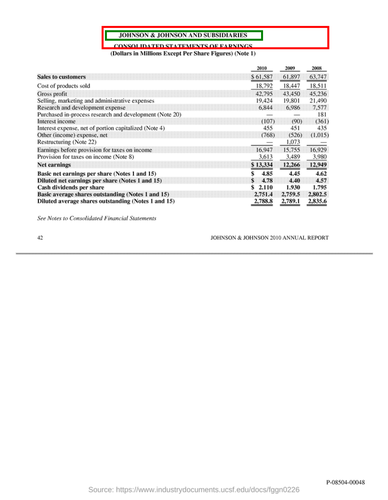}
    \caption{AR = 7.24, ANLS = 0.3713, ACC = 0.2761}
    \label{fig:ar_med}
  \end{subfigure}\hfill
  \begin{subfigure}[t]{0.32\textwidth}
    \centering
    \includegraphics[width=\linewidth]{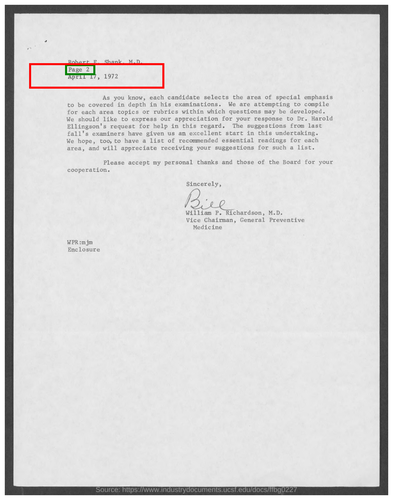}
    \caption{AR = 19.53, ANLS = 0.4328, ACC = 0.3352 }
    \label{fig:ar_high}
  \end{subfigure}

  \caption{Examples of three different examples with different $\mathrm{AR}$. The three different models are from training in Appendix \ref{app:training_details}. Figure \ref{fig:ar_low} shows small $\mathrm{AR}$ but enough to fill the necessary context. Here, the DocVQA performance is lower. This might be due to text being cut off by patches (e.g the number 7, may be cut off to visually look like 1 within the answer region). Figure \ref{fig:ar_med} shows a medium $\mathrm{AR}$, and a stronger performance. Figure \ref{fig:ar_high} shows higher $\mathrm{AR}$, where additional context is included. The predicted answer region still filters out majority of the document, but provide less compact explanations.}
  \label{fig:ar_examples_figure}
\end{figure}

We argue that a given $\mathrm{AR} > 1$  is not necessarily undesirable. The ground truth answers region is often small. Adding extra prediction pixels yields larger $\mathrm{AR}$, even if the added pixels contain a sensible context. For this reason, we do not apply a threshold based on $\mathrm{AR}$ directly during training or inference. Figure \ref{fig:ar_examples_figure} illustrates this using examples from the validation data split. The $\mathrm{AR}$ is reported at different sizes with the DocVQA metrics \cite{mathew2021docvqadatasetvqadocument}. With $\mathrm{AR} = 19.28$, we observed a larger prediction area than required, but still within reason. At the same time, we acknowledge that a very large $\mathrm{AR}$ value indicates over-selection. During training in Appendix \ref{app:training_details}, we observed that when the prediction area was not regularised ($\lambda_{\text{area}} \approx 0.0$), the model predicted unreasonably large regions ($\mathrm{AR} = 334.19$). There is a trade-off between selecting the minimal answer region and enough context for the decoder.  

To further quantify explanation quality, Table \ref{tab:category_analysis} reports localization metrics grouped by prediction accuracy. Correct predictions exhibit 5.4× higher Coverage (0.70 vs. 0.13) and lower AR (12.92 vs. 24.34) than incorrect ones, demonstrating that the model produces tighter and more focused explanations when it answers correctly. This consistent gap confirms that explanation quality is directly linked to prediction quality, validating that \ModelName's explanations are faithful to its decision process rather than arbitrary region selections.

\begin{table}[ht!]
    \centering
    \caption{DocVQA performance metrics grouped by prediction accuracy. We define each category based on the ANLS performance of the group. Specifically, Correct: ANLS $\geq 0.75$; Neutral: $0.50 \leq$ ANLS $< 0.75$; Incorrect: ANLS $< 0.50$. N denotes the number of examples in each group. Experiment was conducted by using our best performing \ModelName{} on examples in the validation split. }
    \begin{tabular}{lccccc}
        \toprule
        \textbf{Group} & \textbf{N} & \textbf{Mean ANLS} & \textbf{IoU} & \textbf{Coverage} & \textbf{AR} \\
        \midrule
        Correct & 1968 & 0.9778 & 0.1327 & 0.7014 & 12.9189 \\
        Neutral & 386 & 0.5711 & 0.0419 & 0.2580 & 20.9525 \\
        Incorrect & 2601 & 0.0000 & 0.0162 & 0.1294 & 24.3361 \\
        Overall & 4955 & 0.4328 & 0.0645 & 0.3666 & 19.5379 \\
        \bottomrule
    \end{tabular}
    \label{tab:category_analysis}
\end{table}

\FloatBarrier
\section{Faithfulness and Robustness Experiment Details}
\label{app:faithfulness_details}

\begin{figure}[htbp]
  \centering
  \begin{subfigure}[t]{0.32\textwidth}
    \centering
    \includegraphics[width=\linewidth]{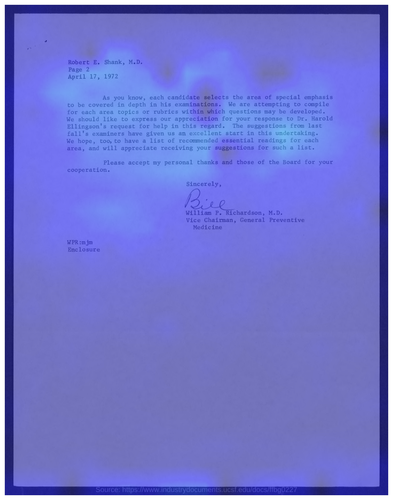}
    \caption{Masking probability = 10\%}
    \label{fig:masking_overlap_examples_a}
  \end{subfigure}\hfill
  \begin{subfigure}[t]{0.32\textwidth}
    \centering
    \includegraphics[width=\linewidth]{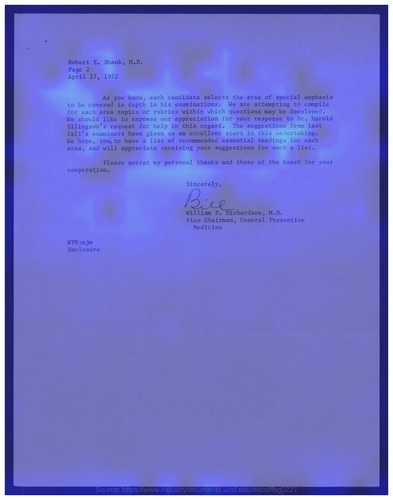}
    \caption{Masking probability = 50\%}
    \label{fig:masking_overlap_examples_b}
  \end{subfigure}\hfill
  \begin{subfigure}[t]{0.32\textwidth}
    \centering
    \includegraphics[width=\linewidth]{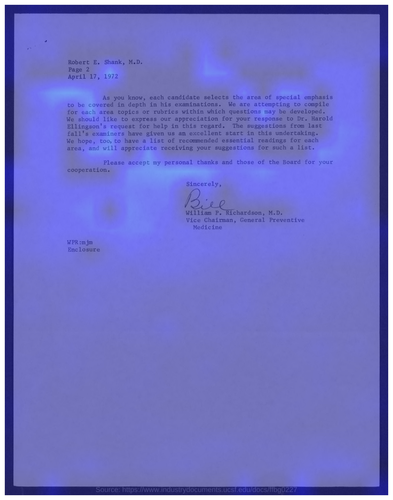}
    \caption{Masking probability = 90\%}
    \label{fig:masking_overlap_examples_c}
  \end{subfigure}

  \caption{Masked question-evidence heatmaps when masking is applied to patches that \textbf{overlap} the question-prior region. The masking probability denotes the fraction of overlapping patches that are removed (set to zero).}
  \label{fig:masking_overlap_examples}
\end{figure}

\begin{figure}[htbp]
  \centering
  \begin{subfigure}[t]{0.32\textwidth}
    \centering
    \includegraphics[width=\linewidth]{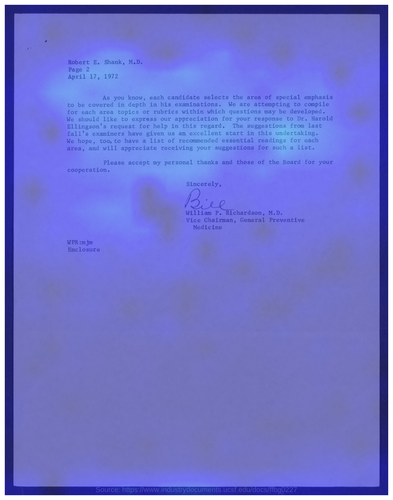}
    \caption{Masking probability = 10\%}
    \label{fig:masking_non_overlap_examples_a}
  \end{subfigure}\hfill
  \begin{subfigure}[t]{0.32\textwidth}
    \centering
    \includegraphics[width=\linewidth]{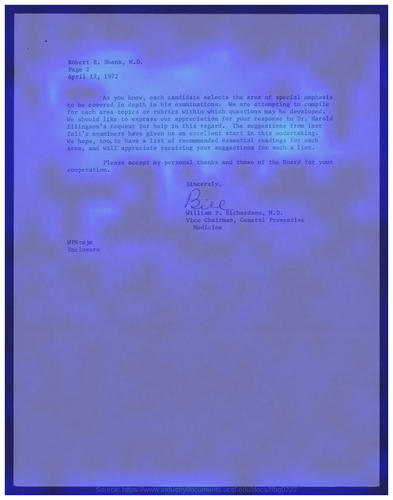}
    \caption{Masking probability = 50\%}
    \label{fig:masking_non_overlap_examples_b}
  \end{subfigure}\hfill
  \begin{subfigure}[t]{0.32\textwidth}
    \centering
    \includegraphics[width=\linewidth]{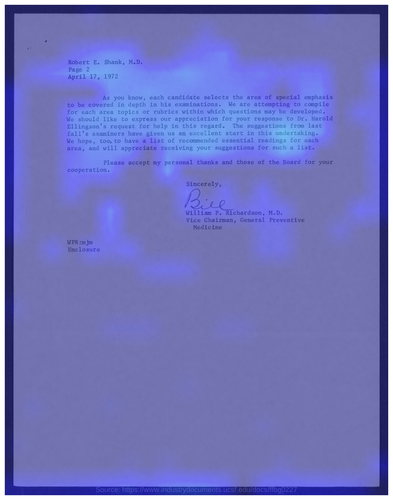}
    \caption{Masking probability = 90\%}
    \label{fig:masking_non_overlap_examples_c}
  \end{subfigure}

  \caption{Masked question-evidence heatmaps when masking is applied to patches outside the question-prior region (\textbf{Non-QP}). The masking probability denotes the fraction of non-overlapping patches that are removed.}
  \label{fig:masking_non_overlap_examples}
\end{figure}

We will perform the experiments with a fully trained \ModelName{} model at inference time. During the forward pass, we attach a forward hook to the model at the stage where the question heatmap is produced, i.e right before the gating module (Figure \ref{fig:architecture}). In the forward hook, we set the selected patches in the predicted question heatmap, or all the other non-overlapping patches in the predicted heatmap, and set them to zero. The overlap is calculated by taking the top-$k=0.70$ of the prior and the predicted heatmap. For the selected region, we mask each patch individually with a given probability. We use Bernoulli-sampling with a given probability. Then the forward pass is completed as normal with the masked heatmap, and we report the results. Figure \ref{fig:masking_overlap_examples} shows how the predicted question heatmap changes according to the masking probability of the overlapping patches. We observe that the important patches that overlay the main content of the page are set to zero. Figure \ref{fig:masking_non_overlap_examples} shows that the non-important patches are set to zero. One might argue that the performance drop reflects residual leakage rather than genuine faithfulness. However, the graded masking probabilities (10\%, 50\%, 90\%) rule this out. Masking overlapping patches causes a sharp, non-linear drop, while masking non-overlapping patches has minimal effect even at 90\%. This asymmetry confirms that the decoder is critically dependent on the gated region. 

\FloatBarrier
\section{Qualitative Analysis of Predictions}
\label{app:analysis_of_predictions}

\begin{figure}[htbp]
  \centering
  \begin{subfigure}[t]{0.32\textwidth}
    \centering
    \includegraphics[width=\linewidth]{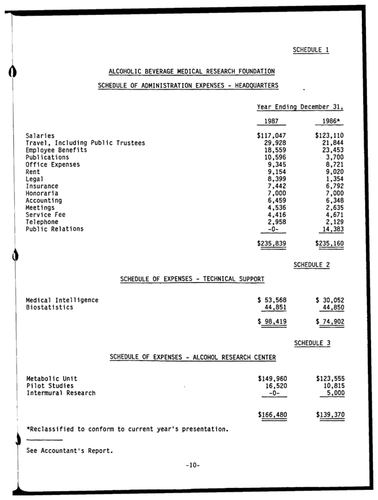}
    \caption{Original document given to the model.}
    \label{fig:example_1_a}
  \end{subfigure}\hfill
  \begin{subfigure}[t]{0.32\textwidth}
    \centering
    \includegraphics[width=\linewidth]{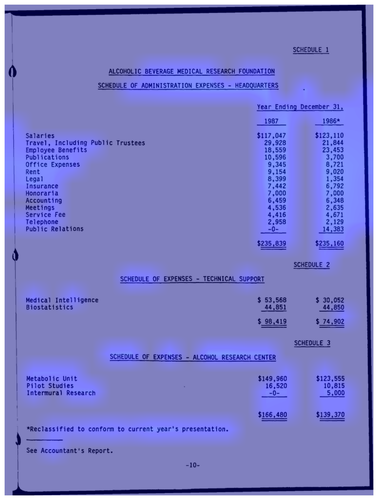}
    \caption{Question-heatmap ($\QMask$), visualized with a jet colormap (\textcolor{blue}{low} $\rightarrow$ \textcolor{orange}{high} relevance).}
    \label{fig:example_1_b}
  \end{subfigure}\hfill
  \begin{subfigure}[t]{0.32\textwidth}
    \centering
    \includegraphics[width=\linewidth]{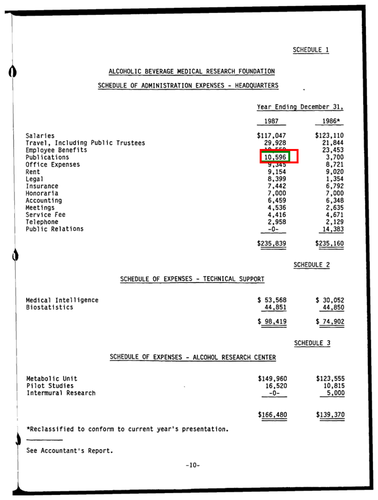}
    \caption{Predicts the answer region as a bounding box ($\ABox$, \textcolor{red}{red}), and the ground truth location ($\ABoxPrior$, \textcolor{ForestGreen}{green}).}
    \label{fig:example_1_c}
  \end{subfigure}

  \caption{Question: ``What is the Expenses for Publications for 1987?". The model predicted from the answer region: ``10,646", and the correct answer was ``10,596". Inspecting the predicted answer region, one can confirm that the model found the correct answer region, but were not able to correctly decode the answer. This model variant had lower accuracy due to low $\mathrm{AR} \approx 2.5$, but provides high faithfulness and compact explanation.}
  \label{fig:example_1}
\end{figure}

\begin{figure}[htbp]
  \centering
  \begin{subfigure}[t]{0.32\textwidth}
    \centering
    \includegraphics[width=\linewidth]{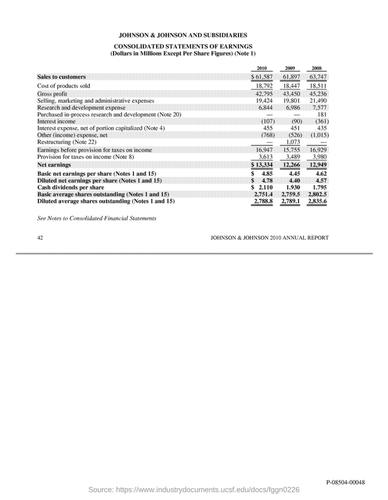}
    \caption{Original document given to the model.}
    \label{fig:example_2_a}
  \end{subfigure}\hfill
  \begin{subfigure}[t]{0.32\textwidth}
    \centering
    \includegraphics[width=\linewidth]{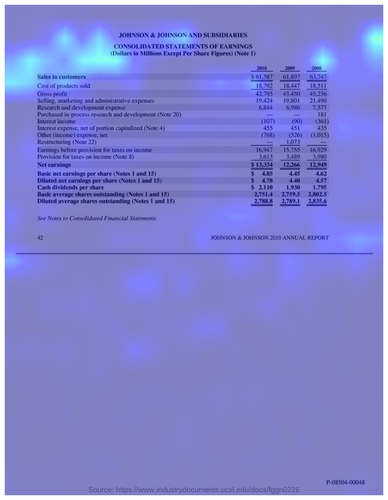}
    \caption{Question-heatmap ($\QMask$), visualized with a jet colormap (\textcolor{blue}{low} $\rightarrow$ \textcolor{orange}{high} relevance).}
    \label{fig:example_2_b}
  \end{subfigure}\hfill
  \begin{subfigure}[t]{0.32\textwidth}
    \centering
    \includegraphics[width=\linewidth]{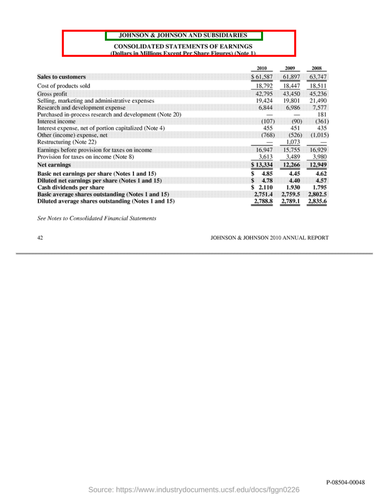}
    \caption{Predicts the answer region as a bounding box ($\ABox$, \textcolor{red}{red}), and the ground truth location ($\ABoxPrior$, \textcolor{ForestGreen}{green}).}
    \label{fig:example_2_c}
  \end{subfigure}

  \caption{Question: ``What is the name of the company mentioned at the top of the page?". The model predicted the correct answer from the answer region: ``Johnson \& Johnson and subsidiaries". The provided question heatmap seems more trivial at first glance, but it highlights part of ``Johnson" and upper regions of the model. Predicted region are higher due to being from the best model with $\mathrm{AR} \approx 19$. The given context is enough for the decoder to correctly decode the answer. This makes the model’s rationale interpretable.}
  \label{fig:example_2}
\end{figure}

\begin{figure}[htbp]
  \centering
  \begin{subfigure}[t]{0.32\textwidth}
    \centering
    \includegraphics[width=\linewidth]{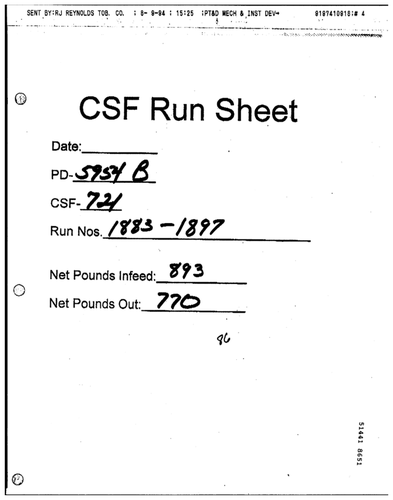}
    \caption{Original document given to the model.}
    \label{fig:example_4_a}
  \end{subfigure}\hfill
  \begin{subfigure}[t]{0.32\textwidth}
    \centering
    \includegraphics[width=\linewidth]{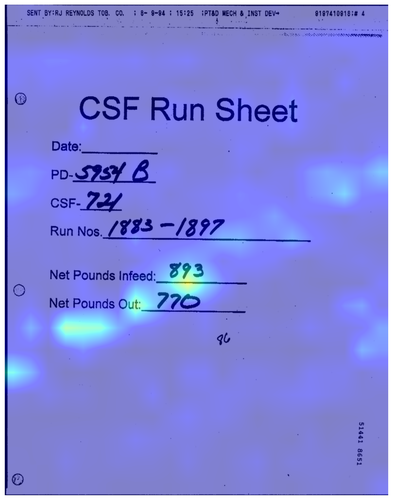}
    \caption{Question-heatmap ($\QMask$), visualized with a jet colormap (\textcolor{blue}{low} $\rightarrow$ \textcolor{orange}{high} relevance).}
    \label{fig:example_4_b}
  \end{subfigure}\hfill
  \begin{subfigure}[t]{0.32\textwidth}
    \centering
    \includegraphics[width=\linewidth]{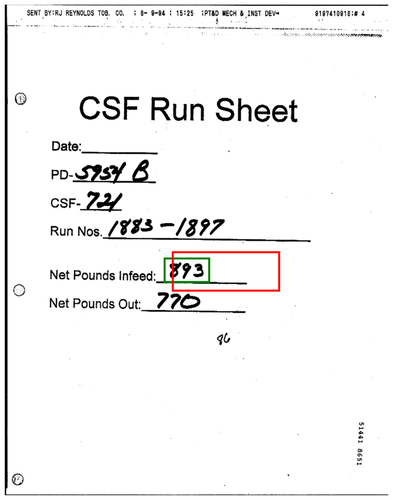}
    \caption{Predicts the answer region as a bounding box ($\ABox$, \textcolor{red}{red}), and the ground truth location ($\ABoxPrior$, \textcolor{ForestGreen}{green}).}
    \label{fig:example_4_c}
  \end{subfigure}

  \caption{Question: ``What is the Net Pound Infeed?". The model predicted the correct answer from the answer region: ``893". The provided question heatmap highlight the region in close proximity of the answer location. Predicted region are higher due to being from the best model with $\mathrm{AR} \approx 19$. The given context is enough for the decoder to correctly decode the answer.}
  \label{fig:example_4}
\end{figure}

We show some examples of predictions from our fully trained \ModelName{} model. On the left are the original document, in the middle is the question heatmap, and on the right is the answer localization. The latter shows the predicted answer location as a red rectangle and the ground truth as a green rectangle. Figures \ref{fig:example_1}, \ref{fig:example_2}, \ref{fig:example_3} and \ref{fig:example_4} show four different examples. 

\begin{figure}[htbp]
  \centering
  \begin{subfigure}[t]{0.32\textwidth}
    \centering
    \includegraphics[width=\linewidth]{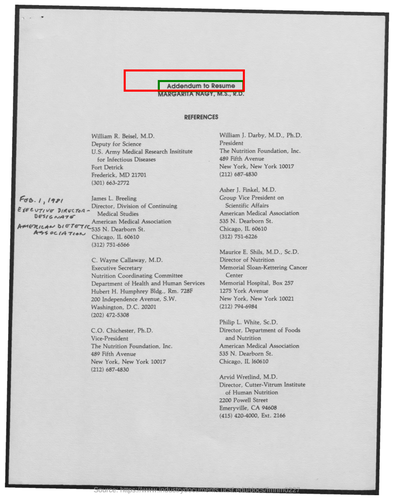}
    \caption{Question: ``What is the title of the document?", Predicted Text Answer: ``Addendum to Resume" and Ground Truth Answer: ``Addendum to Resume".}
    \label{fig:correct_1_a}
  \end{subfigure}\hfill
  \begin{subfigure}[t]{0.32\textwidth}
    \centering
    \includegraphics[width=\linewidth]{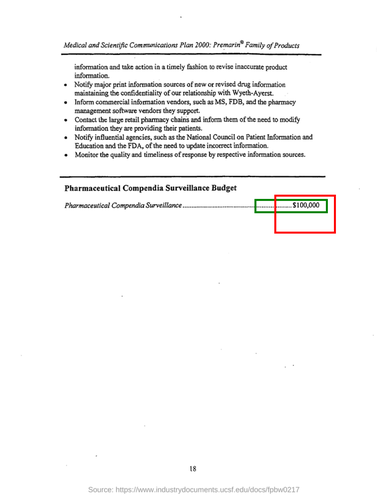}
    \caption{Question: ``What is the Budget Estimate for Pharmaceutical Compendia Surveillance?", Predicted Text Answer: ``\$100,000" and Ground Truth Answer: ``\$100,000".}
    \label{fig:correct_1_b}
  \end{subfigure}\hfill
  \begin{subfigure}[t]{0.32\textwidth}
    \centering
    \includegraphics[width=\linewidth]{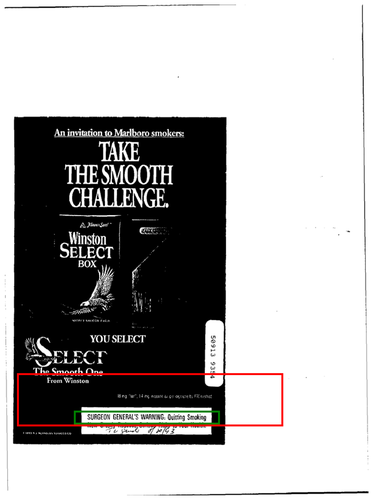}
    \caption{Question: ``What greatly reduces serious risks to your health, according to SURGEON GENERAL 2019S WARNING?", Predicted Text Answer: ``quitting smoking" and Ground Truth Answer: ``Quitting Smoking".}
    \label{fig:correct_1_c_append}
  \end{subfigure}

  \caption{Examples where the model correctly localizes the answer and decodes the correct answer.}
  \label{fig:correct_1}
\end{figure}

\begin{figure}[htbp]
  \centering
  \begin{subfigure}[t]{0.32\textwidth}
    \centering
    \includegraphics[width=\linewidth]{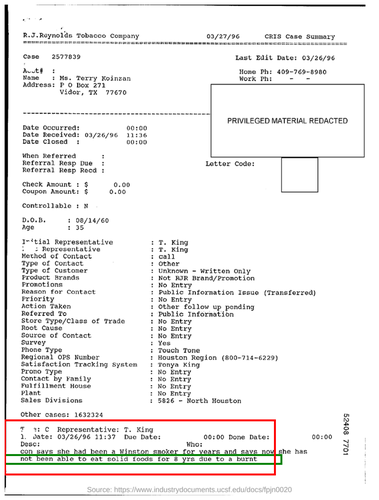}
    \caption{Question: ``For how long has the consumer not been able to eat solid food?", Predicted Text Answer: ``8 yrs" and Ground Truth Answer: ``8 yrs".}
    \label{fig:correct_2_a}
  \end{subfigure}\hfill
  \begin{subfigure}[t]{0.32\textwidth}
    \centering
    \includegraphics[width=\linewidth]{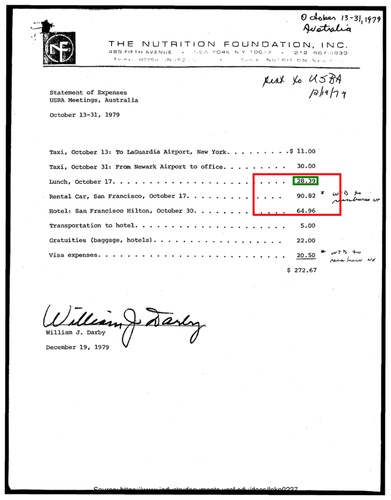}
    \caption{Question: ``What is the expense for lunch on October 17?", Predicted Text Answer: ``28.39" and Ground Truth Answer: ``28.39".}
    \label{fig:correct_2_b}
  \end{subfigure}\hfill
  \begin{subfigure}[t]{0.32\textwidth}
    \centering
    \includegraphics[width=\linewidth]{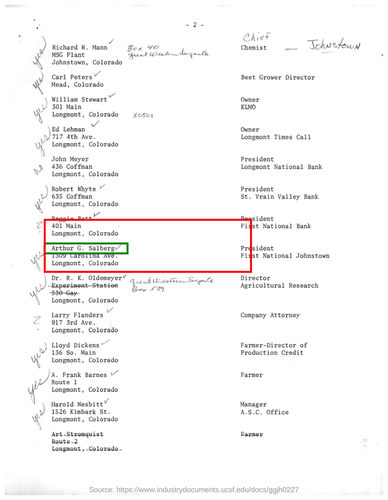}
    \caption{Question: ``Who is the president of First National Johnstown?", Predicted Text Answer: ``Arthur G. Salberg" and Ground Truth Answer: ``Arthur G. Salberg".}
    \label{fig:correct_2_c}
  \end{subfigure}

  \caption{Examples where the model correctly localizes the answer and decodes the correct answer.}
  \label{fig:correct_2}
\end{figure}

\begin{figure}[htbp]
  \centering
  \begin{subfigure}[t]{0.32\textwidth}
    \centering
    \includegraphics[width=\linewidth]{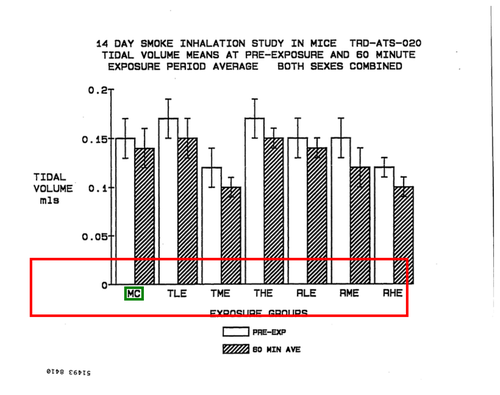}
    \caption{Question: ``Which is the first exposure group on the plot?", Predicted Text Answer: ``MC" and Ground Truth Answer: ``MC".}
    \label{fig:correct_3_a}
  \end{subfigure}\hfill
  \begin{subfigure}[t]{0.32\textwidth}
    \centering
    \includegraphics[width=\linewidth]{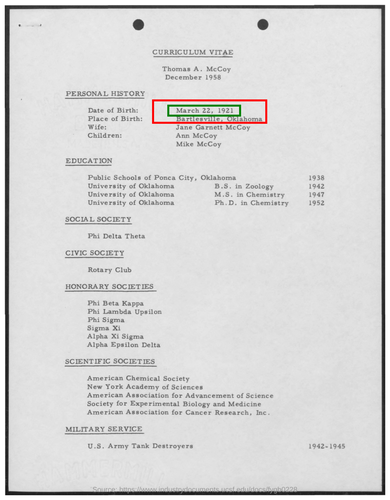}
    \caption{Question: ``What is Mr. McCoy's date of birth?", Predicted Text Answer: ``March 22, 1921" and Ground Truth Answer: ``March 22, 1921".}
    \label{fig:correct_3_b}
  \end{subfigure}\hfill
  \begin{subfigure}[t]{0.32\textwidth}
    \centering
    \includegraphics[width=\linewidth]{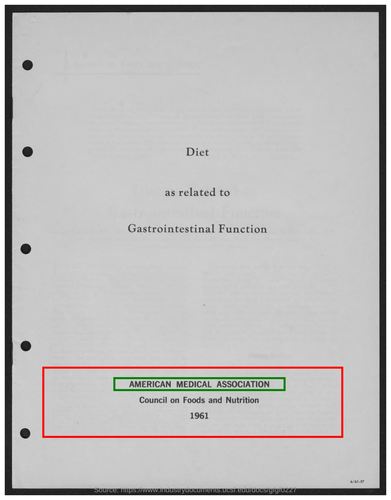}
    \caption{Question: ``What does AMA stand for?", Predicted Text Answer: ``American Medical Association" and Ground Truth Answer: ``American Medical Association".}
    \label{fig:correct_3_c}
  \end{subfigure}

  \caption{Examples where the model correctly localizes the answer and decodes the correct answer.}
  \label{fig:correct_3}
\end{figure}

\begin{figure}[htbp]
  \centering
  \begin{subfigure}[t]{0.32\textwidth}
    \centering
    \includegraphics[width=\linewidth]{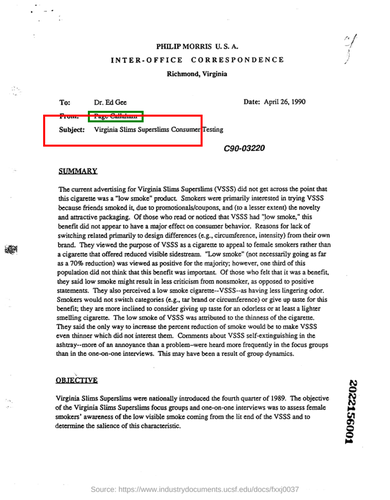}
    \caption{Question: ``What is the name of the sender?", Predicted Text Answer: ``Page Callaham" and Ground Truth Answer: ``Page Callaham".}
    \label{fig:partial_1_a}
  \end{subfigure}\hfill
  \begin{subfigure}[t]{0.32\textwidth}
    \centering
    \includegraphics[width=\linewidth]{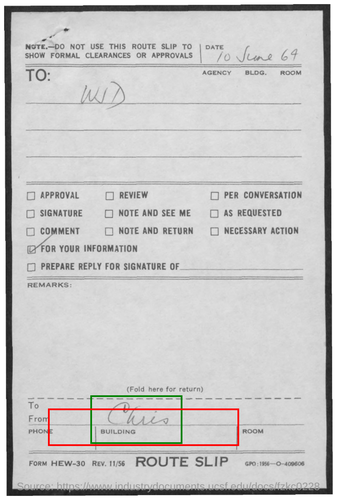}
    \caption{Question: ``Who is this slip from?", Predicted Text Answer: ``CHRIS" and Ground Truth Answer: ``Chris".}
    \label{fig:partial_1_b}
  \end{subfigure}\hfill
  \begin{subfigure}[t]{0.32\textwidth}
    \centering
    \includegraphics[width=\linewidth]{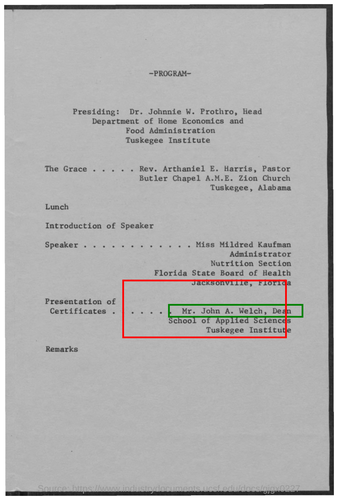}
    \caption{Question: ``Who is doing the presentation of certificates?", Predicted Text Answer: ``Mr. John A. Welch" and Ground Truth Answer: ``Mr. John A. Welch".}
    \label{fig:partial_1_c}
  \end{subfigure}

  \caption{Examples where the model partially includes the correct answer region. However, the model is still able to decode the correct text answer. Figure \ref{fig:partial_1_b} cuts off the upper part of the signature, but the decoder is still able to recover the answer. }
  \label{fig:partial_1}
\end{figure}

\begin{figure}[htbp]
  \centering
  \begin{subfigure}[t]{0.32\textwidth}
    \centering
    \includegraphics[width=\linewidth]{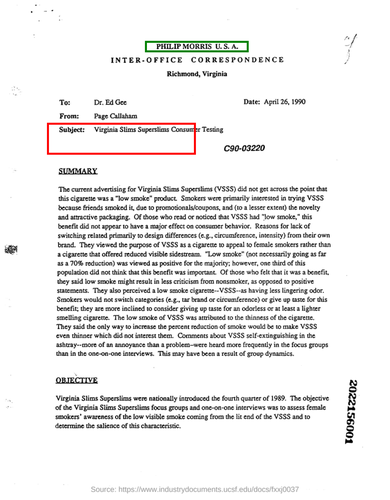}
    \caption{Question: ``What is the name present in the letter drop ?", Predicted Text Answer: ``Virginia Slims Superslims Consumer Testi" and Ground Truth Answer: ``PHILIP MORRIS U.S.A.".}
    \label{fig:incorrect_1_a}
  \end{subfigure}\hfill
  \begin{subfigure}[t]{0.32\textwidth}
    \centering
    \includegraphics[width=\linewidth]{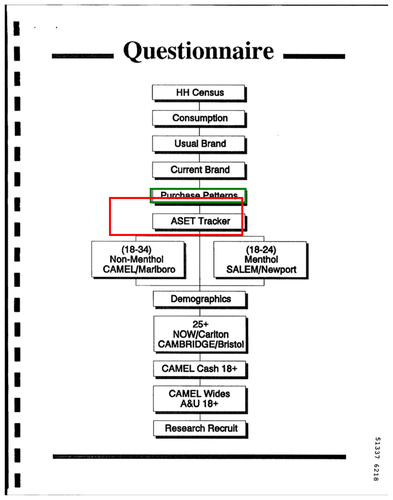}
    \caption{Question: ``After what step is the ASET ( age, sex, ethnicity, type) Tacker?", Predicted Text Answer: ``10" and Ground Truth Answer: ``Purchase Patterns".}
    \label{fig:incorrect_1_b}
  \end{subfigure}\hfill
  \begin{subfigure}[t]{0.32\textwidth}
    \centering
    \includegraphics[width=\linewidth]{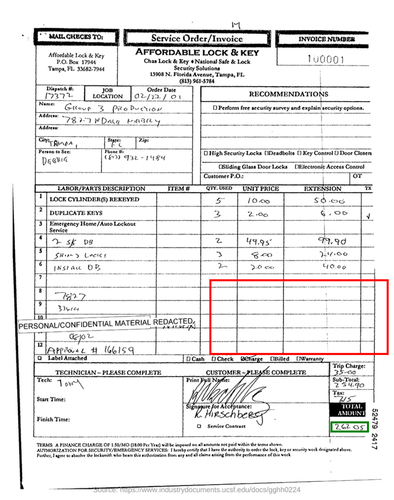}
    \caption{Question: ``What is the total bill amount?", Predicted Text Answer: ``\$1,000.00" and Ground Truth Answer: ``262.05".}
    \label{fig:incorrect_1_c}
  \end{subfigure}

  \caption{Incorrect examples. In Figure \ref{fig:incorrect_1_a}, the predicted answer location looks plausible, but overlaps the incorrect text span. The correct location is at the top of the page. Figures \ref{fig:incorrect_1_b} and \ref{fig:incorrect_1_c} both highlight parts of the document little to no question-relevant information, which lead to incorrect decoding. Most notably is Figure \ref{fig:incorrect_1_c}, which predicts an region that contains no readable text. Despite being wrong, these examples show that users can see where the model grounds its predictions, and that the corresponding decoded text is explicit.}
  \label{fig:incorrect_1}
\end{figure}

\paragraph{Predicted Answer Regions Examples.} We also show additional examples with only the predicted location with a red rectangle and ground truth with a green rectangle. Each example is categorized using the same definition as in Appendix \ref{app:answer_priors}: correct, partial, and incorrect. Figures \ref{fig:correct_1}, \ref{fig:correct_2} and \ref{fig:correct_3} show examples where the location of the answer was correctly predicted. Figure \ref{fig:partial_1} shows three examples where the answer location is partially correctly predicted. We observe that even when the correct answer is partially cut off from the answer location (Figure \ref{fig:partial_1_b}), the decoder is still able to recover the answer. Figure \ref{fig:incorrect_1} shows examples where the model predicts the incorrect answer location, and thus decodes incorrect predictions. The decoding process is then faithful to the predicted answer region by either (i) decoding the most logical answer given the predicted answer region (Figures \ref{fig:incorrect_1_a}), or (ii) outputting an irrelevant answer (Figures \ref{fig:incorrect_1_b} and \ref{fig:incorrect_1_c}).

\FloatBarrier
\section{Qualitative User Evaluation}
\label{app:user_evaluation}

We conducted a qualitative user evaluation to evaluate the utility of the framework. Participation was voluntary and uncompensated. In this Section, we go through the instructions, participants demographic, and each part of the questionnaire. 

\begin{figure}[ht!]
    \centering
    \includegraphics[width=0.9\linewidth, page=1]{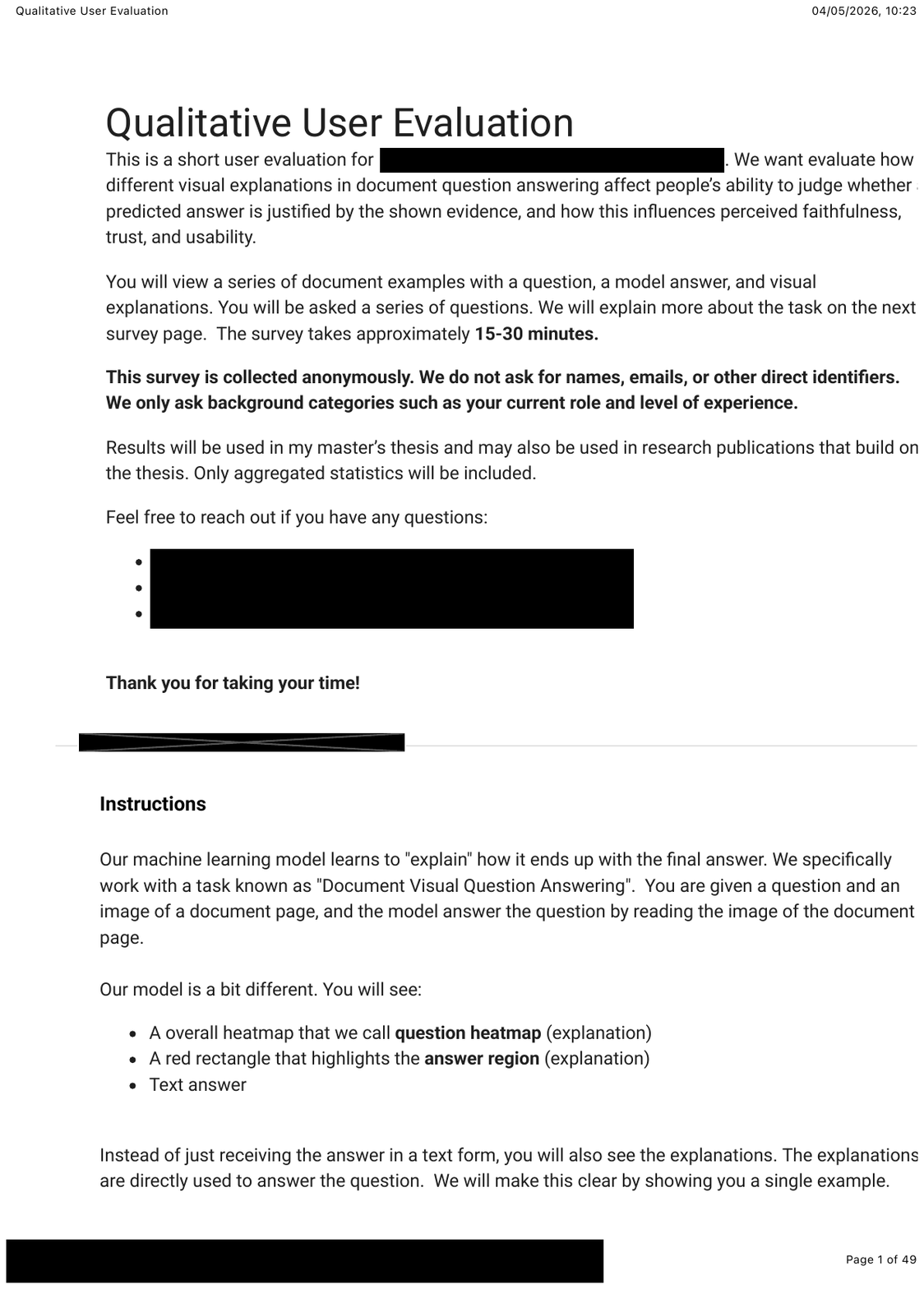}
    \caption{User evaluation instructions (page 1).}
    \label{fig:instructions_p1}
\end{figure}

\begin{figure}[ht!]
    \centering
    \includegraphics[width=0.9\linewidth, page=2]{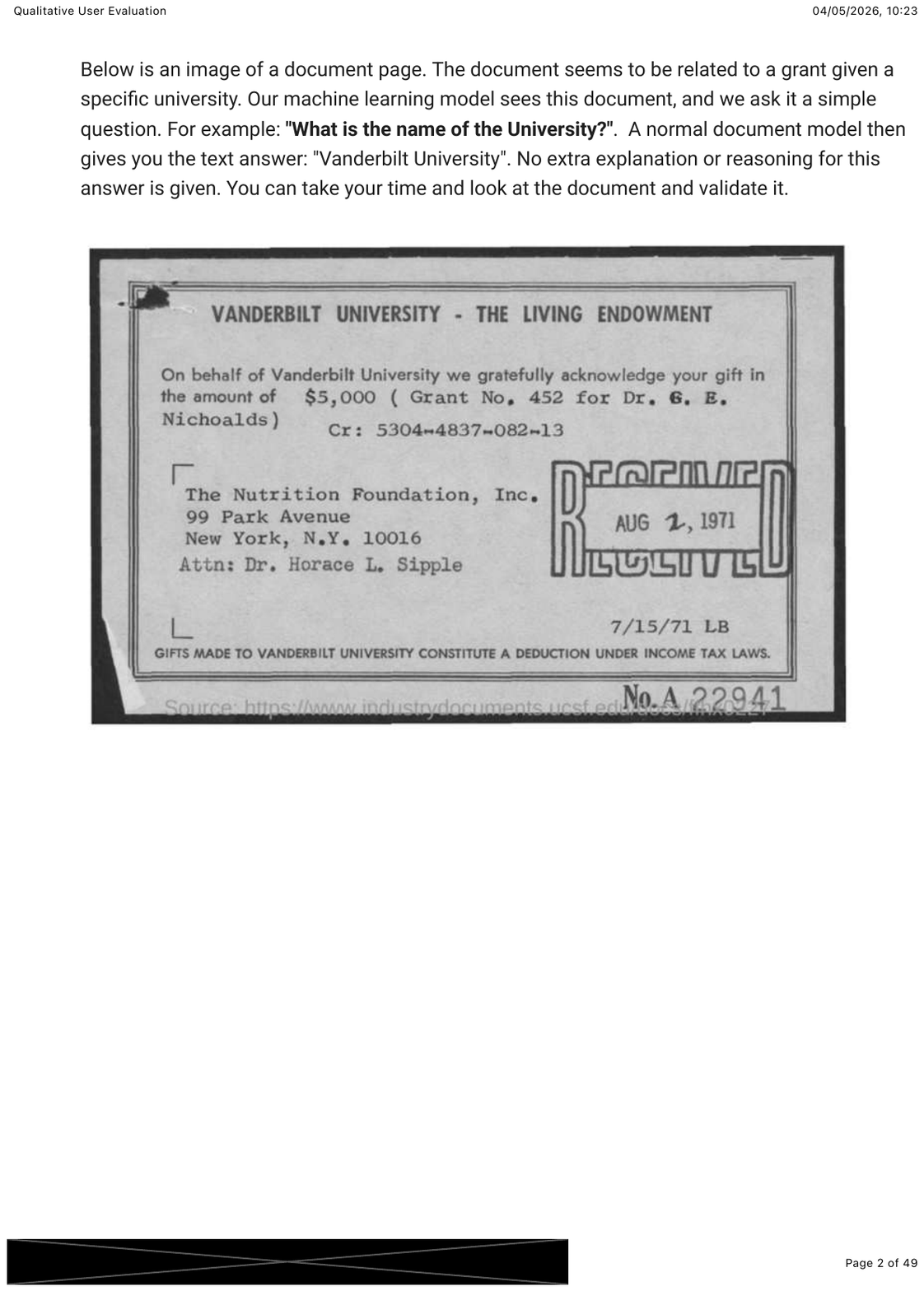}
    \caption{User evaluation instructions (page 2).}
    \label{fig:instructions_p2}
\end{figure}

\begin{figure}[ht!]
    \centering
    \includegraphics[width=0.9\linewidth, page=3]{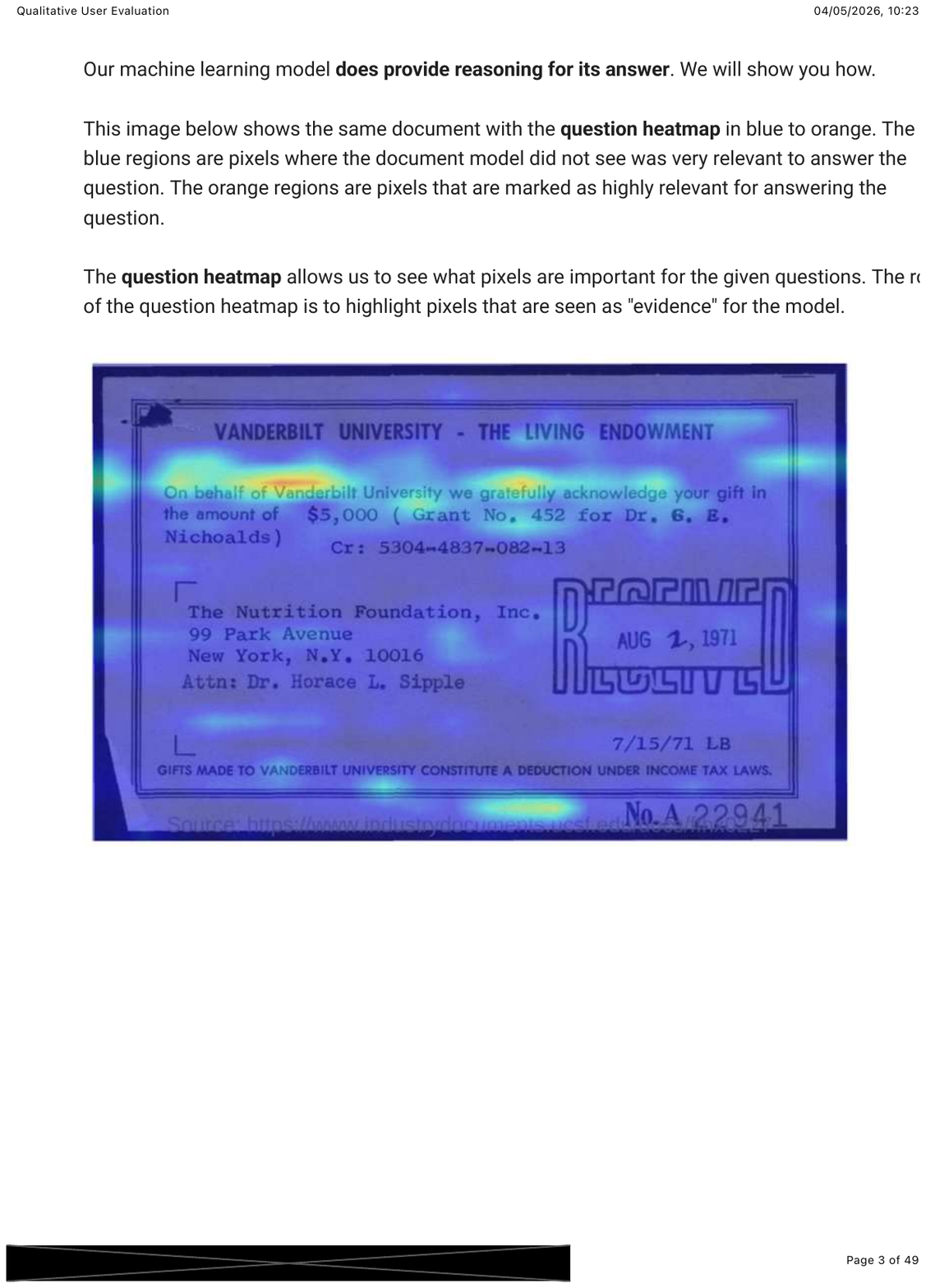}
    \caption{User evaluation instructions (page 3).}
    \label{fig:instructions_p3}
\end{figure}

\begin{figure}[ht!]
    \centering
    \includegraphics[width=0.9\linewidth, page=4]{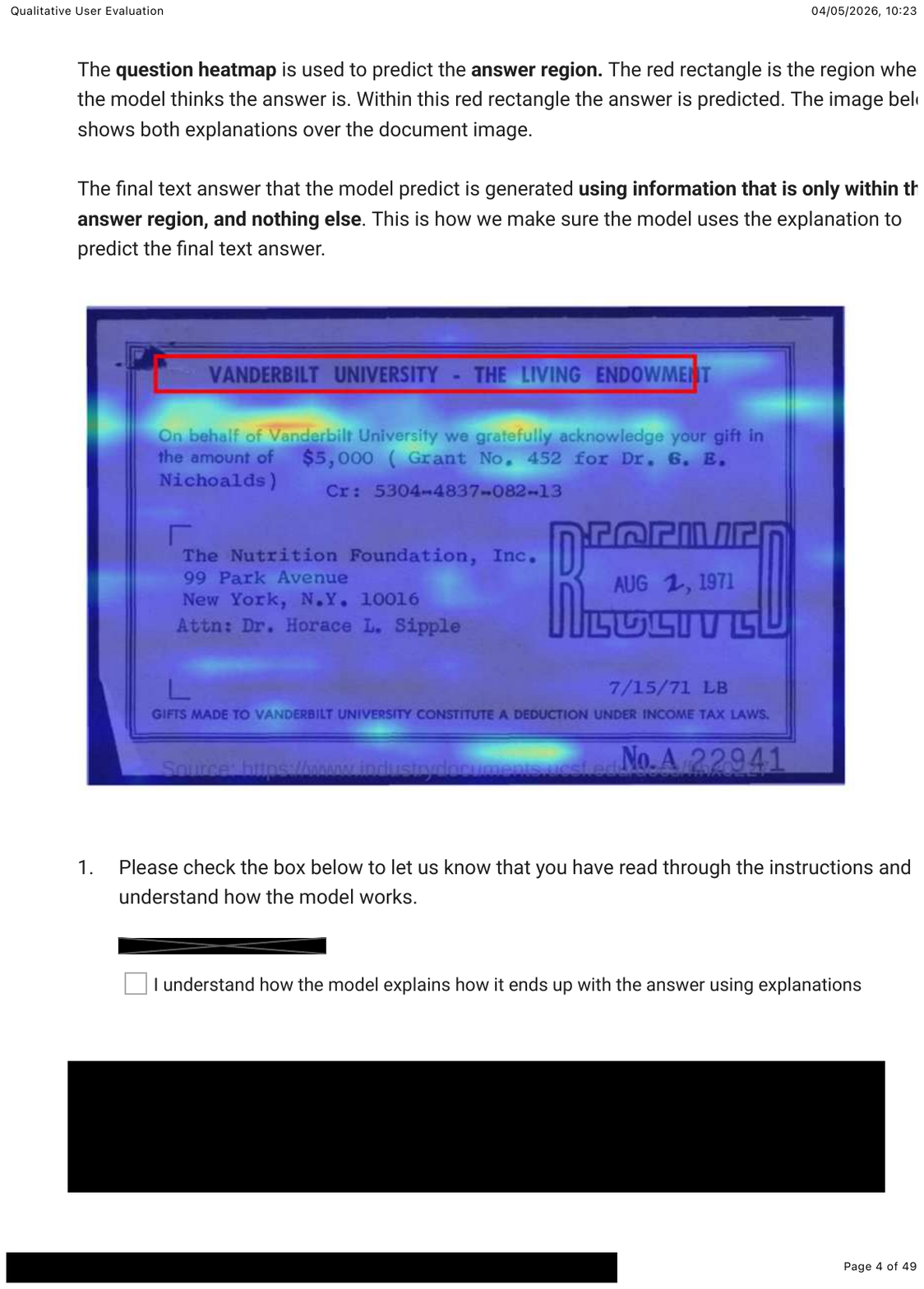}
    \caption{User evaluation instructions (page 4).}
    \label{fig:instructions_p4}
\end{figure}

\FloatBarrier
\paragraph{Instructions.} Figures~\ref{fig:instructions_p1}--\ref{fig:instructions_p4} shows the instructions given to participants. Figure~\ref{fig:instructions_p1} presents the motivation of the evaluation, while Figures~\ref{fig:instructions_p2}--\ref{fig:instructions_p4} explain the inner workings of the model. The instructions are written for a general audience without assuming prior knowledge of DocVQA or explainability methods. We redact identifying information from the pages to preserve anonymity.

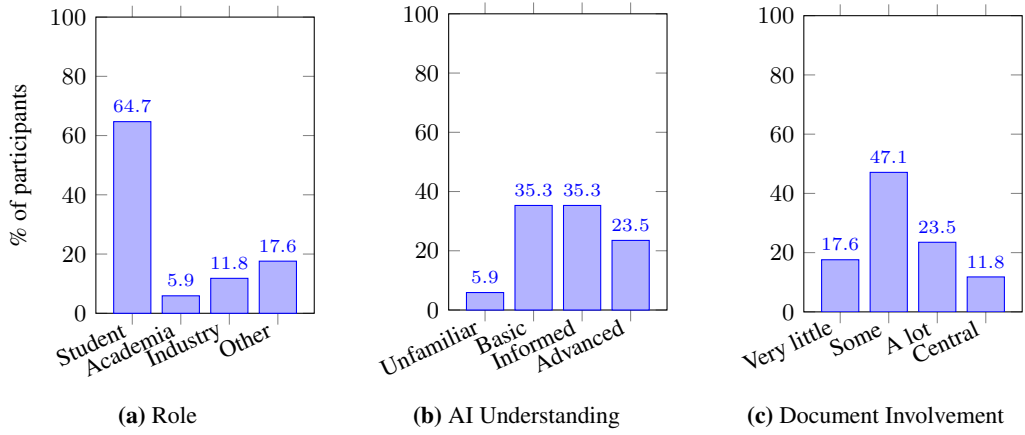
\begin{figure}[ht!]
    \centering
    \begin{subfigure}[t]{0.32\textwidth}
        \centering
        \begin{tikzpicture}
        \begin{axis}[
            width=\linewidth,
            height=5.5cm,
            ybar,
            bar width=14pt,
            ymin=0, ymax=100,
            ylabel={\% of participants},
            symbolic x coords={Student,Academia,Industry,Other},
            xtick=data,
            x tick label style={rotate=25,anchor=east,font=\small},
            nodes near coords,
            nodes near coords align={vertical},
            every node near coord/.append style={font=\scriptsize},
            enlarge x limits=0.25,
            ytick={0,20,40,60,80,100},
            label style={font=\small},
            tick label style={font=\small},
        ]
        \addplot coordinates {
            (Student, \StudentsPCT)
            (Academia, \AcademiaPCT)
            (Industry, \IndustryPCT)
            (Other, \OtherPCT)
        };
        \end{axis}
        \end{tikzpicture}
        \caption{Role}
        \label{fig:demo_role}
    \end{subfigure}\hfill
    \begin{subfigure}[t]{0.32\textwidth}
        \centering
        \begin{tikzpicture}
        \begin{axis}[
            width=\linewidth,
            height=5.5cm,
            ybar,
            bar width=14pt,
            ymin=0, ymax=100,
            symbolic x coords={Unfamiliar,Basic,Informed,Advanced},
            xtick=data,
            x tick label style={rotate=25,anchor=east,font=\small},
            nodes near coords,
            nodes near coords align={vertical},
            every node near coord/.append style={font=\scriptsize},
            enlarge x limits=0.25,
            ytick={0,20,40,60,80,100},
            label style={font=\small},
            tick label style={font=\small},
        ]
        \addplot coordinates {
            (Unfamiliar, \UnfamiliarPCT)
            (Basic, \BasicPCT)
            (Informed, \InformedPCT)
            (Advanced, \AdvancedPCT)
        };
        \end{axis}
        \end{tikzpicture}
        \caption{AI Understanding}
        \label{fig:demo_ai}
    \end{subfigure}\hfill
    \begin{subfigure}[t]{0.32\textwidth}
        \centering
        \begin{tikzpicture}
        \begin{axis}[
            width=\linewidth,
            height=5.5cm,
            ybar,
            bar width=14pt,
            ymin=0, ymax=100,
            symbolic x coords={Very little,Some,A lot,Central},
            xtick=data,
            x tick label style={rotate=25,anchor=east,font=\small},
            nodes near coords,
            nodes near coords align={vertical},
            every node near coord/.append style={font=\scriptsize},
            enlarge x limits=0.25,
            ytick={0,20,40,60,80,100},
            label style={font=\small},
            tick label style={font=\small},
        ]
        \addplot coordinates {
            (Very little, \VeryLittlePCT)
            (Some, \SomePCT)
            (A lot, \ALotPCT)
            (Central, \MostPCT)
        };
        \end{axis}
        \end{tikzpicture}
        \caption{Document Involvement}
        \label{fig:demo_docs}
    \end{subfigure}
    \caption{Participant demographics (\UserEvalParticipants{} participants).}
    \label{fig:user_eval_demographics}
\end{figure}

\FloatBarrier
\paragraph{Demographic of Participants.} Participants included students (\NumStudents, \pgfmathprintnumber{\StudentsPCT}\%), academia (\NumAcademia, \pgfmathprintnumber{\AcademiaPCT}\%), industry practitioners (\NumIndustry, \pgfmathprintnumber{\IndustryPCT}\%), and other (\NumOther, \pgfmathprintnumber{\OtherPCT}\%). AI familiarity ranged from unfamiliar (\NumUnfamiliar) to advanced (\NumAdvanced), and document involvement from very little (\NumVeryLittle) to central (\NumMost). Participant demographics are summarised in Figure~\ref{fig:user_eval_demographics}. 

\paragraph{Part 1: Answer Justification.} Participants were shown \BinaryExamples{} examples (\BinaryGoodExamples{} correct, \BinaryBadExamples{} incorrect predictions\footnote{Participants were not informed of the distribution of correct and incorrect predictions. This is also the case for part 2.}) and asked whether the explanation sufficiently justified the prediction, whether they believed the model predicted correctly, and rated their confidence and the quality of individual explanation components on 7-point Likert scales. This yields $\pgfmathparse{int(\BinaryExamples * \UserEvalParticipants)}\pgfmathresult$ total evaluations. Table~\ref{tab:user_eval_part1} shows the results from part 1.

\begin{table}[ht!]
    \centering
    \small
    \caption{Part 1 Answer justification results. ``Sufficient'' indicates whether participants judged the explanation as sufficient evidence. ``Correct'' indicates the rate at which participants judged the model's prediction to be correct (1 = correct, 0 = incorrect). For the Bad group, a low value indicates that participants successfully identified the prediction as wrong. Confidence, rectangle quality, and heatmap relevance are reported as mean $\pm$ std on 7-point Likert scales.}
    \begin{tabular}{lccccc}
        \toprule
        \textbf{Group} & \textbf{Sufficient} & \textbf{Correct} & \textbf{Confidence} & \textbf{Rect.\ Quality} & \textbf{Heatmap Rel.} \\
        \midrule
        Correct (Ex.\ 1--4) & \pgfmathprintnumber[precision=2]{\GoodSufficientMean} $\pm$ \pgfmathprintnumber[precision=2]{\GoodSufficientStd} & \pgfmathprintnumber[fixed, precision=2]{\GoodCorrectMean} $\pm$ \pgfmathprintnumber[precision=2]{\GoodCorrectStd} & \pgfmathprintnumber[precision=2]{\GoodConfMean} $\pm$ \pgfmathprintnumber[precision=2]{\GoodConfStd} & \pgfmathprintnumber[precision=2]{\GoodRectMean} $\pm$ \pgfmathprintnumber[precision=2]{\GoodRectStd} & \pgfmathprintnumber[precision=2]{\GoodHeatMean} $\pm$ \pgfmathprintnumber[precision=2]{\GoodHeatStd} \\
        Bad (Ex.\ 5--6) & \pgfmathprintnumber[precision=2]{\BadSufficientMean} $\pm$ \pgfmathprintnumber[precision=2]{\BadSufficientStd} & \pgfmathprintnumber[fixed,precision=2]{\BadCorrectMean} $\pm$ \pgfmathprintnumber[precision=2]{\BadCorrectStd} & \pgfmathprintnumber[precision=2]{\BadConfMean} $\pm$ \pgfmathprintnumber[precision=2]{\BadConfStd} & \pgfmathprintnumber[precision=2]{\BadRectMean} $\pm$ \pgfmathprintnumber[precision=2]{\BadRectStd} & \pgfmathprintnumber[precision=2]{\BadHeatMean} $\pm$ \pgfmathprintnumber[precision=2]{\BadHeatStd} \\
        \bottomrule
    \end{tabular}
    \label{tab:user_eval_part1}
\end{table}

Participants reliably distinguished correct from incorrect predictions. For correct examples, participants identified the prediction as correct with a rate of $\pgfmathprintnumber[precision=2]{\GoodCorrectMean} \pm \pgfmathprintnumber[precision=2]{\GoodCorrectStd}$, compared to $\pgfmathprintnumber[fixed,precision=2]{\BadCorrectMean} \pm \pgfmathprintnumber[fixed,precision=2]{\BadCorrectStd}$ for incorrect examples. Confidence remained high in both cases ($\pgfmathprintnumber[precision=2]{\GoodConfMean}$ vs.\ $\pgfmathprintnumber[precision=2]{\BadConfMean}$), indicating that participants were confident in their assessments regardless of prediction correctness. Rectangle quality was rated substantially higher for correct predictions ($\pgfmathprintnumber[precision=2]{\GoodRectMean}$ vs.\ $\pgfmathprintnumber[precision=2]{\BadRectMean}$), confirming that explanation quality is perceptibly linked to prediction quality. Similarly, heatmap relevance was rated higher for correct predictions ($\pgfmathprintnumber[precision=2]{\GoodHeatMean}$ vs.\ $\pgfmathprintnumber[precision=2]{\BadHeatMean}$), indicating that participants found both explanation components more informative when the model's grounding was accurate.

\paragraph{Part 2: Answering with Explanations Only.} Participants were shown the additional \UserAnswerExamples{} examples but without the model's predicted answer. They were asked to answer the question using only the visual explanations (bounding box and heatmap) and the document. This tests whether explanations are actionable, i.e. whether they contain sufficient information for a human to recover the correct answer. Table~\ref{tab:user_eval_part2} shows the results of part 2.

\begin{table}[ht!]
    \centering
    \small
    \caption{Part 2: Answering with explanations only. ``Correct'' indicates whether participants recovered the correct answer using only the visual explanations (0 = "No", 1 = "Yes"). Confidence is reported using a 7-point Likert scale, where 7 is most confident.}
    \begin{tabular}{lcc}
        \toprule
        \textbf{Group} & \textbf{Correct} & \textbf{Confidence} \\
        \midrule
        Good (Ex.\ 1--4) & \pgfmathprintnumber[precision=2]{\PartTwoGoodCorrectMean} $\pm$ \pgfmathprintnumber[precision=2]{\PartTwoGoodCorrectStd} & \pgfmathprintnumber[precision=2]{\PartTwoGoodConfMean} $\pm$ \pgfmathprintnumber[precision=2]{\PartTwoGoodConfStd} \\
        Bad (Ex.\ 5--6) & \pgfmathprintnumber[precision=2]{\PartTwoBadCorrectMean} $\pm$ \pgfmathprintnumber[precision=2]{\PartTwoBadCorrectStd} & \pgfmathprintnumber[precision=2]{\PartTwoBadConfMean} $\pm$ \pgfmathprintnumber[precision=2]{\PartTwoBadConfStd} \\
        \bottomrule
    \end{tabular}
    \label{tab:user_eval_part2}
\end{table}

For correctly predicted examples, participants recovered the correct answer with high accuracy ($\pgfmathprintnumber[precision=2]{\PartTwoGoodCorrectMean} \pm \pgfmathprintnumber[precision=2]{\PartTwoGoodCorrectStd}$) and high confidence ($\pgfmathprintnumber[precision=2]{\PartTwoGoodConfMean} \pm \pgfmathprintnumber[precision=2]{\PartTwoGoodConfStd}$). For incorrect examples, participants were unable to recover the correct answer ($\pgfmathprintnumber[precision=2]{\PartTwoBadCorrectMean} \pm \pgfmathprintnumber[precision=2]{\PartTwoBadCorrectStd}$) and reported lower confidence ($\pgfmathprintnumber[precision=2]{\PartTwoBadConfMean} \pm \pgfmathprintnumber[precision=2]{\PartTwoBadConfStd}$). This confirms that \ModelName{}'s explanations are actionable when the model is correct and do not mislead users into false confidence when the model fails.

\paragraph{Part 3: Visualisation Preferences.} Participants compared two answer localization variants (rectangle vs.\ hard mask) and two heatmap variants (hard mask vs.\ coloured heatmap). The default answer localization draws a bounding box around the predicted region, while the alternative hard mask variant obscures all content outside the region with a dark overlay. Similarly, the default question heatmap uses a coloured gradient to indicate relevance, while the alternative renders a binary hard mask. Examples using the default visualisations are shown in Appendix~\ref{app:analysis_of_predictions}. Table~\ref{tab:user_eval_part3} shows the results.

\begin{table}[ht!]
    \centering
    \small
    \caption{Part 3: Visualisation preferences.}
    \begin{tabular}{llccc}
        \toprule
        \textbf{Component} & \textbf{Metric} & \textbf{Option A} & \textbf{Option B} & \textbf{No Pref.} \\
        \midrule
        \multirow{2}{*}{Answer Box} 
            & Preference & Rectangle: \BoxPrefRect & Mask: \BoxPrefMask & \BoxPrefNoPref \\
            & Trust (Likert) & \pgfmathprintnumber[precision=2]{\BoxRectTrustMean} $\pm$ \pgfmathprintnumber[precision=2]{\BoxRectTrustStd} & \pgfmathprintnumber[precision=2]{\BoxMaskTrustMean} $\pm$ \pgfmathprintnumber[precision=2]{\BoxMaskTrustStd} & -- \\
        \midrule
        \multirow{2}{*}{Heatmap}
            & Preference & Hard: \HeatPrefHard & Coloured: \HeatPrefColor & \HeatPrefNoPref \\
            & Misleading (Likert) & \pgfmathprintnumber[precision=2]{\HeatHardMisleadMean} $\pm$ \pgfmathprintnumber[precision=2]{\HeatHardMisleadStd} & \pgfmathprintnumber[precision=2]{\HeatColorMisleadMean} $\pm$ \pgfmathprintnumber[precision=2]{\HeatColorMisleadStd} & -- \\
        \bottomrule
    \end{tabular}
    \label{tab:user_eval_part3}
\end{table}

For answer localization, participants preferred the rectangle variant (\BoxPrefRect{} vs.\ \BoxPrefMask), citing exact span identification and less occlusion. For the question heatmap, participants preferred the coloured variant (\HeatPrefColor{} vs.\ \HeatPrefHard), citing better overview and less noise. However, the mixed preferences suggest that users may benefit from the ability to customise how explanations are rendered. A minority of participants reported no preference between variants.

\paragraph{Part 4: Post-Questionnaire.} After completing Parts 1-3, participants rated statements on perceived faithfulness, trust, and usability using 7-point Likert scales ($1 = $ strongly disagree, $7 = $ strongly agree). Table~\ref{tab:user_eval_part4} shows each statement with their agreement evaluation.

\begin{table}[ht!]
    \centering
    \small
    \caption{Part 4: Post-questionnaire results (7-point Likert scale). Items marked with $\dagger$ are negatively worded.}
    \begin{tabular}{p{0.65\linewidth}cc}
        \toprule
        \textbf{Statement} & \textbf{Mean} & \textbf{Std} \\
        \midrule
        \multicolumn{3}{l}{\textit{Faithfulness}} \\
        \UserQ{FA} The explanation generally highlighted evidence relevant to the question. & \pgfmathprintnumber[precision=2]{\FAMean} & \pgfmathprintnumber[precision=2]{\FAStd} \\
        \UserQ{FB} When the model produced an incorrect answer, the explanations tended to make this apparent. & \pgfmathprintnumber[precision=2]{\FBMean} & \pgfmathprintnumber[precision=2]{\FBStd} \\
        \UserQ{FC} The highlighted regions matched what the model relied on. & \pgfmathprintnumber[precision=2]{\FCMean} & \pgfmathprintnumber[precision=2]{\FCStd} \\
        \UserQ{FD} The answer rectangle was consistent with where I would expect the answer to be located. & \pgfmathprintnumber[precision=2]{\FDMean} & \pgfmathprintnumber[precision=2]{\FDStd} \\
        \UserQ{FE} The answer rectangle helped me verify whether the model's answer was supported. & \pgfmathprintnumber[precision=2]{\FEMean} & \pgfmathprintnumber[precision=2]{\FEStd} \\
        \UserQ{FF}$^\dagger$ In some cases, the answer rectangle highlighted text that did not support the answer. & \pgfmathprintnumber[precision=2]{\FFMean} & \pgfmathprintnumber[precision=2]{\FFStd} \\
        \UserQ{FG} The question heatmap was consistent with where I would look to answer the question. & \pgfmathprintnumber[precision=2]{\FGMean} & \pgfmathprintnumber[precision=2]{\FGStd} \\
        \UserQ{FH} The question heatmap helped explain why the model predicted the shown answer location. & \pgfmathprintnumber[precision=2]{\FHMean} & \pgfmathprintnumber[precision=2]{\FHStd} \\
        \UserQ{FI}$^\dagger$ In some cases the question heatmap looked plausible but highlighted irrelevant regions. & \pgfmathprintnumber[precision=2]{\FIMean} & \pgfmathprintnumber[precision=2]{\FIStd} \\
        \midrule
        \multicolumn{3}{l}{\textit{Trust}} \\
        \UserQ{TA} The explanations helped me decide when to trust the predicted answer. & \pgfmathprintnumber[precision=2]{\TAMean} & \pgfmathprintnumber[precision=2]{\TAStd} \\
        \UserQ{TB} I would rely more on the system when the highlighted evidence is strong and specific. & \pgfmathprintnumber[precision=2]{\TBMean} & \pgfmathprintnumber[precision=2]{\TBStd} \\
        \UserQ{TC} I would double-check answers even when the evidence looks strong. & \pgfmathprintnumber[precision=2]{\TCMean} & \pgfmathprintnumber[precision=2]{\TCStd} \\
        \midrule
        \multicolumn{3}{l}{\textit{Usability}} \\
        \UserQ{UA} The visualisations were easy to interpret quickly. & \pgfmathprintnumber[precision=2]{\UAMean} & \pgfmathprintnumber[precision=2]{\UAStd} \\
        \UserQ{UB}$^\dagger$ The visualisations felt visually cluttered. & \pgfmathprintnumber[precision=2]{\UBMean} & \pgfmathprintnumber[precision=2]{\UBStd} \\
        \UserQ{UC} Overall, the explanations were presented in a user-friendly way. & \pgfmathprintnumber[precision=2]{\UCMean} & \pgfmathprintnumber[precision=2]{\UCStd} \\
        \bottomrule
    \end{tabular}
    \label{tab:user_eval_part4}
\end{table}

Key findings of the post-questionnaire show that participants agreed that the explanations highlighted relevant evidence (\UserQref{FA}: $\pgfmathprintnumber[precision=2]{\FAMean} \pm \pgfmathprintnumber[precision=2]{\FAStd}$), that the answer rectangle helped verify predictions (\UserQref{FE}: $\pgfmathprintnumber[precision=2]{\FEMean} \pm \pgfmathprintnumber[precision=2]{\FEStd}$), and that the answer rectangle was consistent with the expected answer locations (\UserQref{FD}: $\pgfmathprintnumber[precision=2]{\FDMean} \pm \pgfmathprintnumber[precision=2]{\FDStd}$). Participants also reported that explanations helped decide when to trust predictions (\UserQref{TA}: $\pgfmathprintnumber[precision=2]{\TAMean} \pm \pgfmathprintnumber[precision=2]{\TAStd}$) and that they would double-check answers even when evidence looks strong (\UserQref{TC}: $\pgfmathprintnumber[precision=2]{\TCMean} \pm \pgfmathprintnumber[precision=2]{\TCStd}$), suggesting that explanations support verification rather than inducing blind trust. Overall usability was rated positively (\UserQref{UC}: $\pgfmathprintnumber[precision=2]{\UCMean} \pm \pgfmathprintnumber[precision=2]{\UCStd}$).

The user evaluation confirms three key findings: (1) participants reliably distinguished correct from incorrect predictions using \ModelName{}'s explanations, (2) explanations are actionable (users recovered correct answers from explanations alone with high accuracy), and (3) explanations support verification without inducing blind trust, as participants reported they would double-check model answers even when evidence appears strong.

\end{document}